\documentclass[final,twocolumn,10pt]{elsart3p}
\pdfoutput=1 
\usepackage{array}
\usepackage{natbib}
\usepackage{float}
\usepackage{graphicx}
\usepackage{amsmath}
\usepackage{amsfonts}
\usepackage{amssymb}
\usepackage{booktabs}
\usepackage{./ctable}
\usepackage{multirow}
\usepackage{hyperref}

\DeclareFontFamily{U}{msb}{}
\DeclareFontShape{U}{msb}{m}{n}{ <5> <6> <7> <8> <9> gen * msbm
               <10> <10.95> <12> <14.4> <17.28> <20.74> <24.88> msbm10}{} 
\DeclareSymbolFont{AMSb}{U}{msb}{m}{n}
\DeclareMathSymbol{\R}{\mathalpha}{AMSb}{"52}                                           
\newlength\onefig
\newlength\twofigs
\divide\twofigs by 2%
\newlength\twosmallfigs
\newlength\threefigs
\divide\threefigs by 3%
\newlength\threesmallfigs
\newlength\fourfigs
\divide\fourfigs by 4%
\newlength\sixfigs
\divide\sixfigs by 6%

\newcommand{\SF}{\bfseries \sffamily}
\newcommand{\bx} {{\mathbf x}}
\newcommand{\bX} {{\mathbf X}}

\newcommand{\br} {{\mathbf r}}

\newcommand{\bfi} {{\boldsymbol \phi}}
\newcommand{\barre}{\rule[0mm]{18pc}{0.5pt}}%

\begin{document}
\begin{frontmatter}
\title{Low Dimensional Embedding of fMRI
  datasets\thanksref{whereto}}
\thanks[whereto]{Submitted to Neuroimage, Sep. 2007. Revised Jan. 2008.}
\author[label1]{Xilin Shen,}
\author[label1]{Fran\c{c}ois G. Meyer\corauthref{cor1}}
\ead{fmeyer@Colorado.Edu\\}
\ead[url]{ece.colorado.edu\hspace*{-1pt}/\hspace*{-1pt}$\sim$fmeyer}
\corauth[cor1]{Corresponding author.}
\address[label1]{Department of Electrical Engineering, University of Colorado at Boulder}
\begin{abstract}
  We propose a novel method to embed a functional magnetic resonance
  imaging (fMRI) dataset in a low-dimensional space.  The embedding
  optimally preserves the local functional coupling between fMRI time
  series and provides a low-dimensional coordinate system for
  detecting activated voxels. To compute the embedding, we build a
  graph of functionally connected voxels. We use the commute time,
  instead of the geodesic distance, to measure functional distances on
  the graph. Because the commute time can be computed directly from
  the eigenvectors of (a symmetric version) the graph probability
  transition matrix, we use these eigenvectors to embed the dataset in
  low dimensions. After clustering the datasets in low dimensions,
  coherent structures emerge that can be easily interpreted.  We
  performed an extensive evaluation of our method comparing it to
  linear and nonlinear techniques using synthetic datasets and {\em in
    vivo} datasets.  We analyzed datasets from the EBC competition
  obtained with subjects interacting in an urban virtual reality
  environment. Our exploratory approach is able to detect
  independently visual areas (V1/V2, V5/MT), auditory areas, and
  language areas. Our method can be used to analyze fMRI collected
  during ``natural stimuli''.
\end{abstract}

\end{frontmatter}
\section{Introduction}
At a microscopic level, a large number of internal variables
associated with various physical and physiological phenomena
contribute to dynamic changes in functional magnetic resonance imaging
(fMRI) datasets.  FMRI provides a large scale (as compared to the
scale of neurons) measurement of neuronal activity, and we expect that
many of these variables will be coupled resulting in a low dimensional
set for all possible configurations of the activated fMRI signal. We
assume therefore that activated fMRI time series can be parametrized
by a small number of variables. This assumption is consistent with the
usage of low dimensional parametric models for detecting activated
voxels \citep{Petersson99b}. This assumption is also consistent with
the empirical findings obtained with principal components analysis
(PCA) and independent components analysis (ICA)
\citep{Biswal99,McKeown03}, where a small number of components are
sufficient to describe the variations of most activated temporal
patterns.  Both PCA and ICA make very strong assumptions about the
components: orthogonality and statistical independence,
respectively. Such constraints are convenient mathematically but have
no physiological justification, and complicate unnecessarily the
interpretation of the components \citep{Friston98b}. A second
limitation of PCA and ICA is that both methods only provide a linear
decomposition of the data \citep{Friston05}. There is no physiological
reason why the fMRI signal should be a linear combination of
eigen-images or eigen-time series. In practice, the first components
identified by PCA are often related to physiological artifacts
(e.g. breathing), or coherent spontaneous fluctuations
\citep{Raichle06a}. These artifacts can be responsible for most of the
variability in the dataset. Stimulus triggered changes, which are much
more subtle, rarely appear among the first components.

The contribution of this paper is a novel exploratory method to
construct an optimal coordinate system that reduces the dimensionality
of the dataset while preserving the functional connectivity between
voxels \citep{Sporns00}.  First, we define a distance between time
series that quantifies the functional coupling \citep{Fox05}, or
connectivity between the corresponding voxels. We then construct an
embedding that preserves this functional connectivity across the
entire brain.  After embedding the dataset in a lower dimensional
space, time series are clustered into coherent groups.  This new
parametrization results in a clear separation of the time series into:
(1) response to a stimulus, (2) coherent physiological signals, (3)
artifacts, and (4) background activity. We performed an extensive
evaluation of our method comparing it to linear and nonlinear
techniques using synthetic datasets and {\em in vivo} datasets.
\section{Methods}
\subsection{Overview of our approach}
Our goal is to find a new parametrization of an fMRI dataset,
effectively replacing the time series by a small set of features, or
coordinates, that facilitate the identification of task-related
hemodynamic responses to the stimulus. The new coordinates will also
be able to reveal the presence of physical or physiological processes
that have an intrinsic low dimension. Such processes can be described
with a small number of parameters (dimensions) in an appropriate
representation (set of basis functions). These time series should be
contrasted with noise time series that have a very diffuse
representation in any basis. Example of such low-dimensional processes
include task-related hemodynamic responses, non-task-related
physiological rhythms (breathing and heart beating motion). We expect
that only a small fraction of all time series will have a
low-dimensional representation. The remaining time series will be
engaged in a spontaneous intrinsic activity \citep{Raichle06a}.  We
call these time series {\em background time series}. As shown in our
experiments, background time series are more complex, and cannot be
well approximated with a small number of parameters.

We now introduce some notations that will be used throughout the paper.
Let $\bX$ denote an fMRI dataset composed of $T$ scans, each comprised of $N$
voxels, which is represented as a $N \times T$ matrix,
\begin{equation}
\bX=
\begin{bmatrix}
x_1(1) &  \cdots & x_1(T) \\
 \vdots  & \vdots & \vdots  \\
x_N(1) &  \cdots & x_N(T)
\end{bmatrix}.
\end{equation}
Row $i$ of $\bX$ is the time series $\bx_i=[x_i(1), \cdots,x_i(T)]$
generated from voxel $i$.  Column $j$ is the $j^\text{th}$ scan
unrolled as a $1 \times N$ vector.  In this work, we regard $\bx_i$ as
a point in $\R^T$, with $T$ coordinates. 

We seek a new parametrization of the dataset that optimally preserves
the local functional coupling between time series. Most methods of
reduction of dimensionality used for fMRI are linear: each $\bx_i$ is
projected onto a set of components $\bfi_k$. The resulting
coefficients $<\bx_i,\bfi_k>,k=0,\cdots,K-1$ serve as the new
coordinates in the low dimensional representation.  However, in the
presence of nonlinearity in the organization of the $\bx_i$ in $\R^T$,
a linear mapping may distort local functional correlations. This
distortion will make the clustering of the dataset more difficult. Our
experiments with {\em in vivo} data confirm that the subsets formed by
the low-dimensional time series have a nonlinear geometry and cannot
be mapped onto a linear subspace without significant distortion. We
propose therefore to use a nonlinear map $\Psi$ to represent the
dataset $\bX$ in low dimensions.  Because the map $\Psi$ is able to
preserve the local functional coupling between voxels, low dimensional
coherent structures can easily be detected with a clustering
algorithm. Finally, the temporal and the spatial patterns associated
with each cluster are examined and the cluster that corresponds to the
task-related response is identified. In summary, our approach includes
the following three steps:
\begin{enumerate}
\item Low dimensional embedding of the dataset;
\item Clustering of the dataset using the new parameterization;
\item Identification of the set of activated time series.
\end{enumerate}
\begin{figure}[H]
\centerline{
  \includegraphics[width=18pc]{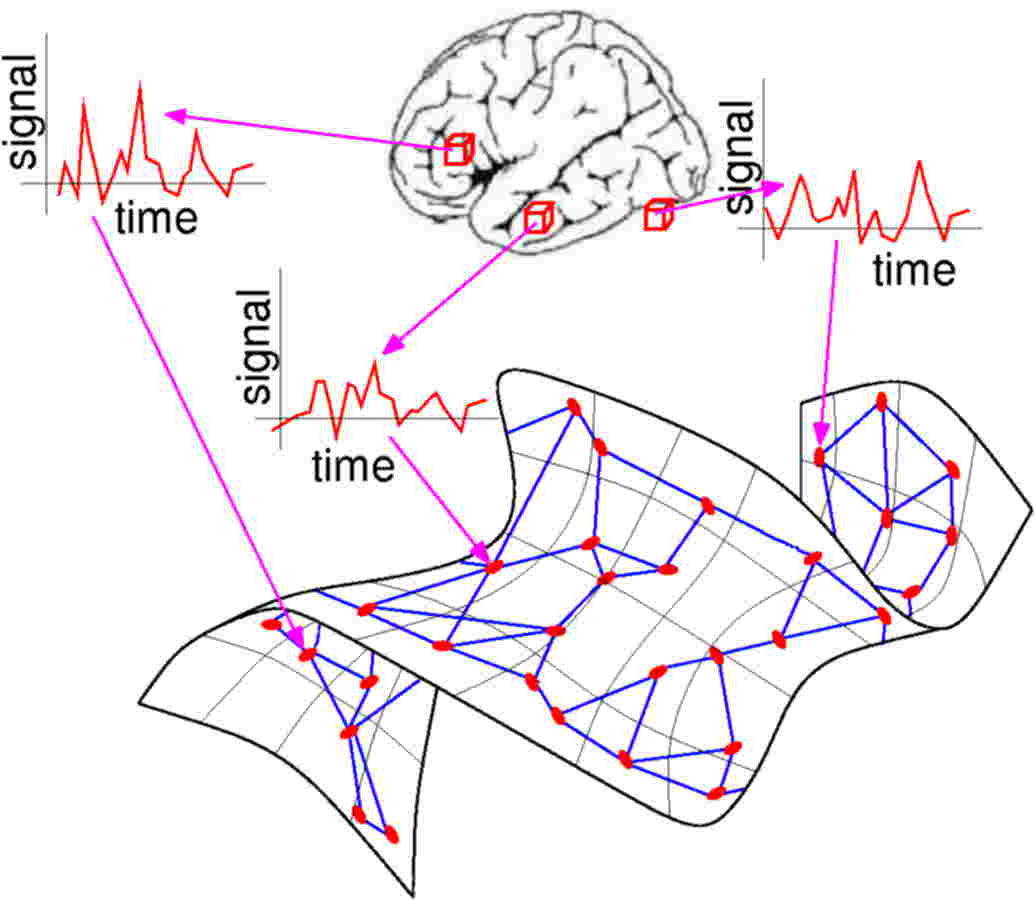}}
\caption{The network of functionally correlated voxels, represented by a
  graph, encodes the functional connectivity between time series.
  \label{network}}
\end{figure}
\subsection{The connectivity graph: a network of functionally correlated
  voxels \label{connectgraph}}
In order to construct the nonlinear map $\Psi$ we need to estimate the
functional correlation between voxels. We characterize the functional
correlation between voxels with a network. Similar networks have been
constructed to study functional connectivity in
\citep{Achard06,Caclin06,Eguiluz05,Fox05,Sporns00}.  We represent the
network by a graph $G$ that is constructed as follows.  The time series
$\bx_i$ originating from voxel $i$ becomes the node (or vertex) $\bx_i$ of
the graph\footnote{We slightly abuse notation here: $\bx_i$ is a time
series, as well as a node on the graph.}. Edges between vertices quantify
the functional connectivity. Each node $\bx_i$ is connected to its $n_n$
nearest neighbors according to the Euclidean distance between the time
series, $\| \bx_i - \bx_j\|= (\sum_{t=1}^T (x_i(t) - x_j(t))^2)^{1/2}$
(Fig. \ref{network}).  The weight $W_{i,j}$ on the edge $\{i,j\}$
quantifies the functional proximity between voxels $i$ and $j$ and is
defined by
\begin{equation}
W_{i,j} = 
\begin{cases}
e^{\displaystyle - \|\bx_i-\bx_j\|^2/ \sigma^2}, 
& \text{if $\bx_i$ is connected to $\bx_j$},\\
0 & \text{otherwise.}
\end{cases}
\label{weight}
\end{equation}
The scaling factor $\sigma$ modulates the definition of proximity
measured by the weight (\ref{weight}). If $\sigma$ is very large, then
for all the nearest neighbors $\bx_j$ of $\bx_i$, we have $W_{i,j}
\approx 1$ (see (\ref{weight})), and the transition probability is the
same for all the neighbors, $P_{i,j} = 1/n_n$.  This choice of
$\sigma$ promotes a very fast diffusion of the random walk through
the dataset, and blurs the distinction between activated and
background time series. On the other hand, if $\sigma$ is extremely
small, then $P_{i,j} = 0$ for all the neighbors such that $\|\bx_i -
\bx_j \|> 0$. Only if the distance between $\bx_i$ and $\bx_j$ is zero
(or very small) the transition probability is non zero. This choice of
$\sigma$ accentuates the difference between the time series, but is
more sensitive to noise. In practice, we found that the universal choice
\begin{equation*}
\sigma = n \times \min_{i<j} \|\bx_i-\bx_j\|
\end{equation*}
where $n\in (0,5]$ (we used  $n=2$ for all experiments) is usually optimal.

The weighted graph ${G}$ is fully characterized by the $N \times N$
weight matrix $\mathbf W$ with entries $W_{i,j}$. Let  $\mathbf D$  be
the diagonal degree matrix with entries $D_{i,i} =
\sum_{j}W_{i,j}$. The spatial coordinates of the voxels
$i$ and $j$ are currently not used in the computation of $W_{i,j}$.
We know that spatial information can be useful: truly activated voxels
tend to be spatially clustered. However, spatial proximity should be
measured along the cortical ribbon, and not in the 3-D volume. 
\subsection{A new way to measure functional distances between voxels
\label{commute}}
Once the network of functionally connected voxels is created, we need
to define a distance between any two vertices $\bx_i$ and $\bx_j$ in
the network. This distance should reflect the topology of the graph,
but should also be able to distinguish between strongly connected
vertices (when the voxels $i$ and $j$ belong to the same functional
region) and weakly connected vertices (when $i$ and $j$ have similar
time series $\bx_i$ and $\bx_j$, but belong to different functional
regions). The Euclidean distance that is used to construct the graph
is only useful locally: we can use it to compare voxels that are
functionally similar, but we should use a different distance to
compare voxels that may not be functionally similar. As shown in the
experiments, the shortest distance $\delta(\bx_i,\bx_j)$ between two
nodes $\bx_i$ and $\bx_j$ measured along the edges of the graph is
very sensitive to short-circuits created by the noise in the data. A
standard alternative to the geodesic distance is the commute time,
$\kappa(\bx_i,\bx_j)$, that quantifies the expected path length
between $\bx_i$ and $\bx_j$ for a random walk started at $\bx_i$
\citep{Bremaud99}.

We review here the concept of commute time in the context of a random walk on
a graph. We show how the commute time can be computed easily
from the eigenvectors of $\mathbf D^{-1/2} \mathbf W \mathbf D^{-1/2}$. Let
us consider a random walk $Z_n$ on the connectivity graph. The walk starts
at $\bx_i$, and evolves on the graph with the transition probability
$\mathbf P = \mathbf D^{-1} \mathbf W$,
\begin{equation}
P_{i,j} = W_{i,j}/(\sum_j W_{i,j}).
\label{transition}
\end{equation}
If the walk is at $\bx_i$, it jumps to one of the nearest neighbors,
$\bx_j$, with probability $P_{i,j}$.  The walk first visits all
voxels in the same functional area before jumping to a different functional
area. Indeed, if voxel $i$ and $j$ are in the same functional area, and
voxel $k$ is in a different functional area, then we expect that $\|\bx_i -
\bx_j\| \ll \|\bx_i - \bx_k\|$, and therefore $P_{i,j} \gg P_{i,k}$. This
observation can be quantified by computing the average hitting time that
measures the number of steps that it takes for the random walk started at
$\bx_i$ to hit $\bx_j$ for the first time \citep{Bremaud99},
\begin{equation*}
H(\bx_i,\bx_j) = E_i[T_j] \quad \text{with} \quad T_j=\min \{n\ge 0; Z_n =j\}.
\end{equation*}
The hitting time is not symmetric, and cannot be a distance on the
graph. A proper distance is provided by a symmetric version of $H$,
called the commute time, 
\citep{Bremaud99}, 
\begin{equation}
\kappa(\bx_i,\bx_j) = H(\bx_i,\bx_j) + H(\bx_j,\bx_i) = E_i[T_j] + E_j[T_i].
\end{equation}
As one would expect, $\kappa(\bx_i,\bx_j)$ increases with the geodesic
distance $\delta(\bx_i,\bx_j)$.  Unlike the geodesic distance,
$\kappa(\bx_i,\bx_j)$ decreases when the number of paths between the nodes
increases.  

{\noindent \em Commute time and clustering coefficient}\\
The commute time is greatly influenced by the richness of the connections
between any two nodes of the network. This concept can be quantified by the
clustering coefficient. Let $\bx_i$ be a node with $n_n$ neighbors. The
neighbors of $\bx_i$ may, or may not, be neighbors of one another. To asses
the transitivity of the connections, we can compute the total number of
edges, $e_i$, that exist between all the neighbors of $\bx_i$.  The
clustering coefficient \citep{Albert02} is $C_i = 2e_i/[n_n(n_n -1)]$.
The maximum value of $C_i$ is $1$ and is achieved when each neighbor of
$\bx_i$ is connected to all the other neighbors of $\bx_i$ (they form a
clique). If the average clustering coefficient, computed over all nodes of
the network, is close to 1, then there will always be multiple routes
between any two nodes $\bx_i$ and $\bx_j$, and the commute time will
\begin{figure}[H]
\centerline{
  \includegraphics[width=18pc]{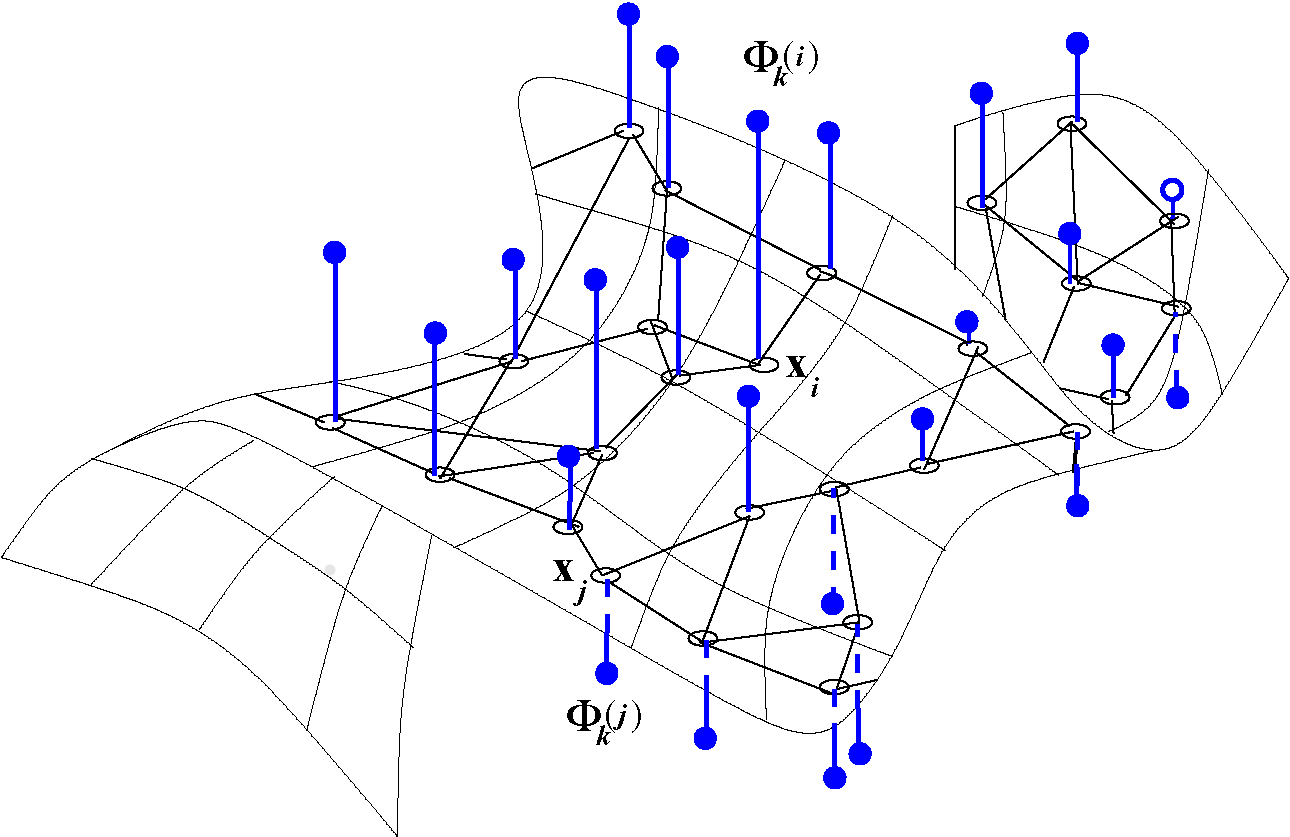}}
\caption{The eigenvector $\bfi_k$ as a function defined on the nodes of the graph.  
  \label{eigenfunction}}
\end{figure}
\noindent  remain
small. \cite{Eguiluz05} measured clustering coefficients in networks of
functionally connected voxels in fMRI that were indicative of scale-free
small-world networks.  \cite{Achard06} identified networks of richly
connected hubs in the cortex, and have also shown that the functional
network (as measured by fMRI) exhibits the ``small world'' property. The
commute time provides a distance on the graph that takes into account the
abundance of connections that may exist between two nodes of the graph.

{\em A spectral decomposition of commute time}\\
As explained in the Appendix, the commute time can be conveniently computed
from the eigenvector $\bfi_1, \cdots, \bfi_N$ of the symmetric matrix
$\mathbf D^{-1/2} \mathbf P \mathbf D^{-1/2}$. The corresponding
eigenvalues are between $-1$ and $1$, and can be labeled such that $-1 \leq
\lambda_N \cdots \leq \lambda_2 < \lambda_1 =1$.  Each eigenvector $\bfi_k$
is a vector with $N$ coordinates, one for each node of the graph $G$. 
We therefore write
\begin{equation}
\bfi_k = 
\left[
\bfi_k(1), \bfi_k(2),\cdots, \bfi_k(N)
\right]^T
\end{equation}
to emphasize the fact that we consider $\bfi_k$ to be a function defined on
the nodes of the graph (Fig. \ref{eigenfunction}).  According to
(\ref{isometry0}), the commute time can be expressed as
\begin{equation}
\kappa(\bx_i,\bx_j)=\sum_{k=2}^N \frac{1}{1 -\lambda_k} 
\left (\frac{\bfi_k(i)}{\sqrt{\pi_i}} -\frac{\bfi_k(j)}{\sqrt{\pi_j}} \right)^2,
\label{isometry}
\end{equation}
where $\pi_i = d_i/\sum_{i,j}W_{i,j}$ is the stationary
distribution associated with $\mathbf P$, ${\mathbf \pi}^T {\mathbf P} =
{\mathbf \pi}^T$. The right hand side of (\ref{isometry}) is the sum of $N-1$ 
 squared contributions of the form 
$(\bfi_k(i)/\sqrt{\pi_i} -\bfi_k(j)/\sqrt{\pi_j} )/\sqrt{1 -\lambda_k}$.
Each contribution is the difference between two terms: $\bfi_k(i)/\sqrt{\pi_i}$ and
$\bfi_k(j)/\sqrt{\pi_j}$, which are associated with nodes $\bx_i$ and
$\bx_j$,  respectively. 
%
\begin{figure}[H]
\makebox{\SF Algorithm 1: Construction of the embedding}
  \barre
  \begin{itemize}
  \item [] {\SF Input}: 
    \begin{itemize}
    \item $\bx_i(t), t=0,\cdots,T-1$, $i=1,\cdots,N$, 
    \item $\sigma$; $n_n$; $K$: dimension of the embedding,
    \end{itemize}
  \item []{\SF Algorithm:}
    \begin{enumerate}
    \item construct the graph defined by the $n_n$ nearest neighbors of each $\bx_i$
    \item compute $\mathbf W$ and $\mathbf D$. Find the first $K$ eigenvectors, $\bfi_k$, of ${\mathbf
        D}^{-\frac{1}{2}}{\mathbf W}{\mathbf D}^{-\frac{1}{2}}$ 
    \end{enumerate}
  \item  [] {\SF Output:}  $K$ coordinates of each $\bx_i$, 
    $\frac{1}{\sqrt{\pi_i}} \left\{\bfi_k(i)/\sqrt{1-\lambda_k}
      \right\}, k=2,\cdots,K+1$   
  \end{itemize}
  \barre
  \caption{Construction of the embedding
    \label{algo1}}
\end{figure}

\subsection{Embedding of the dataset
\label{algorithm}}
We define a mapping from $\bx_i$ to a vector of size $N-1$,
\begin{equation}
\bx_i \mapsto  
 \frac{1}{\sqrt{\pi_i}}
\left [
\frac{\bfi_2(i)}{\sqrt{1-\lambda_2}}, \cdots \cdots,
\frac{\bfi_N(i)}{\sqrt{1-\lambda_N}} \right ]^T.
\label{embed0}
\end{equation}
The idea was first proposed in \citep{Berard94} to embed
manifolds. Recently, the same idea has been revisited in the machine
learning literature \citep{Belkin03,Coifman06a}. According to
(\ref{isometry}), the commute time is simply the Euclidean distance
measured between the new coordinates.  In practice, we need not use all the
eigenvectors in (\ref{embed0}). Indeed, because $-1 \leq \lambda_N \cdots
\leq \lambda_2 < \lambda_1 =1$ we have
$1/\sqrt{1-\lambda_2} > 1/\sqrt{1-\lambda_3} > \cdots
1/\sqrt{1-\lambda_N}$.
We can therefore neglect $\bfi_k(j)/\sqrt{1-\lambda_k}$ in (\ref{embed0})
for large $k$, and reduce the dimensionality of the embedding by using only
the first $K$ coordinates. Finally, we define the map $\Psi$ from $\R^T$ to $\R^K$,
\begin{equation}
 \bx_i  \mapsto  \Psi(\bx_i) = 
 \frac{1}{\sqrt{\pi_i}}
\left [
\frac{\bfi_2(i)}{\sqrt{1-\lambda_2}}, \cdots \cdots,
\frac{\bfi_{K+1}(i)}{\sqrt{1-\lambda_{K+1}}} \right ]^T.
\label{embed}
\end{equation}
The low dimensional parametrization (\ref{embed}) provides a good
approximation when the spectral gap $\lambda_1 -\lambda_2 = 1
-\lambda_2$, is large.  The construction of the embedding is
summarized in Fig. \ref{algo1}.  Unlike PCA which gives a set of
vectors on which to project the dataset, the embedding (\ref{embed})
directly yields the new coordinates for each time series $\bx_i$. The
new coordinates of $\bx_i$ are given by concatenating the
$i^\text{th}$ coordinates of $\bfi_k$, $k=2,\cdots,K+1$. The
first eigenvector $\bfi_0$ is constant and is not used.
{\em What is the connection to PCA ?}\\
Because each $\bfi_k$ is also an eigenvector of the graph Laplacian,
$\mathbf L = \mathbf I - {\mathbf D}^{\frac{1}{2}}{\mathbf P}{\mathbf
  D}^{-\frac{1}{2}}$, it minimizes the ``distortion'' induced by $\bfi$ and
measured by the Rayleigh ratio \citep{Belkin03},
\begin{equation}
\min_{\bfi, \|\bfi\|=1}
\frac{\sum_{{i,j}} W_{i,j}(\bfi(i) - \bfi (j) )^2}{\sum_{i} d_i\bfi^2(i)},
\label{rayleigh}
\end{equation}
where $\bfi_k$ is orthogonal to the previous eigenvector $\{\bfi_0,
\bfi_1,\cdots,\bfi_{k-1}\}$.  The gradient of $\bfi$ at the vertex
$\bx_i$ on the graph can be computed as a linear combination of terms
of the form $(\bfi(i) - \bfi (j))$, where $j$ and $i$ are
connected. Therefore the numerator of the Rayleigh ratio
(\ref{rayleigh}) is a weighted sum of the gradients of $\bfi$ at all
nodes of the graph, and quantifies the average local distortion
created by the map $\bfi$.  A function that minimizes
(\ref{rayleigh}), will still introduce some global distortions. Only
an isometry will preserve all distances between the time series, and
the isometry which is optimal for dimension reduction is given by
PCA. However, as shown in the experiments, the first PCA components
are usually not able to capture the nonlinear structure formed by the
set of time series. As a result, PCA fails to reveal the organization
of the dataset in terms of low dimensional activated time series and
background time series.  \cite{Thirion03c} have used kernel PCA to
analyze the distributions of the coefficients of a model fitted to
fMRI time series. A block design fMRI dataset from a macaque monkey is
studied. By visual inspection, the authors show that their method can
organize the coefficients according to the relative strength of the
activation. The eigenvectors of the Laplacian have also been used to
construct maps of spectral coherence of fMRI data in
\citep{Thirion06}.

{\em How many new coordinates do we need ?}\\
We can estimate $K$, the number of coordinates in the embedding
(\ref{embed}), by calculating the number of $\bfi_k$ needed to
reconstruct the low dimensional structures present in the dataset. As
opposed to PCA, the embedding defined by (\ref{embed}) is not designed
to minimize the reconstruction error, it only minimizes the average
local distortion  (\ref{rayleigh}). Nevertheless, we can
take advantage of the fact that the eigenvectors $\{\bfi_k\}$
constitute an orthonormal basis for the set of functions defined on
the vertices of the graph \citep{Chung97}. In particular, we make the
trivial observation that the scan at time $t$, $\bx(t) =
[x_1(t),\cdots,x_N(t)]^T$, is a function defined on the nodes of the
graph: $\bx(t)$ at node $\bx_i$ is the value of the fMRI signal
at voxel $i$, $x_i(t)$. We can therefore expand $\bx(t)$ using the
$\bfi_k$,
\begin{equation*}
\begin{split}
\bx(t) & = 
\sum_{k=1}^{K} <\bx(t), \bfi_k> \bfi_k\\
 &+\sum_{k=K+1}^{N} <\bx(t), \bfi_k> \bfi_k\\
& = \hat{\bx}^K(t) + \br (t),
\end{split}
\end{equation*}
where $\hat{\bx}^K(t)= \sum_{k=1}^{K} <\bx(t), \bfi_k> \bfi_k$ is the
approximation to the $t^\text{th}$ scan using the first $K$ eigenvectors,
and $\br(t)$ is the residual error.  We can compute a similar approximation
for all the scans ($t=1, \cdots, T$), and compute the temporal average of
the relative approximation error at a given voxel $i$,
\begin{equation}
\varepsilon_i(K) = \frac {\sum_{t=1}^T (x_i(t)-\hat{x_i}^K(t))^2}
{\sum_{t=1}^T x^2_i(t)}.
\label{residue}
\end{equation}
Finally, one can compute the average of $\varepsilon_i (K)$ over all a
group of voxels in the same functional area $\cal R$,
$\varepsilon_{\cal R}(K) = (\sum_{i \in {\cal R}} \varepsilon_i
(K))/|{\cal R}|$.  We expect $\varepsilon_{\cal R}(K)$ to decay fast
with $K$ if the time series within $\cal R$ are well approximated by
$\bfi_1,\cdots,\bfi_K$. In practice, for each region $\cal R$ we find
$K_{\cal R}$ after which $\varepsilon_{\cal R}$ stops having a fast
decay. We then choose $K$ the number of coordinates to be the largest
$K_{\cal R}$ among all regions $\cal R$.  Examples are shown in
Fig. \ref{checker_embedding}-right, where $K$ is chosen at the
``knee'' of the curves $\varepsilon_{\cal R}(K)$.
\subsection{$k$-means clustering
\label{cluster}}
\begin{figure}[H]
\centerline{
  \includegraphics[width=12pc]{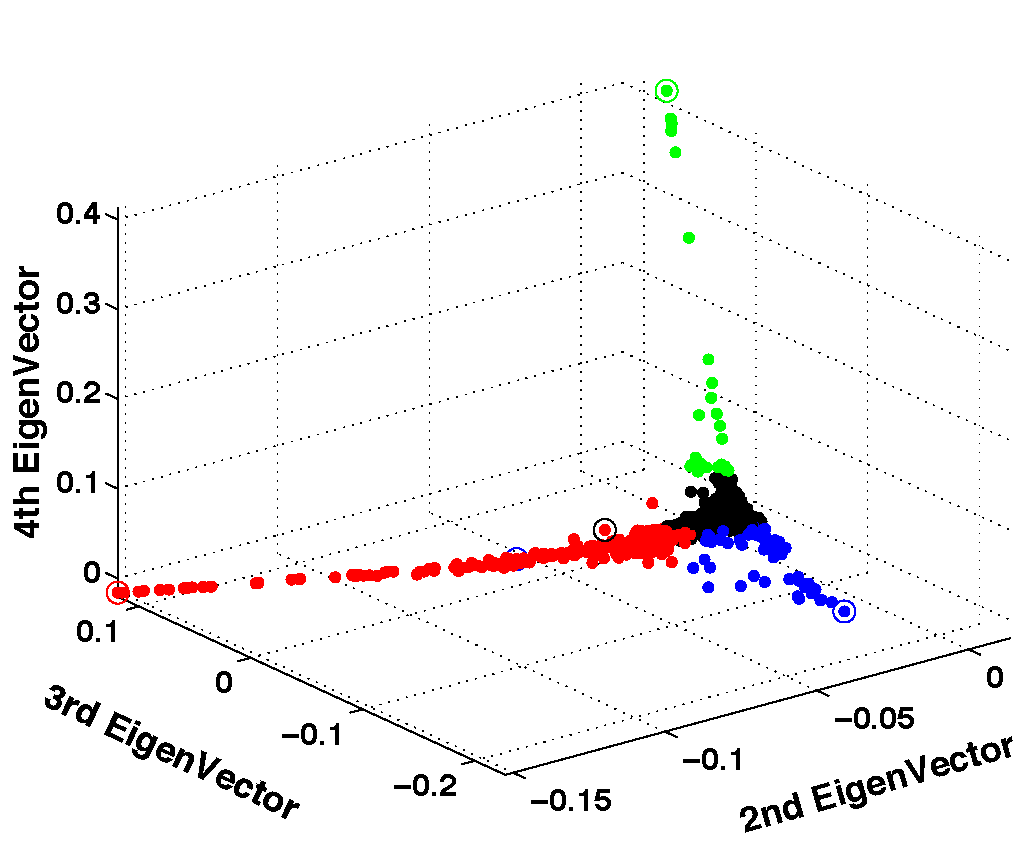}
}
\centerline{
  \includegraphics[width=8pc]{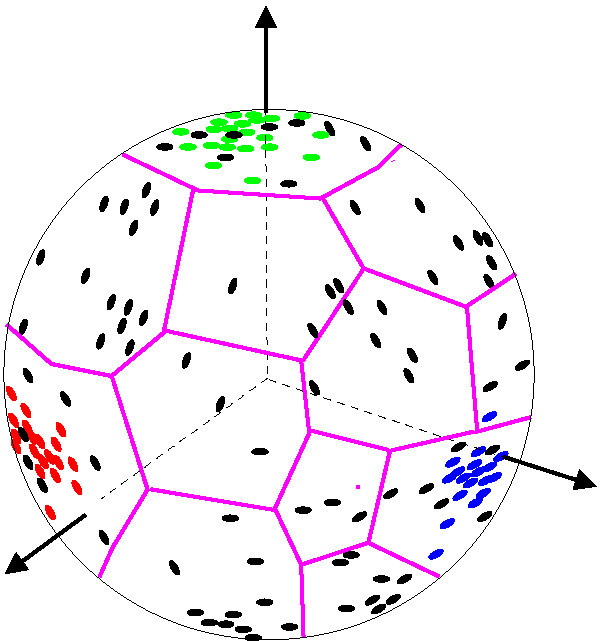}
}
\caption{Left: low dimensional embedding of the dataset analyzed in section
  \ref{blockdesign}. Each dot $i$ is $\Psi(\bx_i)$, the image of the time
  series $\bx_i$ through the mapping $\Psi$. Right: the $\Psi(\bx_i)$ are
  projected on the sphere, and the projections are clustered on the sphere.
  \label{clusterover}}
\end{figure}

The map $\Psi$ defined by (\ref{embed}) provides a set of new
coordinates, $\Psi(\bx_i)$, for each $\bx_i$. We cluster the set
$\left\{\Psi(\bx_i), i = 1,\cdots, N\right\}$ into a small number of
coherent structures and a large background component. We use a
variation of the $k$-means clustering algorithm for this task. The low
dimensional parameterization of the dataset usually has a ``star''
shape (Fig. \ref{clusterover}-left), where the out-stretching ``arms''
of the star are related to activated time series or strong
physiological artifacts, and the center blob corresponds to background
activity. Our goal is to segment each of the ``arms'' from the center
blob. We project all the $\Psi(\bx_i)$ on a unit sphere centered
around the origin (Fig. \ref{clusterover}-right).  We then cluster the
projections on the sphere: the distance between two points on the
sphere is measured by their angle. The center component (shown in
black in Fig. \ref{clusterover}-left) is usually spread all over the
sphere, and is mixed with the branches (Fig. \ref{clusterover}-right).
The time series from the background component can be separated from the other time series by measuring the
distance of $\Psi(\bx_i)$ to the origin
(Fig. \ref{clusterover}-left). The number of clusters%
\noindent  can be chosen to
be equal to $K+1$. This choice is based on experience that each
eigenvector (each dimension) will contribute to an independent arm,
and the background time series will contribute to the last cluster.
This choice may over-segment the dataset. This is usually obvious from
the visual inspection scatter plot, and the corresponding spatial
maps. We iteratively refine the estimate of the number of clusters, by
merging small clusters at each iteration.
\section{Results
\label{results}}
In this section we describe the results of experiments conducted on
synthetic and {\em in-vivo} datasets. We construct the embedding
according to the algorithm described in Fig~\ref{algo1}, and the
clustering algorithm, described in section \ref{cluster}, divides the
embedded dataset into coherent groups. We interpret the coherent
structures in terms of a task-related hemodynamic response, and
physiological artifacts. Voxels that correspond to task-related
activation are identified and activation maps are generated
accordingly. We evaluate our approach using two different criteria.
First, we compare the parameterization created by our approach with
the parametrization produced by PCA. The comparison is based on%
\begin{figure}[H]
\centerline{
  \includegraphics[width=9pc]{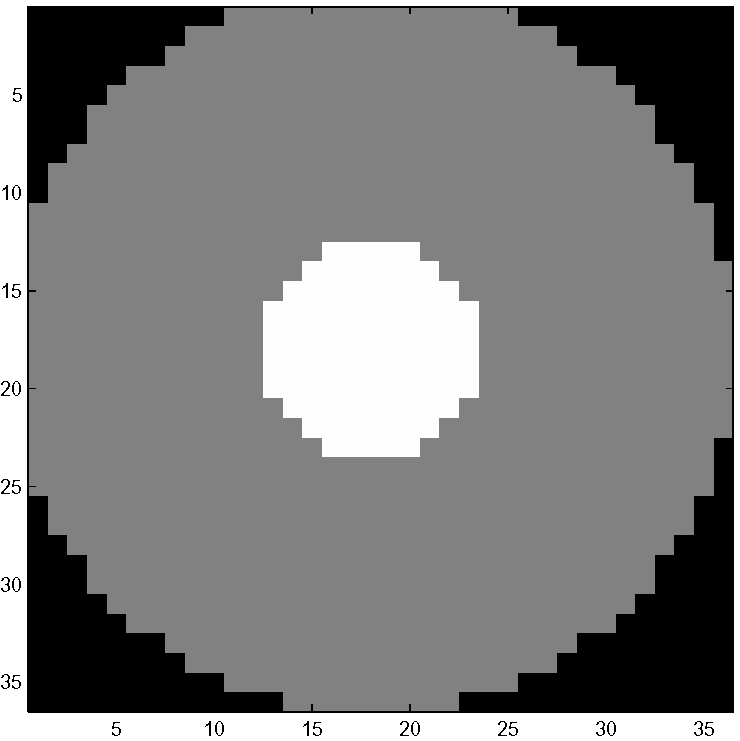}
  \includegraphics[width=9pc]{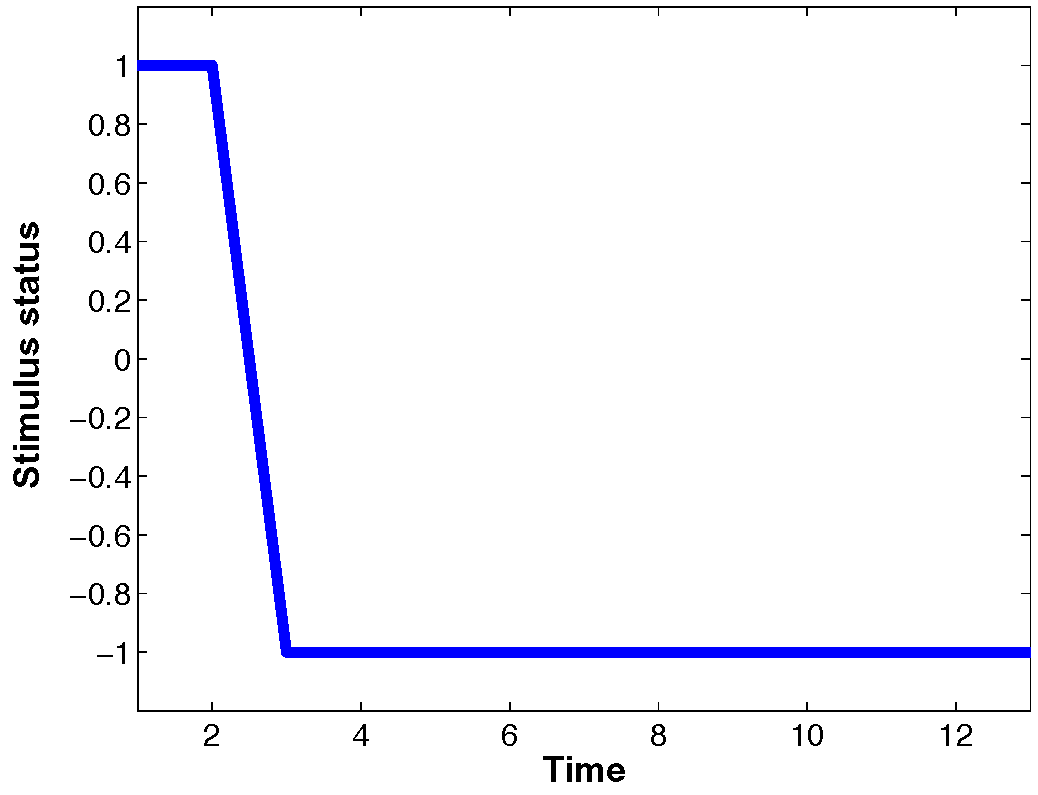}
}
\caption{Left: synthetic brain: activation (white), non-activation (gray), outside the
  brain (black). Right: stimulus time series.
\label{ground_truth}}
\end{figure}
\noindent  our
ability to identify and extract well defined structures from the new
parameterization. Our second criterion is the comparison of the
activation maps obtained with our approach with the ones generated by
the General Linear Model (GLM).  Five datasets are selected for the
analysis: a synthetic dataset, a block design dataset, an
event-related dataset, and two datasets from the EBC competition
\cite{EBC}.
\subsection{Synthetic data}
\begin{figure}[H]
\centerline{
  \includegraphics[width=12pc]{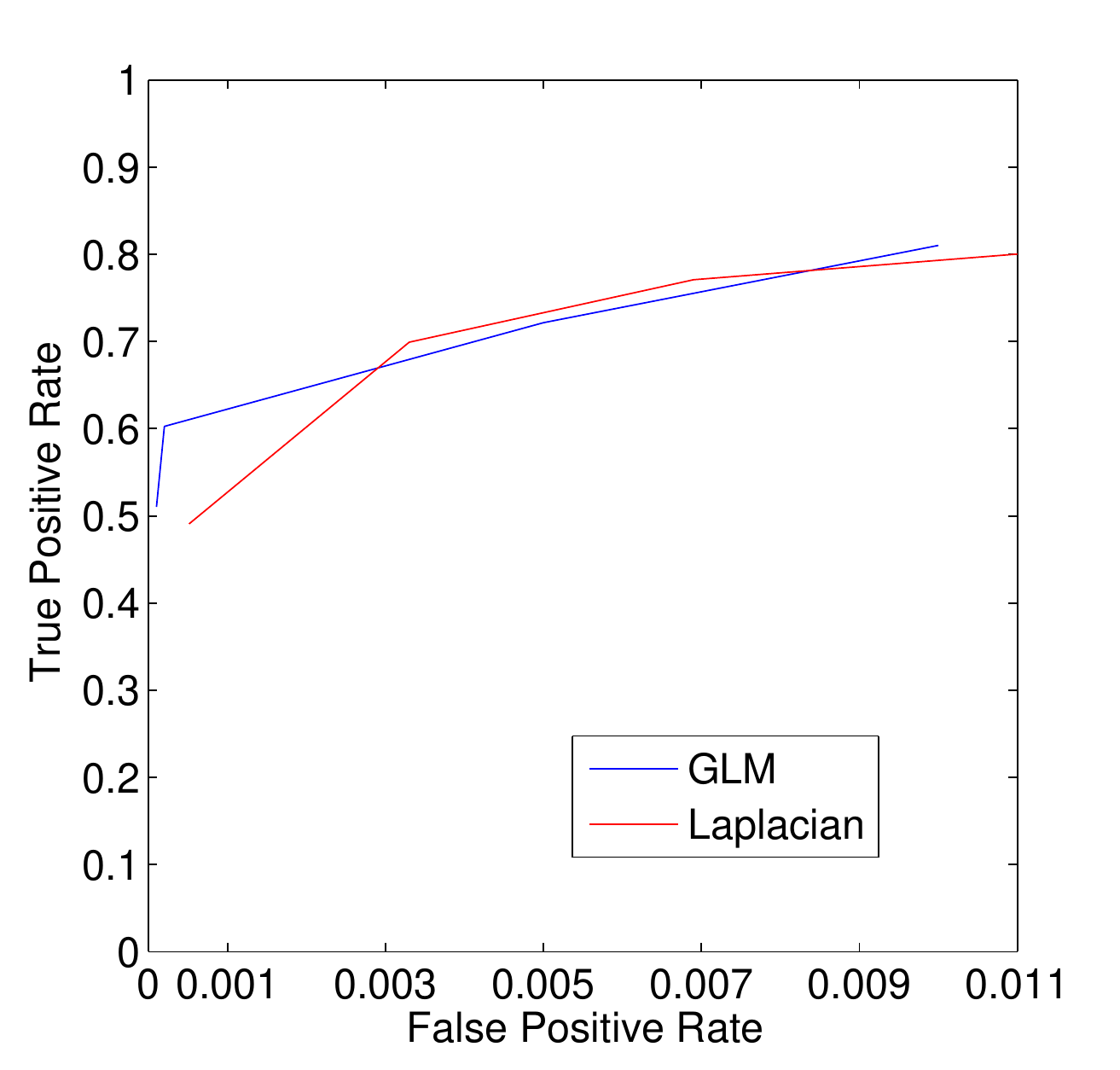}
}
\caption{ROC curves: comparison of our approach to the  GLM.
\label{roccurves}}
\end{figure}

The synthetic datasets were designed by blending activation into
background, non activated, time-series that were extracted from a real {\em
  in vivo} dataset (described in section \ref{blockdesign}). We discarded
time series exhibiting large variance. Activated time-series were
constructed by adding an activation pattern $f(t)$ to the background time series. The
activation pattern $f(t)$ was obtained by convolving the canonical
hemodynamic filter $h(t)$ used in SPM \citep{Friston05}
\begin{equation}
  h(t) = \alpha
  \left(
    \frac{t}{d_1}
  \right)^{a_1} 
  e^{- (t-d_1)/b_1} 
  - c \left( \frac{t}{d_2}\right)^{a_2} 
  e^{- (t-d_2)/ b_2},
  \label{hrf}
\end{equation}
with the stimulus time-series $g(t)$ (Fig. \ref{ground_truth}-left).  The two parameters $\alpha$ and $b_1$
were randomly distributed according to two uniform distributions, on
$[0.8,1.2]$ and on $[5,10]$ respectively. The other parameters were
fixed and chosen as follows, $a_1= 6, a_2=12,b_2=0.9,$ and $c=0.35$.
By varying $b_1$ and $\alpha$ independently, we generated a family of
hemodynamic responses with different peak dispersions and scales.  We generated $20$ independent
realizations according to this design. Each dataset consisted of a
white disk of activated voxels inside a circular gray brain of
background voxels (Fig. \ref{ground_truth}-left). Black voxels were in
the air. There were altogether $1067$ voxels inside the circular
brain, with $97$ activated%
\begin{figure}[H]
\centerline{
  \includegraphics[width=9pc]{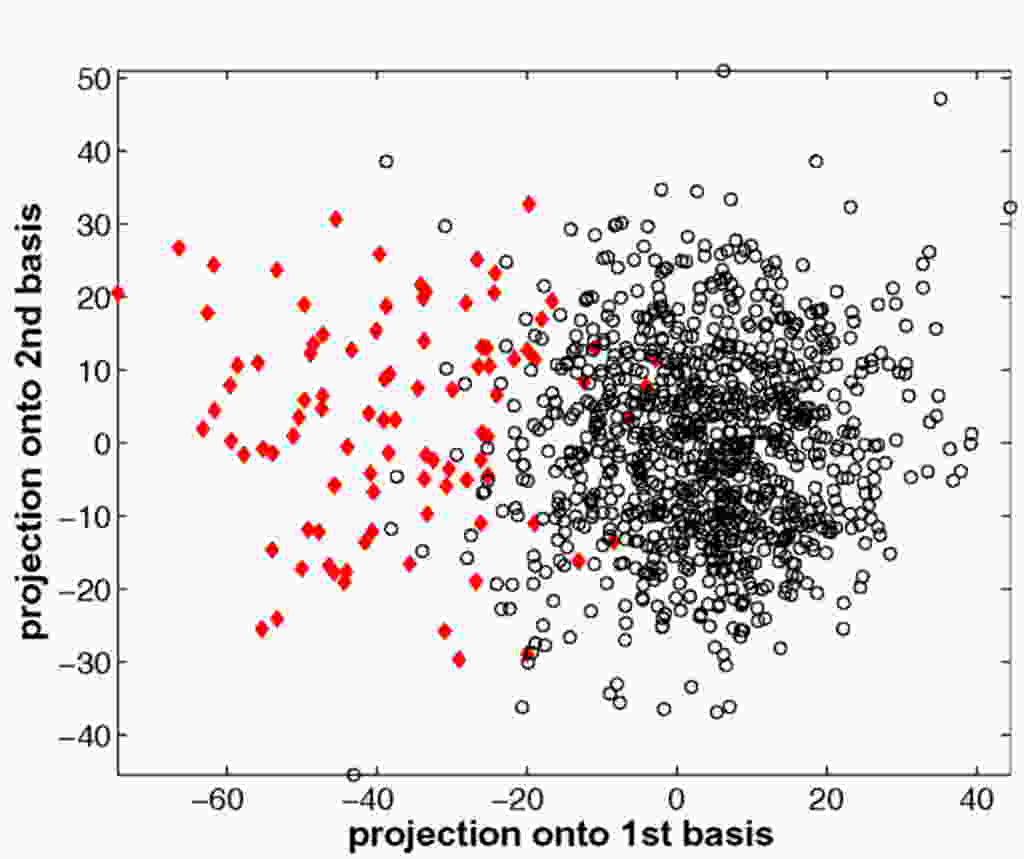}
  \includegraphics[width=9pc]{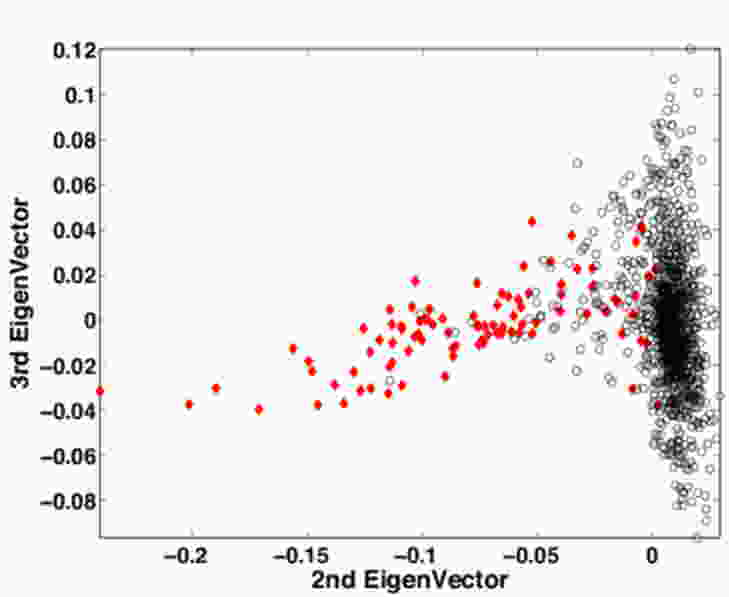}
}
\caption{Activated  (red) and background (black) time series 
projected on the first two PCA components (left), and parametrized by
$\Psi$ (right).
\label{syn_embedding}}
\end{figure}
\begin{figure}[H]
\centerline{
  \includegraphics[width=9pc]{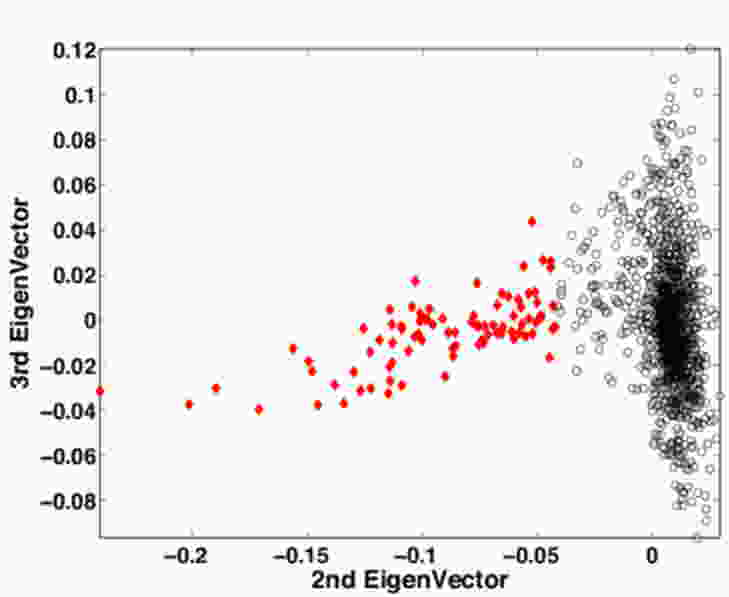}
  \includegraphics[width=9pc]{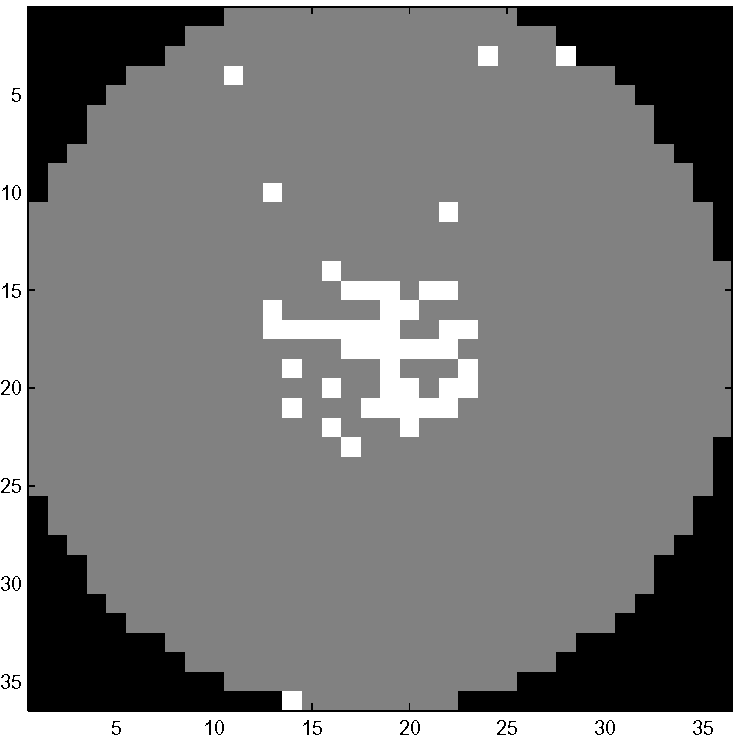}
}
\caption{Left: clustering of $\left\{ \Psi(\bx_i),  i =1, \cdots,N \right
  \}$ into $2$ clusters: activated (red) and  background (black). Right:
  activated  (white) and background (gray) pixels. 
  \label{syn_kmean_2clusters}}
\end{figure}
\noindent voxels, ($9\%$ activation).
Fig. \ref{syn_embedding}-left shows the projections on the first two
principal axes of the $1067$ time series of one realization. The
projections are color-coded according to their status: activated (red
diamond) and background (black circle).  The parameterization given by
our approach is shown in \ref{syn_embedding}-right. We used only $K=2$
coordinates in (\ref{embed}) for $\Psi$. Activated time series are
distributed along a thin strip that extends outward from the main
cluster. This low dimensional structure is compact and easy to
identify.  In comparison, the two dimensional representation given by
PCA (Fig. \ref{syn_embedding}-left) is less conspicuous: activated
time series (red) overlap with background time series (black).  After
embedding the dataset into two dimensions, the dataset is partitioned
into two clusters. Fig.  \ref{syn_kmean_2clusters}-left shows the
result of clustering: the labels (red for activated, black for
background) are based on the clustering only.  The corresponding
activation map is shown in Fig. \ref{syn_kmean_2clusters}-right. We
compared our algorithm with a linear model equipped with the perfect
knowledge of the hemodynamic response $h(t)$ (with $b_1=1$ and
$\alpha=1$). A Student $t$-test was applied to the regression
coefficient to test its significance, and voxels with a $p$-value
smaller than a threshold were considered activated.
Fig. \ref{roccurves} provides a quantitative comparison%
\noindent  of our
approach with the linear model using a receiver operator
characteristic (ROC) curve. The true activation rate (one minus the
type II error) is plotted against the type I error (false alarm
rate). The ROC curve was computed over 20 trials. Each trial included
different activation strengths $\alpha$. In this experiment, the
linear model has access to an oracle, in the form of the perfect
knowledge of the hemodynamic response $h(t)$, and should therefore
perform very well. In fact, if the noise added to $f(t)$ were to be
white, we know from the matched filter theorem that the linear model
would provide the optimal detector. Here,the noise is extracted from
the data, and is probably not white \citep{Bullmore01}.  As shown in
Fig. \ref{roccurves}, our approach performs as well as the GLM for a
type I error in the range [0.003,0.009]. At low type I error, our
approach misses activations.
\subsection{In vivo data I: block design dataset
\label{blockdesign}}
\begin{figure}[H]
\centerline{
  \includegraphics[width=9pc]{fig3a.png}
  \includegraphics[width=9pc]{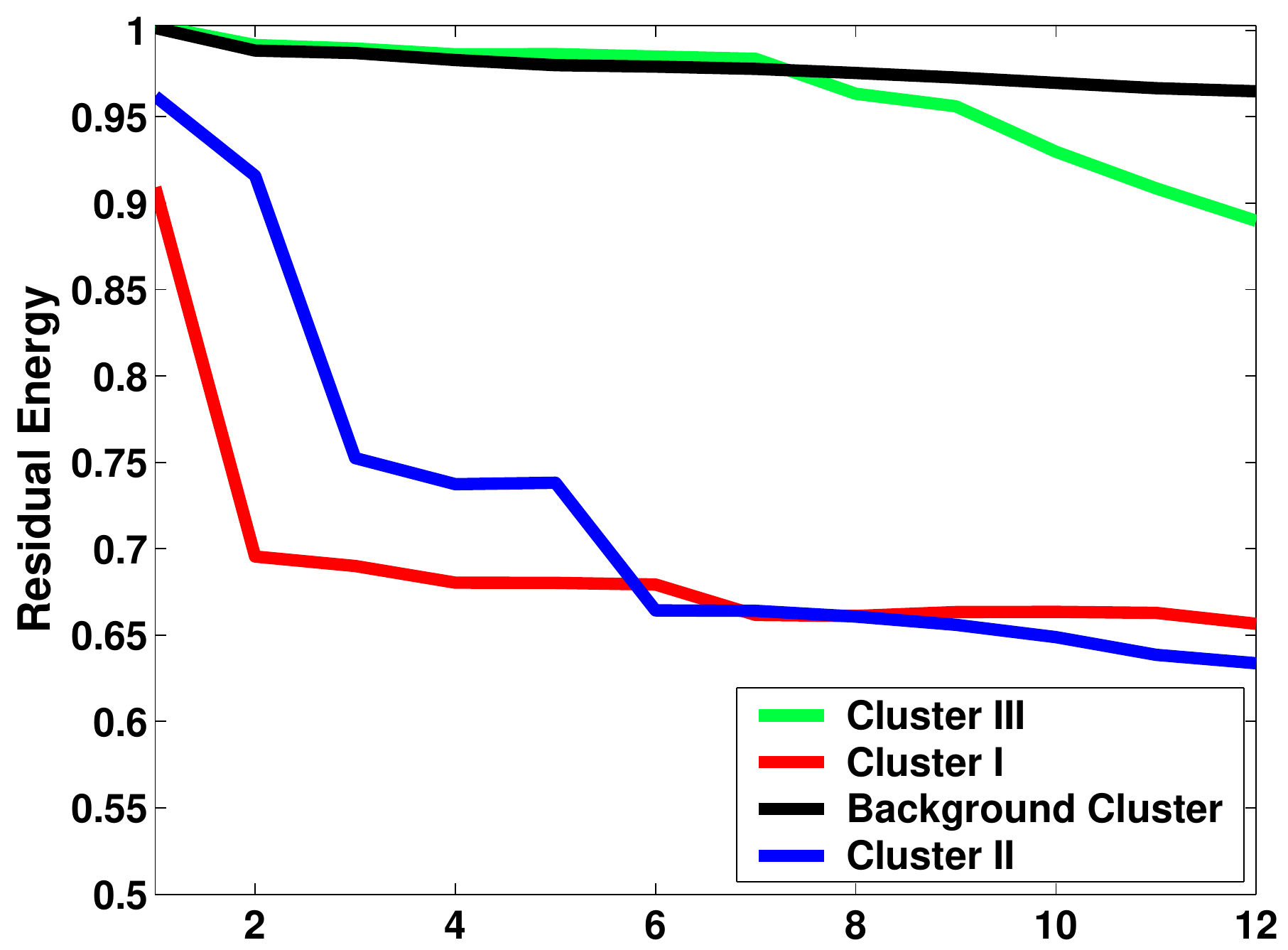}
}
\caption{Three-dimensional embedding: $\left \{\Psi (\bx_i),  i =1,\cdots, 3084 \right
  \}$:  cluster~I (red), cluster~II (blue), cluster~III  (green), and
  background (black). Time series marked by a   circle are shown in
  Fig. \ref{checker_kmeans_ts}. Right: residual error $\varepsilon_{\cal R}
  (K)$ as a  function of the number of coordinates $K$. 
  \label{checker_embedding}}
\end{figure}

\begin{figure}[H]
\centerline{
  \includegraphics[angle=180,width=9pc]{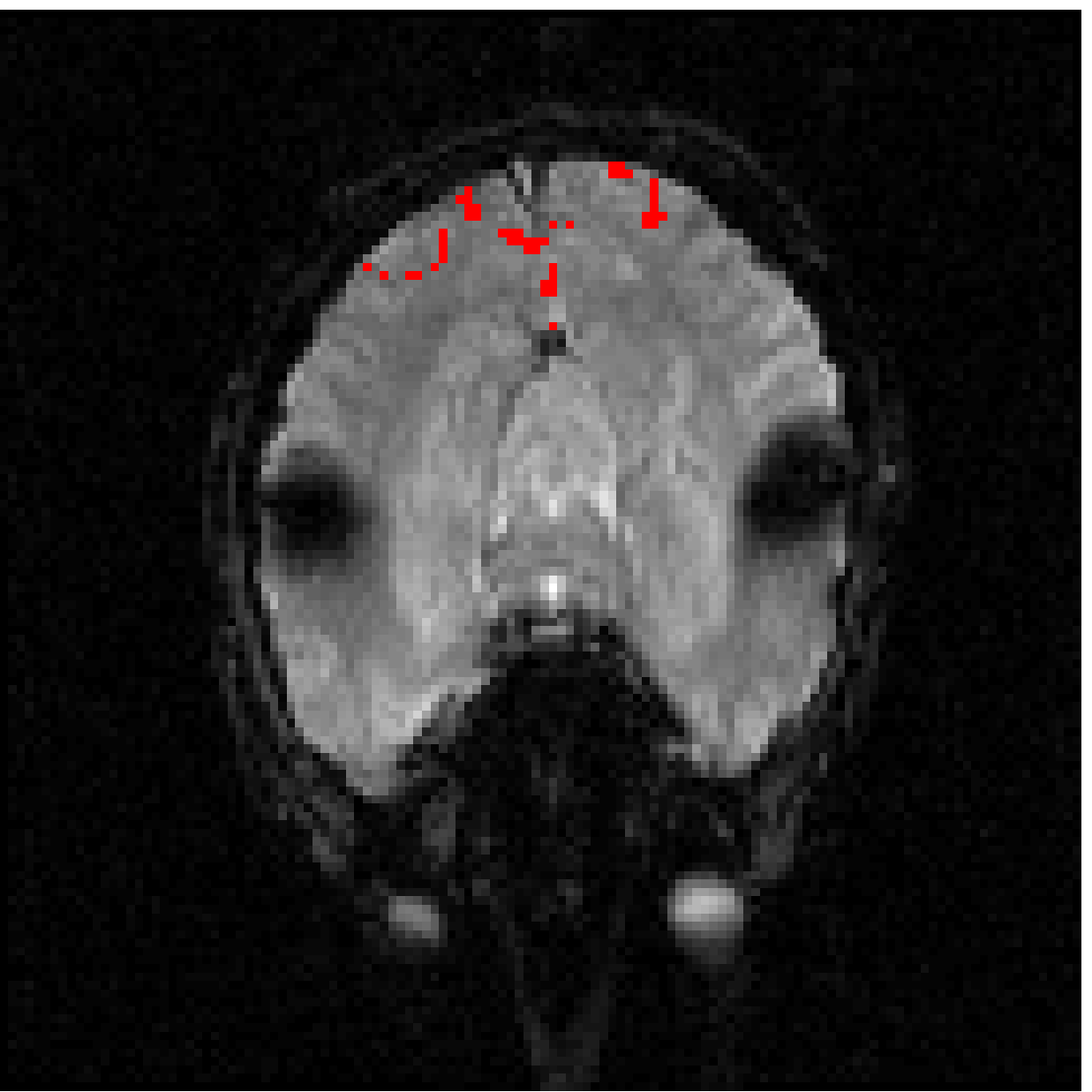}
  \includegraphics[angle=180,width=9pc]{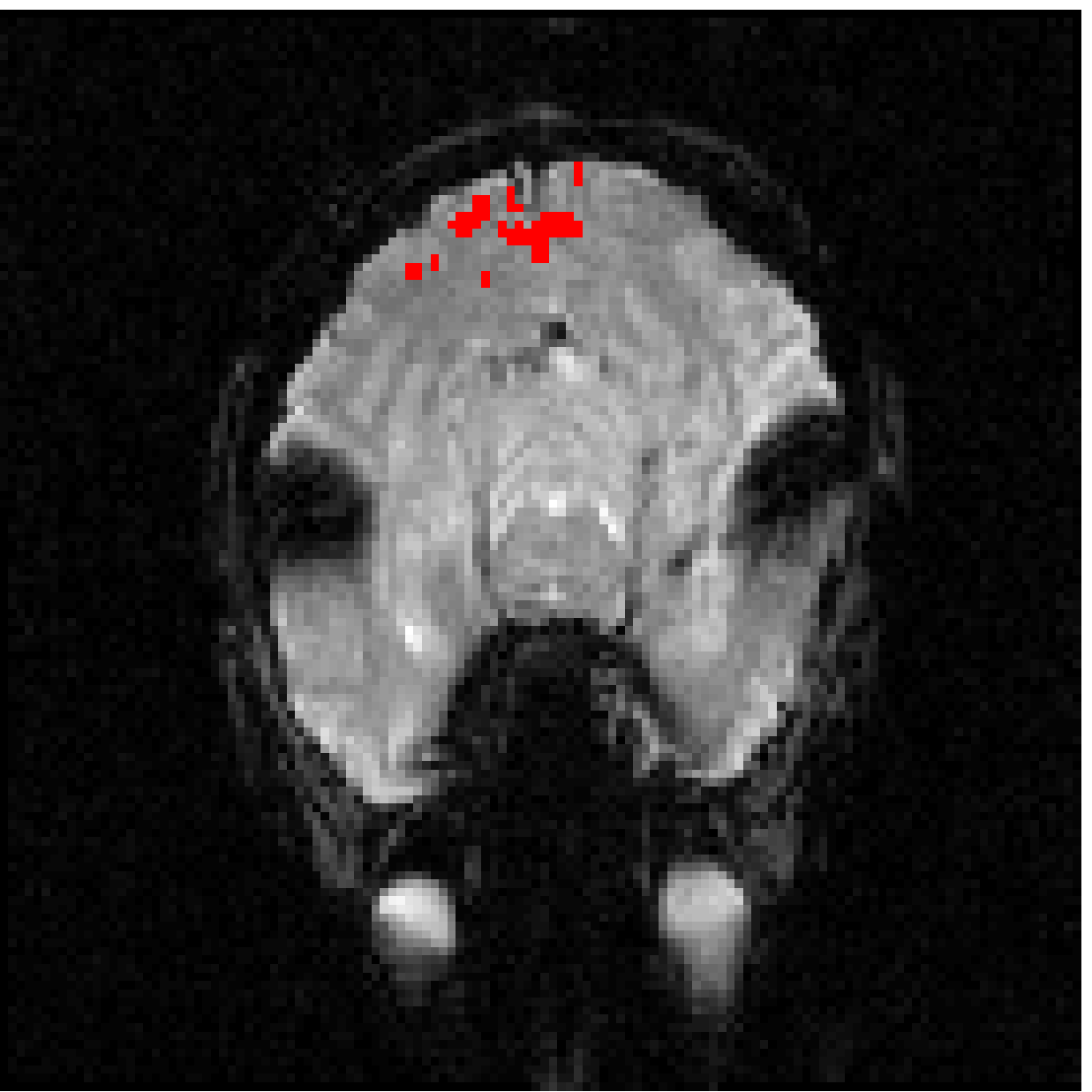}
}
\centerline{
  \includegraphics[angle=180,width=9pc]{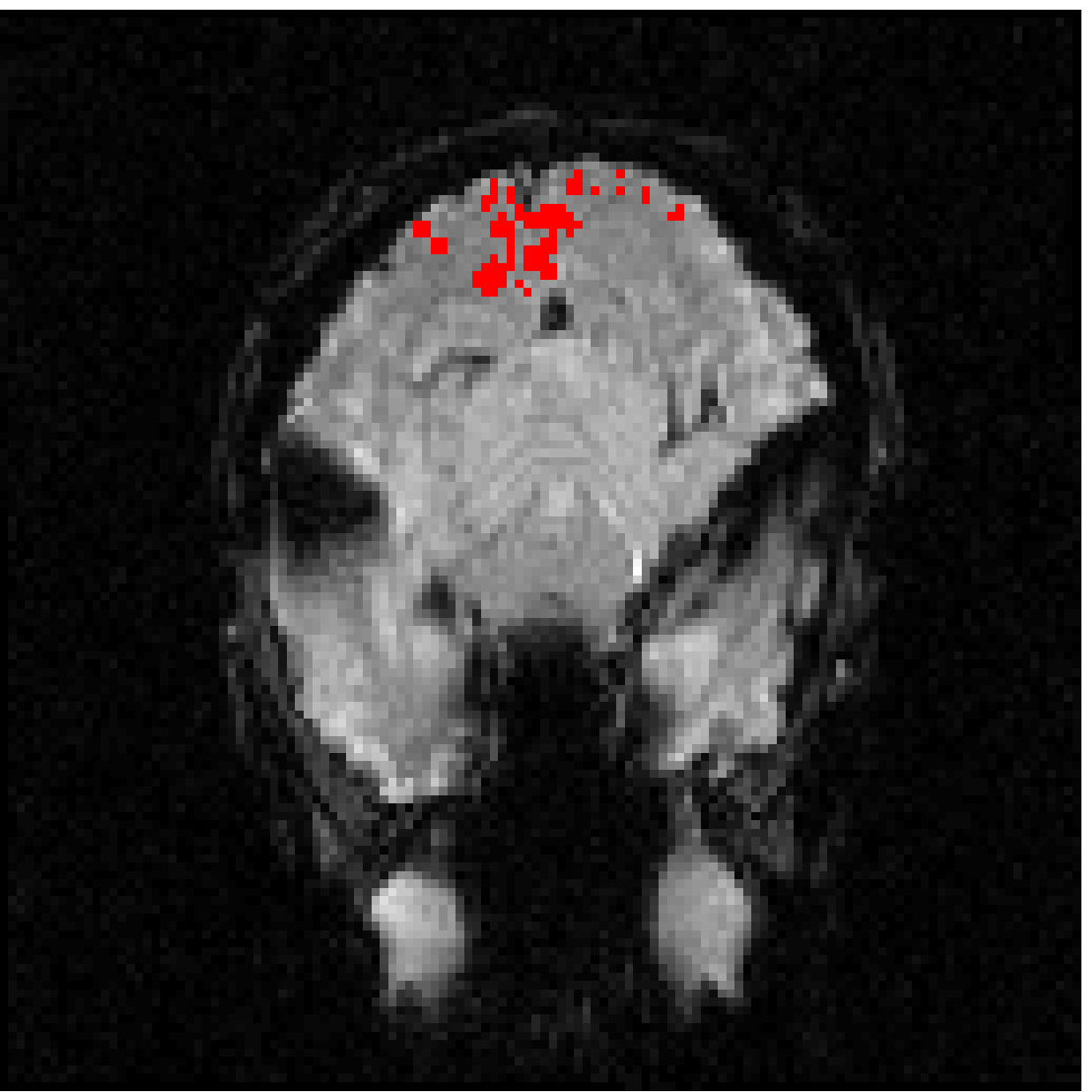}
  \includegraphics[angle=180,width=9pc]{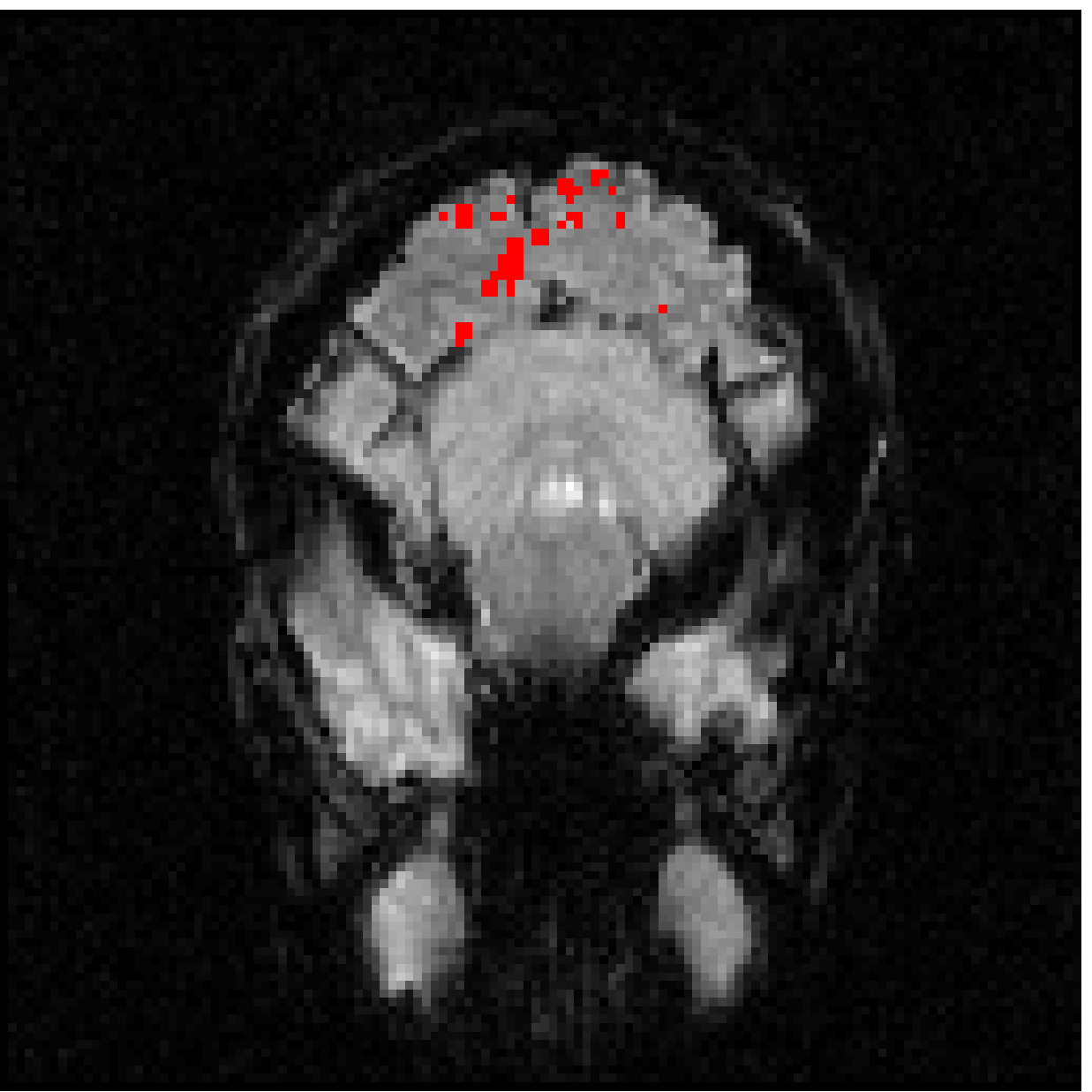}
}
\caption{Top two rows:  voxels  are color-coded according to the cluster
  color (except for the background cluster). We interpret  cluster~I (red)
  as voxels in the visual cortex recruited by the stimulus. 
Bottom two rows: activation maps obtained using the linear regression model ($p=0.001$). 
\label{checker_map}}
\end{figure}

We apply our technique to a block design dataset that demonstrates
activation of the visual cortex \citep{Tanabe01a}.  A flashing
checkerboard image was presented to a subject for 30 seconds, and a
blank image was presented for the next 30 seconds. This alternating
cycle was repeated four times.  Images were acquired with a $1.5$T
Siemens MAGNETOM Vision equipped with a standard quadrature head coil
and an echoplanar subsystem (TR = $3$s, xy dimension: $128\times 128$,
voxel size = $1.88\times 1.88\times 3$mm, 12 contiguous
slices). Eighty images were obtained. We analyze a volume that
contains the calcicarin cortex (Brodman areas 17)
(Fig. \ref{checker_map}). There are altogether $3084$ intracranial
time series in the volume. The linear trend of each time series is
removed. Fig. \ref{checker_embedding}-left displays the image of the
time series by the%
\begin{figure}[H]
\centerline{
  \includegraphics[width=9pc]{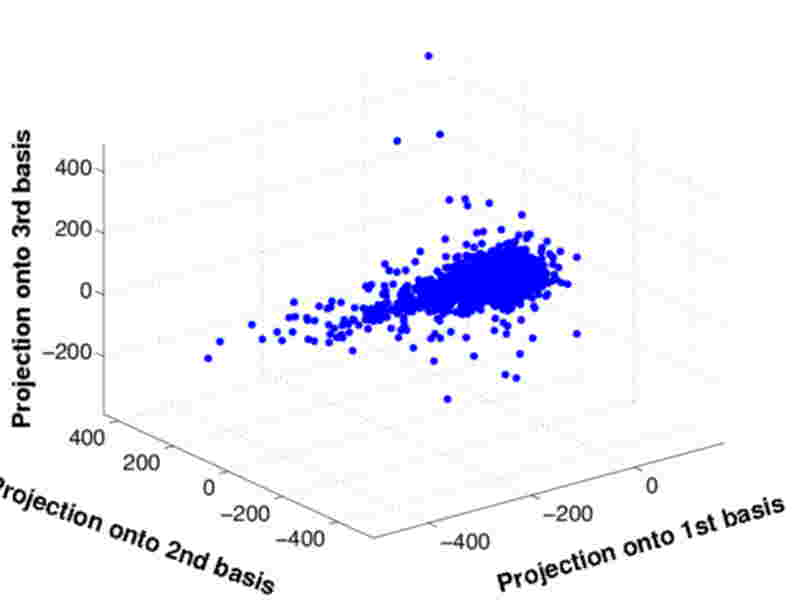}
  \includegraphics[width=9pc]{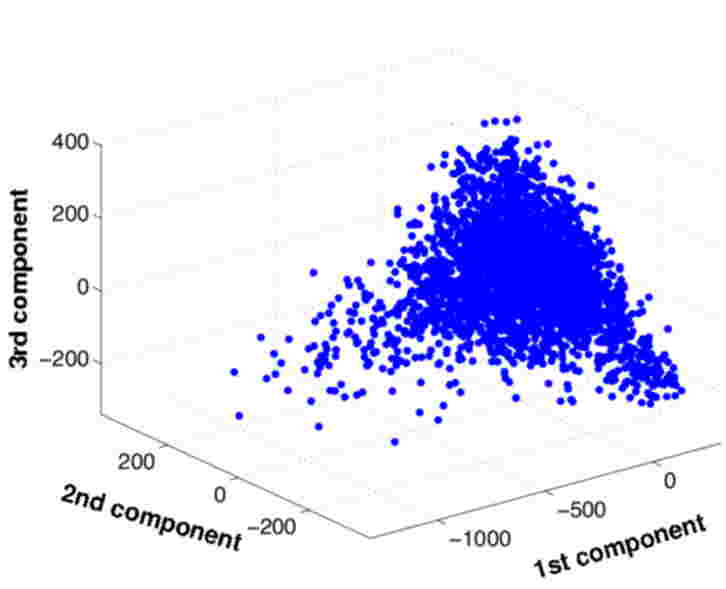}
}
\caption{Low dimensional embedding obtained by PCA (left), and ISOMAP (right).
\label{checker_pca}}
\end{figure}
\begin{figure}[htbp]
\centerline{
\raisebox{6pc}{\hspace*{3pc}\SF A} 
\hspace*{-3pc}\includegraphics[width=9pc]{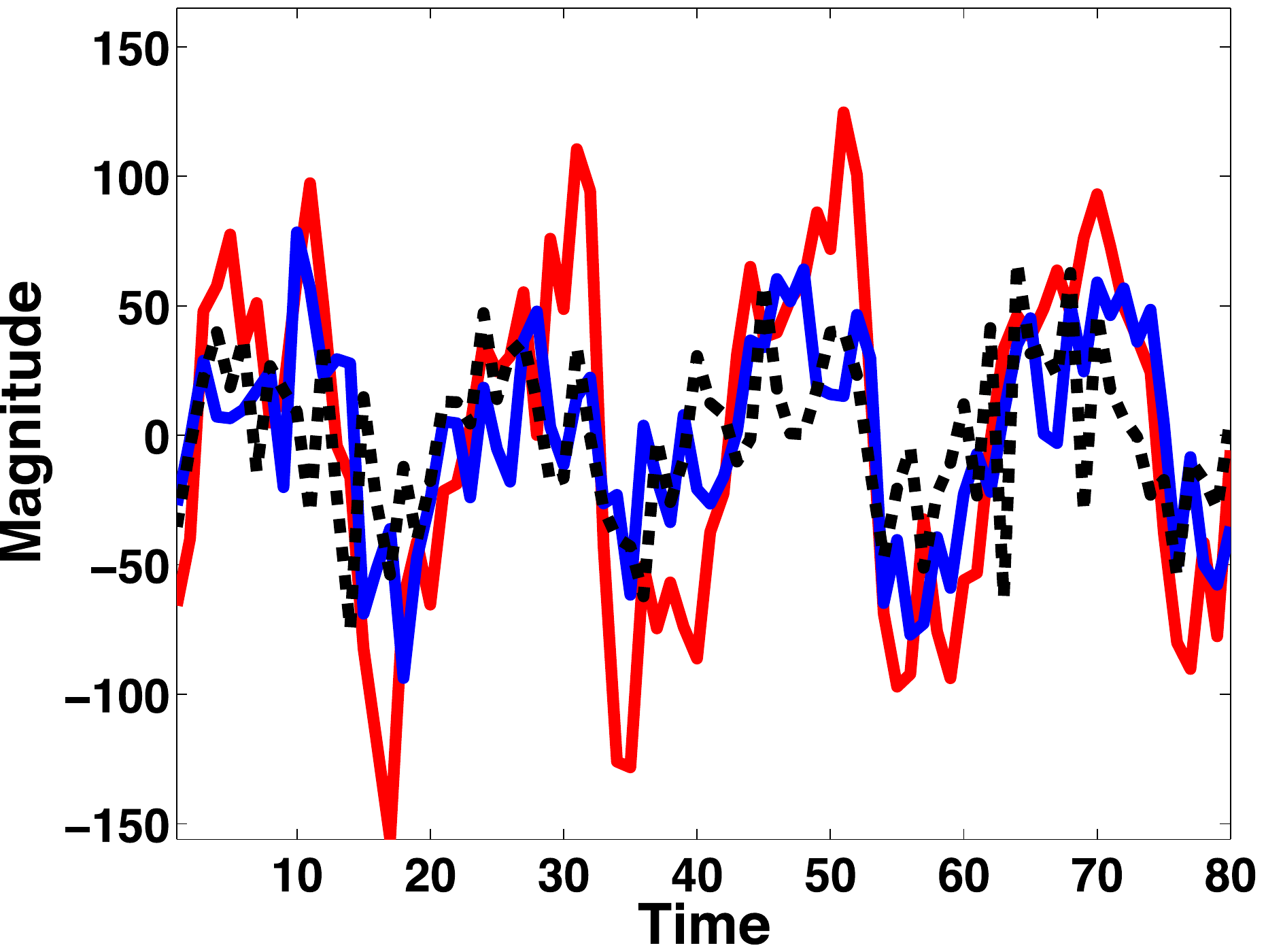}
\raisebox{6pc}{\hspace*{3pc}\SF B} 
\hspace*{-3pc}\includegraphics[width=9pc]{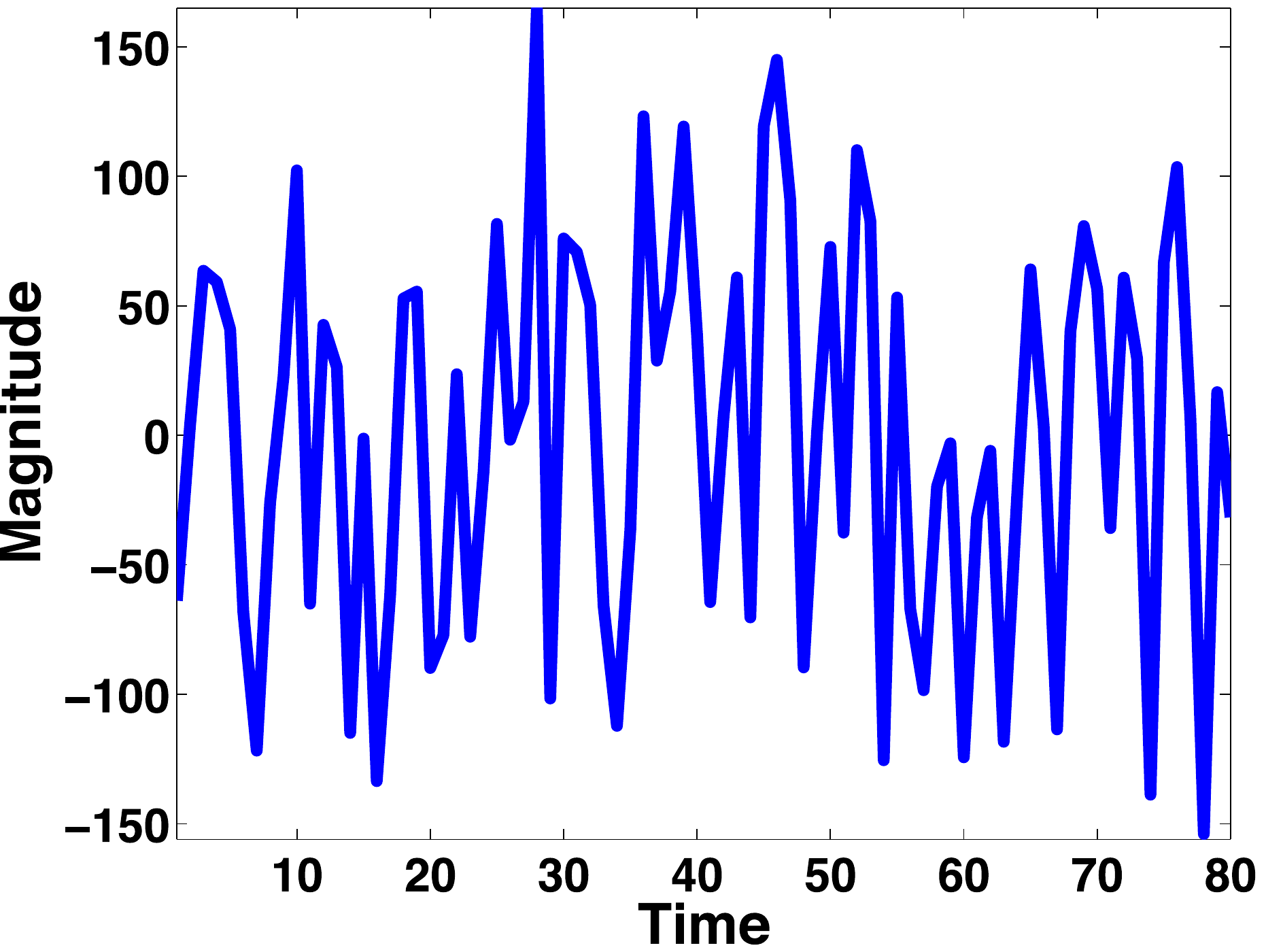}
}
\centerline{
\raisebox{6pc}{\hspace*{3pc}\SF C} 
\hspace*{-3pc}\includegraphics[width=9pc]{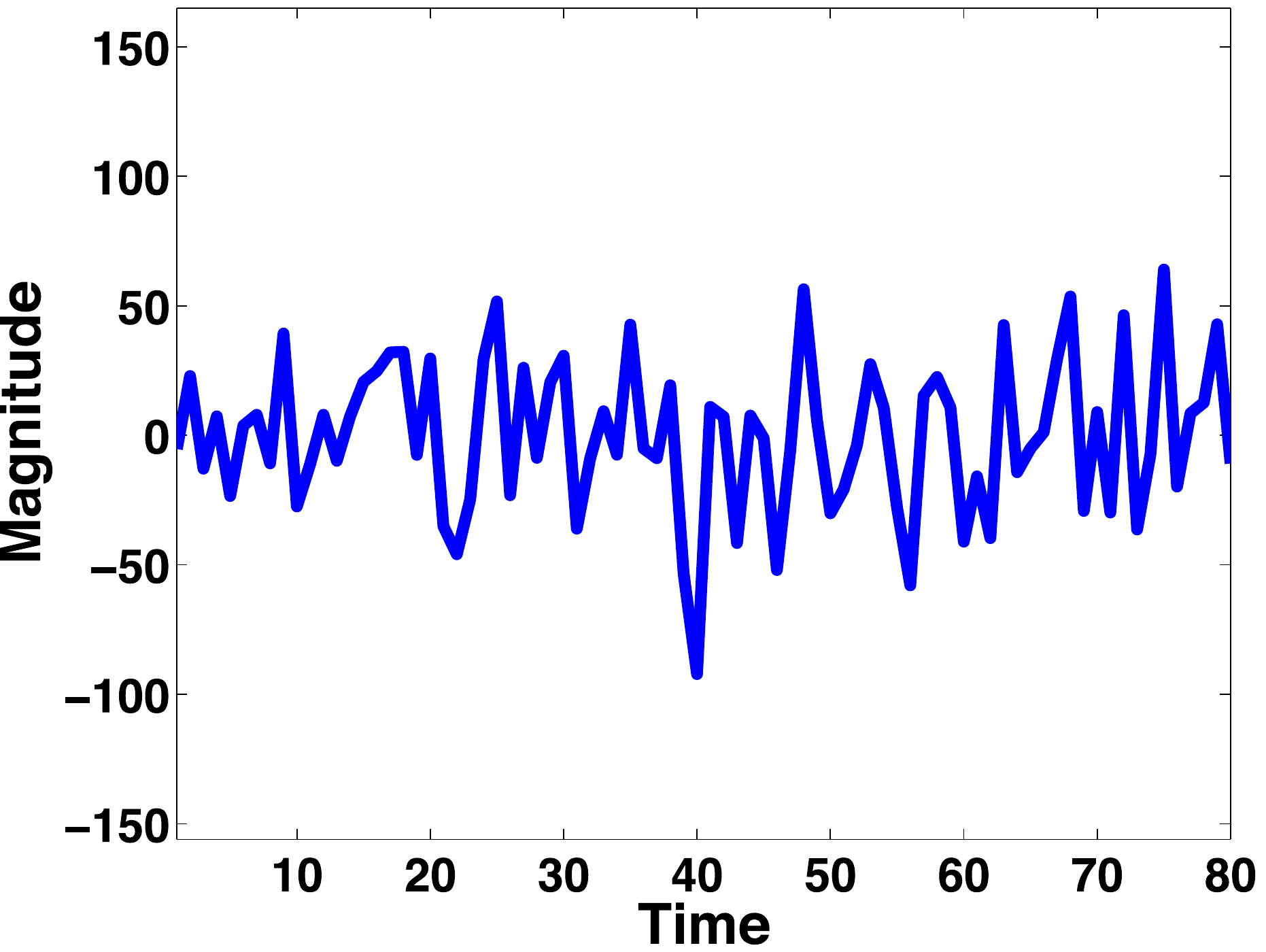}
}
\caption{Time series from the cluster~I ({\SF A}),
  cluster~II ({\SF B}), and cluster~III ({\SF C)}.
  \label{checker_kmeans_ts} }
\end{figure}
\noindent embedding, $\Psi (\bx_i), \; i = 1, \cdots,
3084$. The time series are color-coded according to the result of the
clustering.  The dataset is partitioned into four clusters.  For
comparison purposes, we show the embedding generated by ISOMAP
\citep{Tenenbaum00} (Fig. \ref{checker_pca}-left), and the embedding
obtained by projecting the time series on the first three PCA axes
(Fig. \ref{checker_pca}-right). ISOMAP makes it possible to
reconstruct a low dimensional nonlinear structure embedded in high
dimension. The algorithm computes the%
\begin{figure}[H]
\centerline{\hspace*{-1mm}\raisebox{10mm}{11}
\includegraphics[angle=90,width = 8pc]{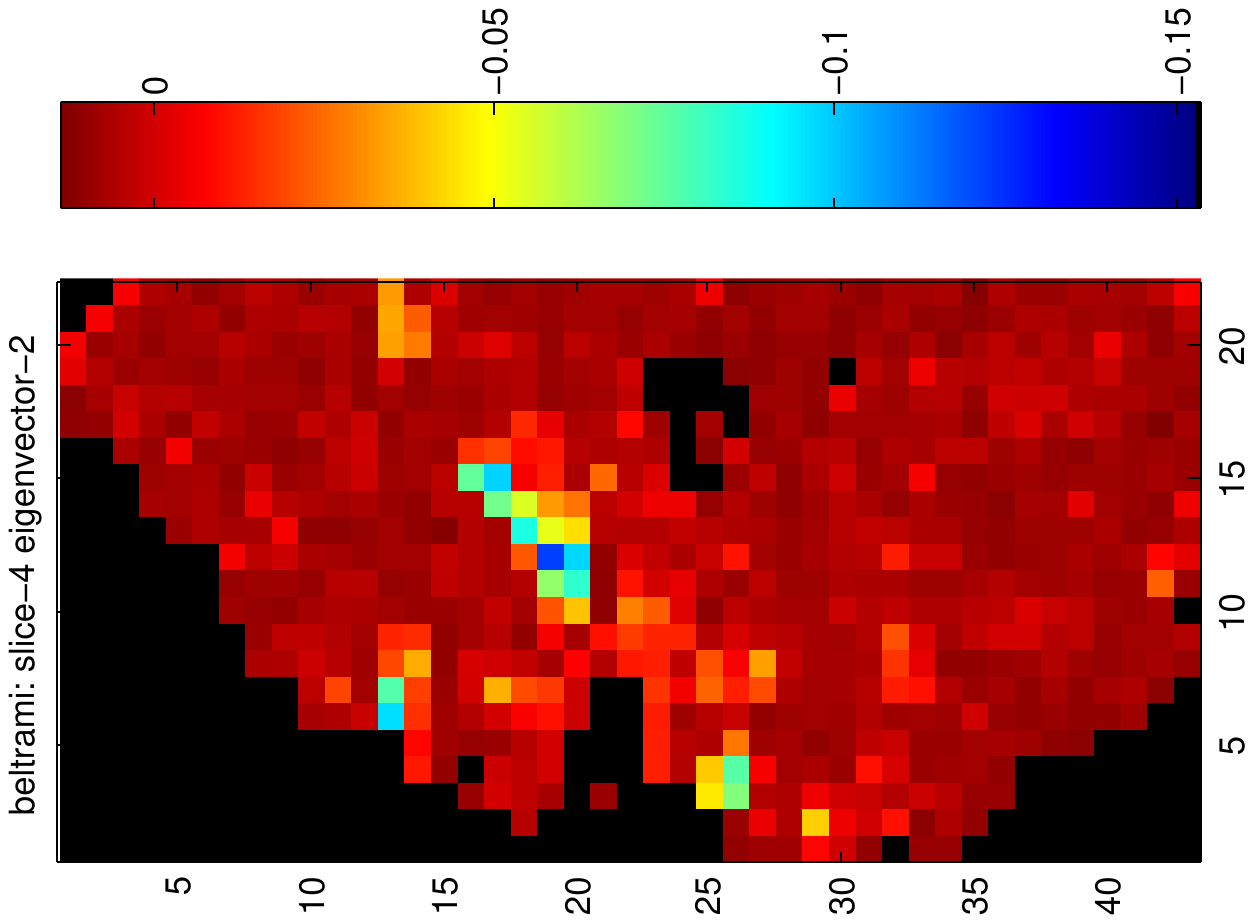}
\includegraphics[angle=90,width = 8pc]{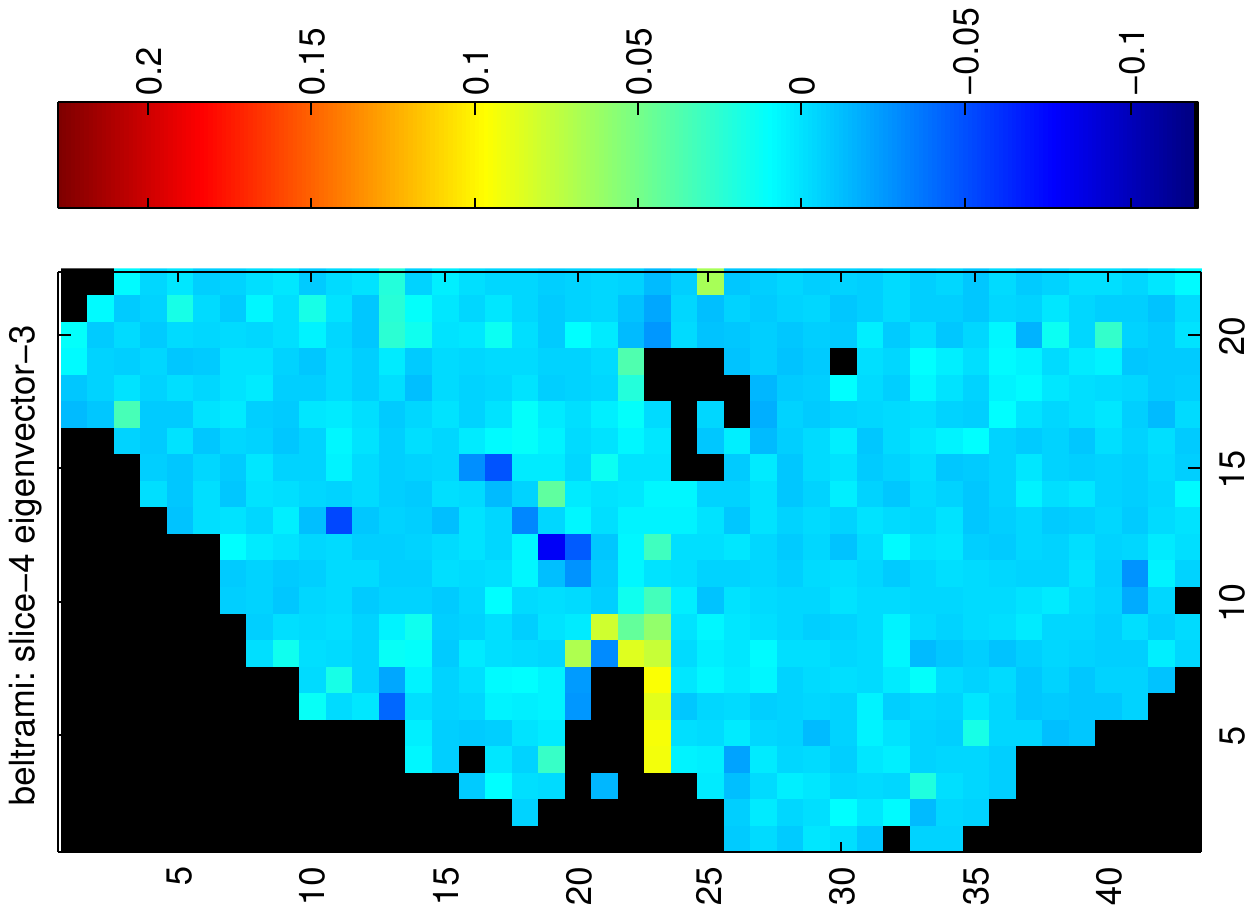}
}
\centerline{\hspace*{-2mm}\raisebox{10mm}{10}
\includegraphics[angle=90,width = 8pc]{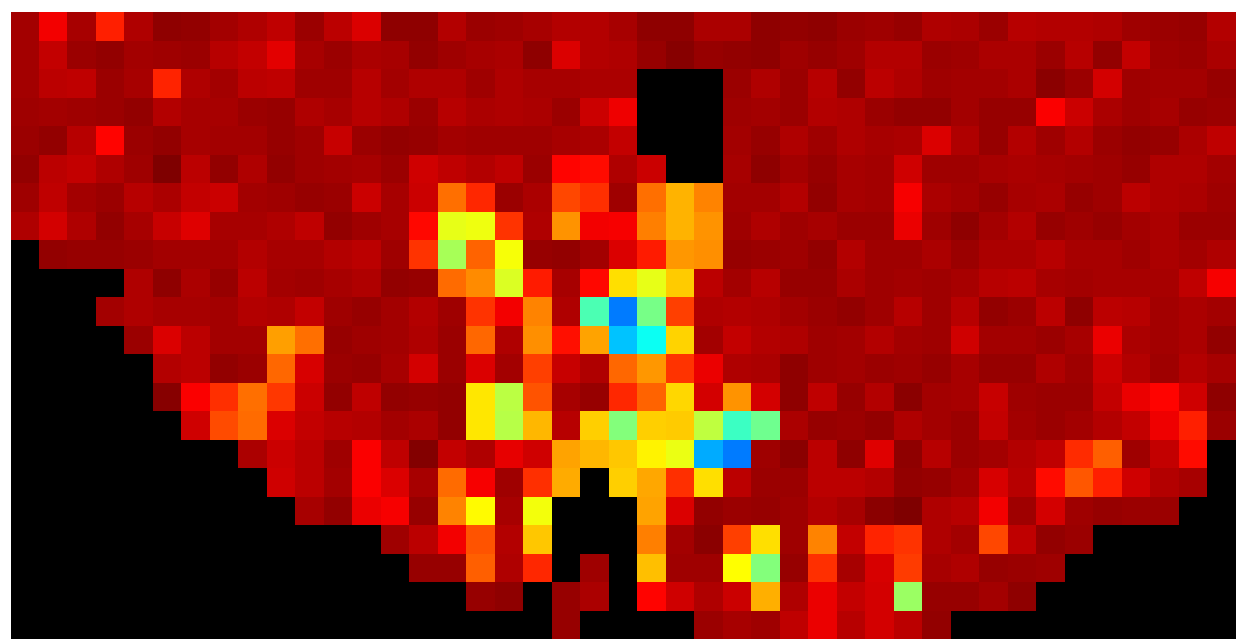}
\includegraphics[angle=90,width = 8pc]{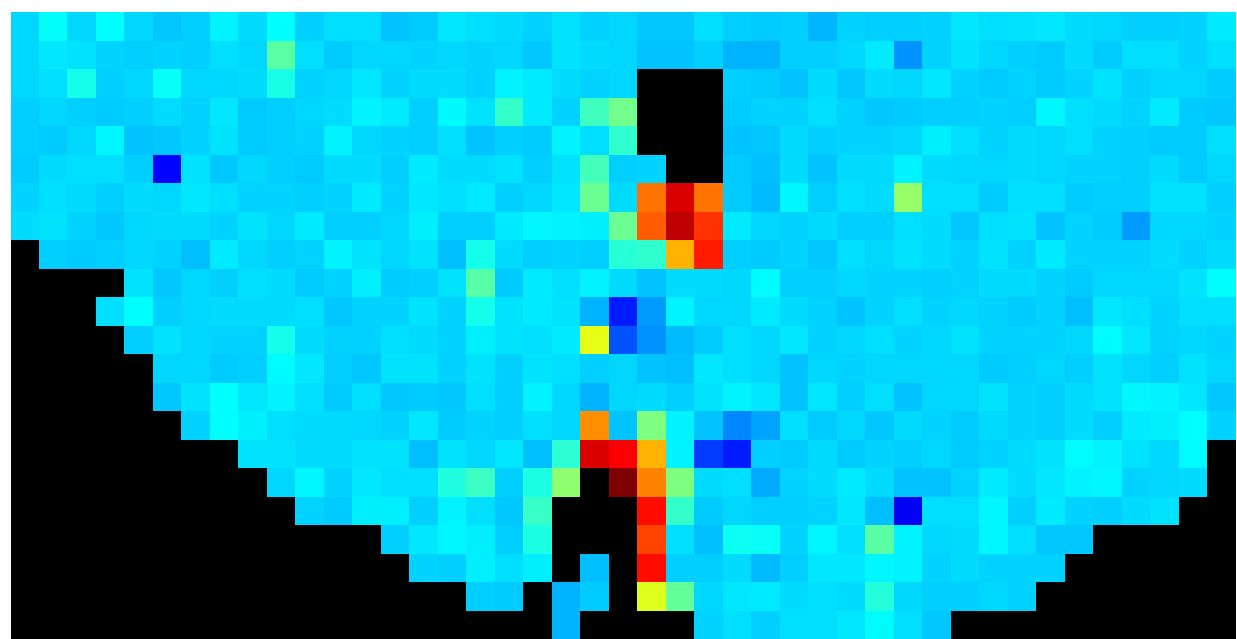}
}
\centerline{\raisebox{9mm}{9}
\includegraphics[angle=90,width = 8pc]{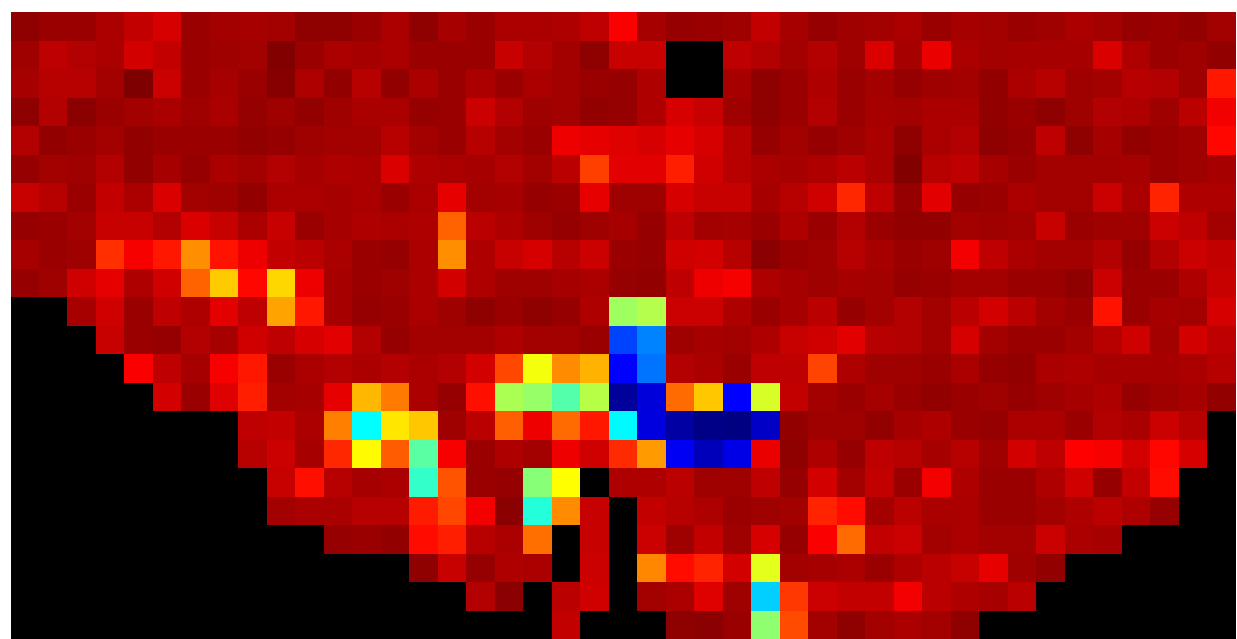}
\includegraphics[angle=90,width = 8pc]{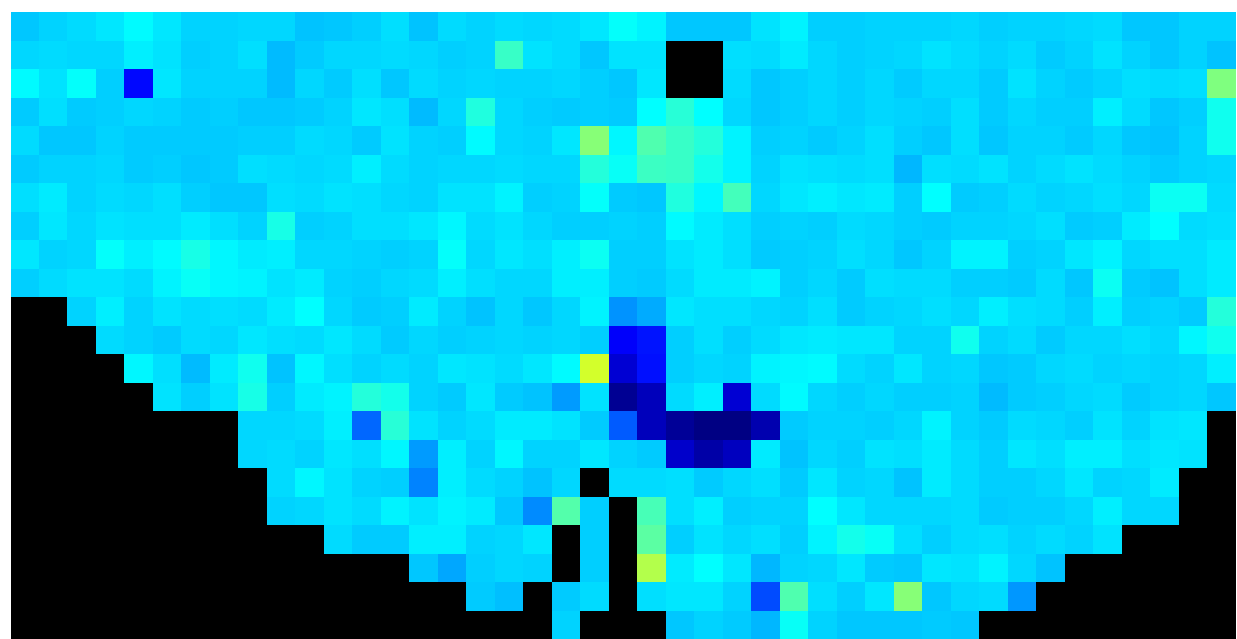}
}
\centerline{\raisebox{9mm}{8}
\includegraphics[angle=90,width = 8pc]{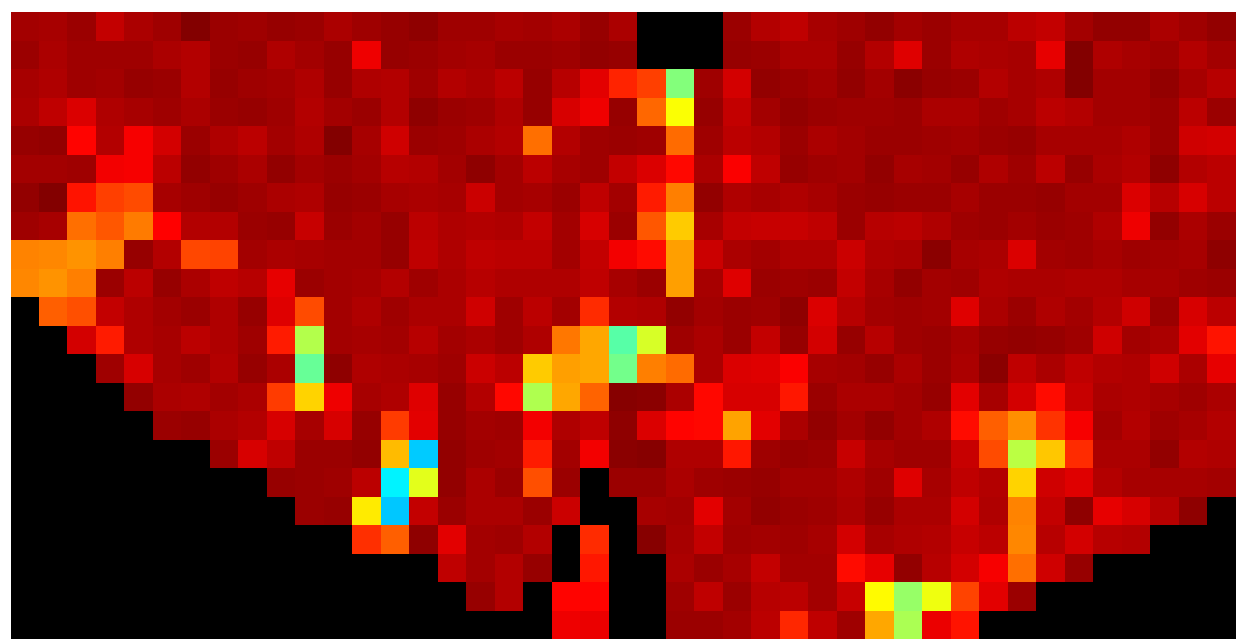}
\includegraphics[angle=90,width = 8pc]{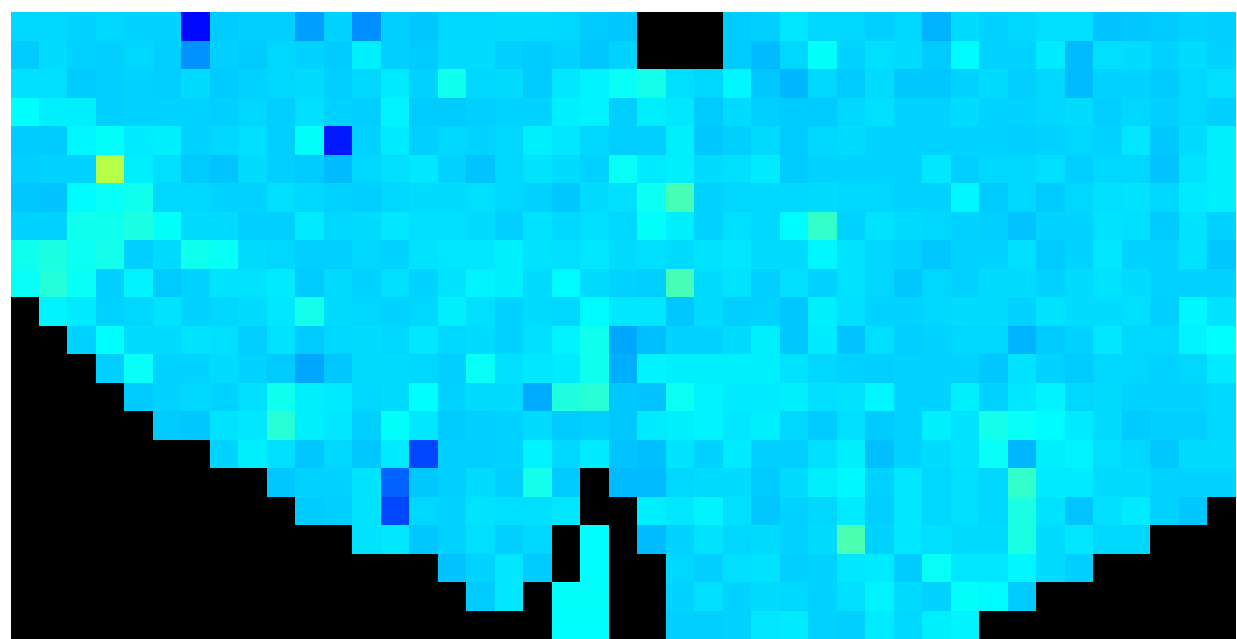}
}                                                 
\centerline{\hfill $\bfi_2$  \hfill $\bfi_3$ \hfill} 
\caption{Eigenvectors $\bfi_2$ and $\bfi_3$ visualized as images.
\label{checker_eigen}}
\end{figure}
\noindent  pairwise geodesic distance
between any two points in the dataset, and uses multidimensional
scaling to embed the dataset in low dimension. As explained in section
\ref{commute}, the geodesic distance is not appropriate for fMRI data
that are very noisy. For this reason an embedding based on commute
time is more robust.  The representation given by PCA and ISOMAP,
shown in Fig. \ref{checker_pca} left and right, are less conspicuous
than the representation obtained by our approach. No low dimensional
structures are apparent in these two plots.  The plot of the residual
error (Fig. \ref{checker_embedding}-right) indicates that eigenvectors
$\bfi_2$ and $\bfi_3$ create the largest drop in the residual energy,
and should be the most useful.  We use $K=3$ coordinates: $\bfi_2,
\bfi_3$ and $\bfi_4$ to embed the dataset. In order to interpret the
role of the clusters, we select several time series from each cluster
(identified by circle in the scatter plot in
Fig. \ref{checker_embedding}) and plot them in Fig.
\ref{checker_kmeans_ts}. The time series from the red cluster
(Fig. \ref{checker_kmeans_ts}-{\SF A}) are typical hemodynamic
responses to a periodic stimulus. The time series at the tip of the
red cluster, marked by a red circle,  is the red curve in
Fig. \ref{checker_kmeans_ts}-{\SF A}.  It is the strongest activation
pattern. Times series in the middle of the red cluster (blue and black
circles) exhibit weaker activation (blue and black curves).
We interpret cluster~I as voxels activated by the stimulus.
The embedding has organized the time series according to the strength
of the activation: strong activation at the tip and weak activation at
the base%
\begin{figure}[H]
\centerline{
  \includegraphics[angle=180,width=9pc]{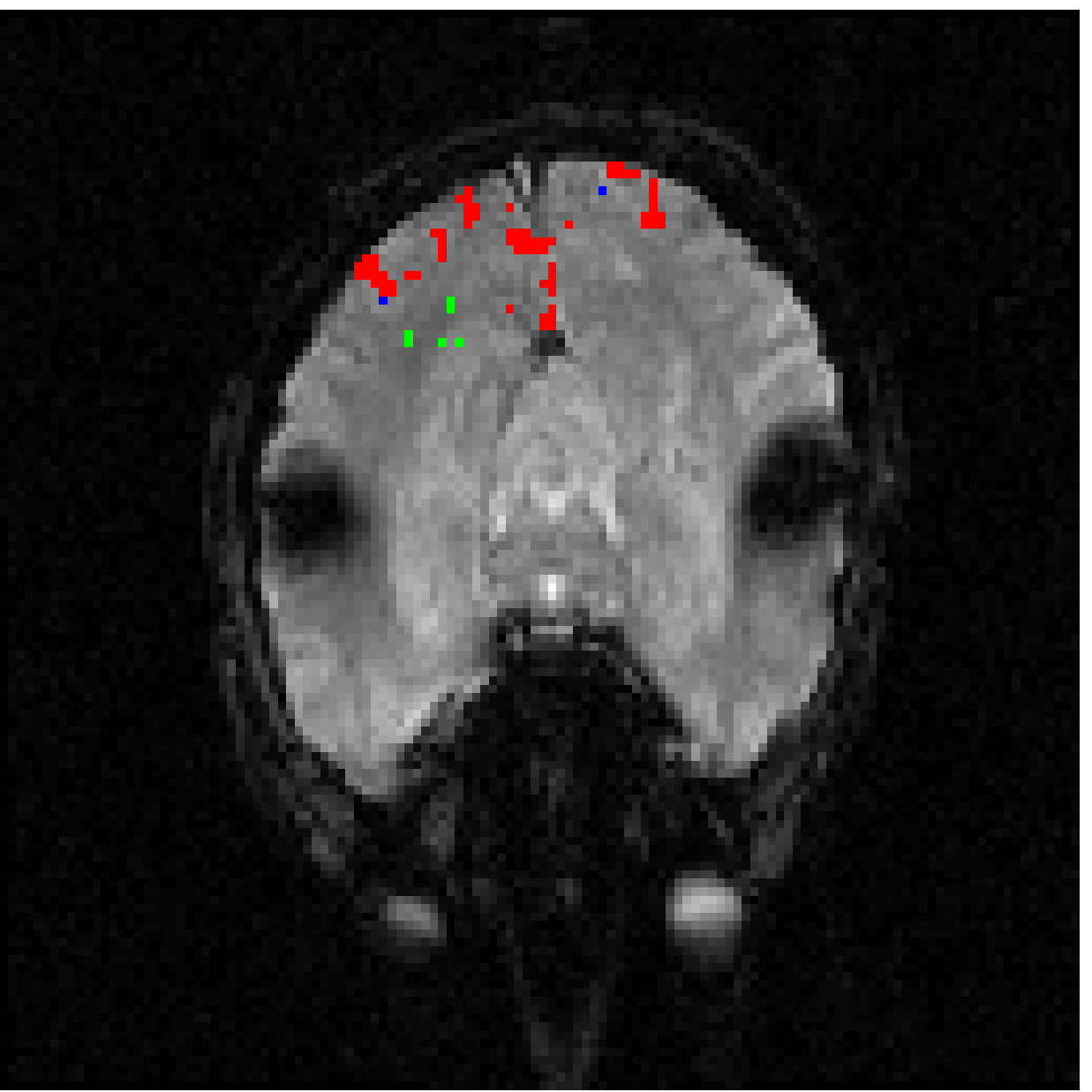}
  \includegraphics[angle=180,width=9pc]{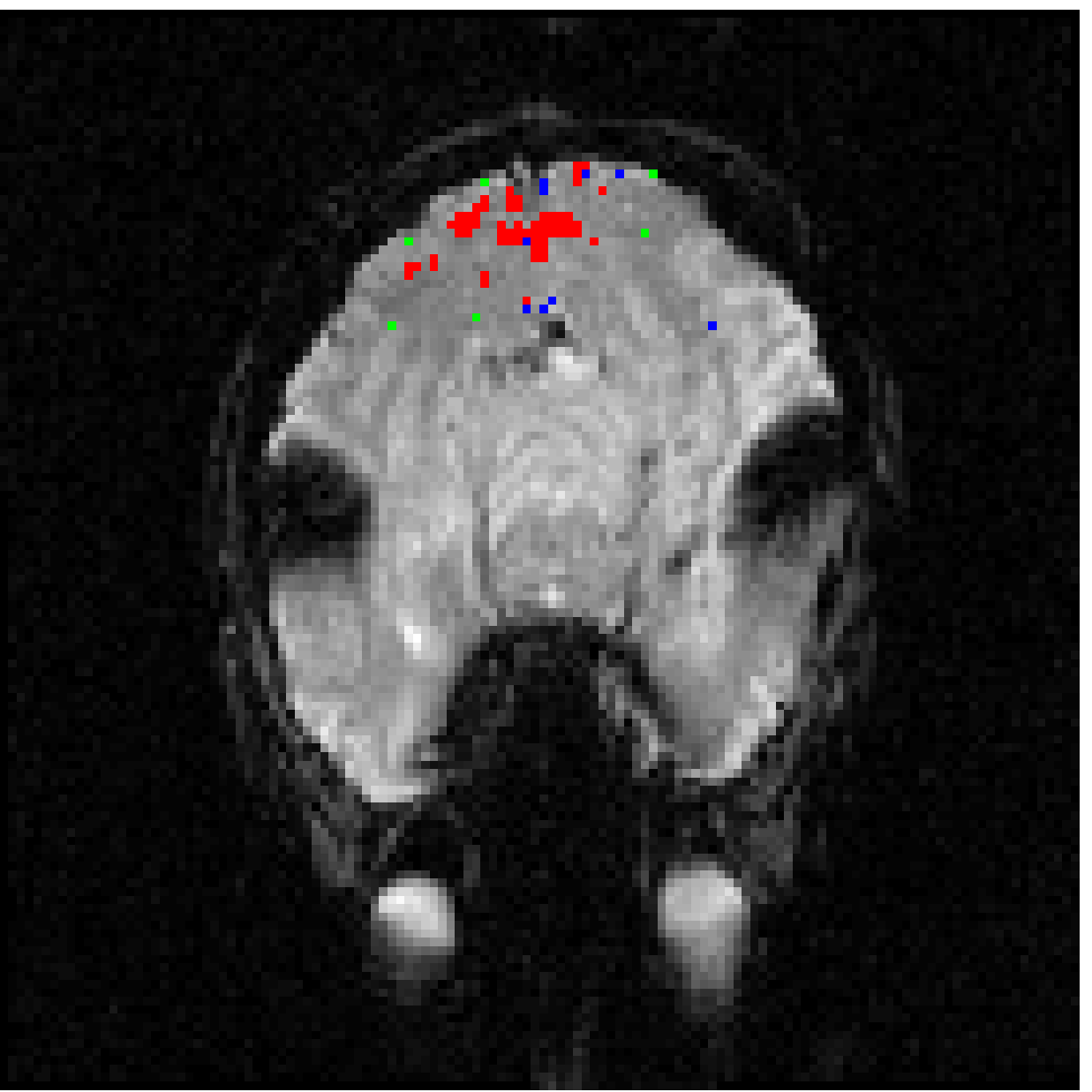}
}
\centerline{
  \includegraphics[angle=180,width=9pc]{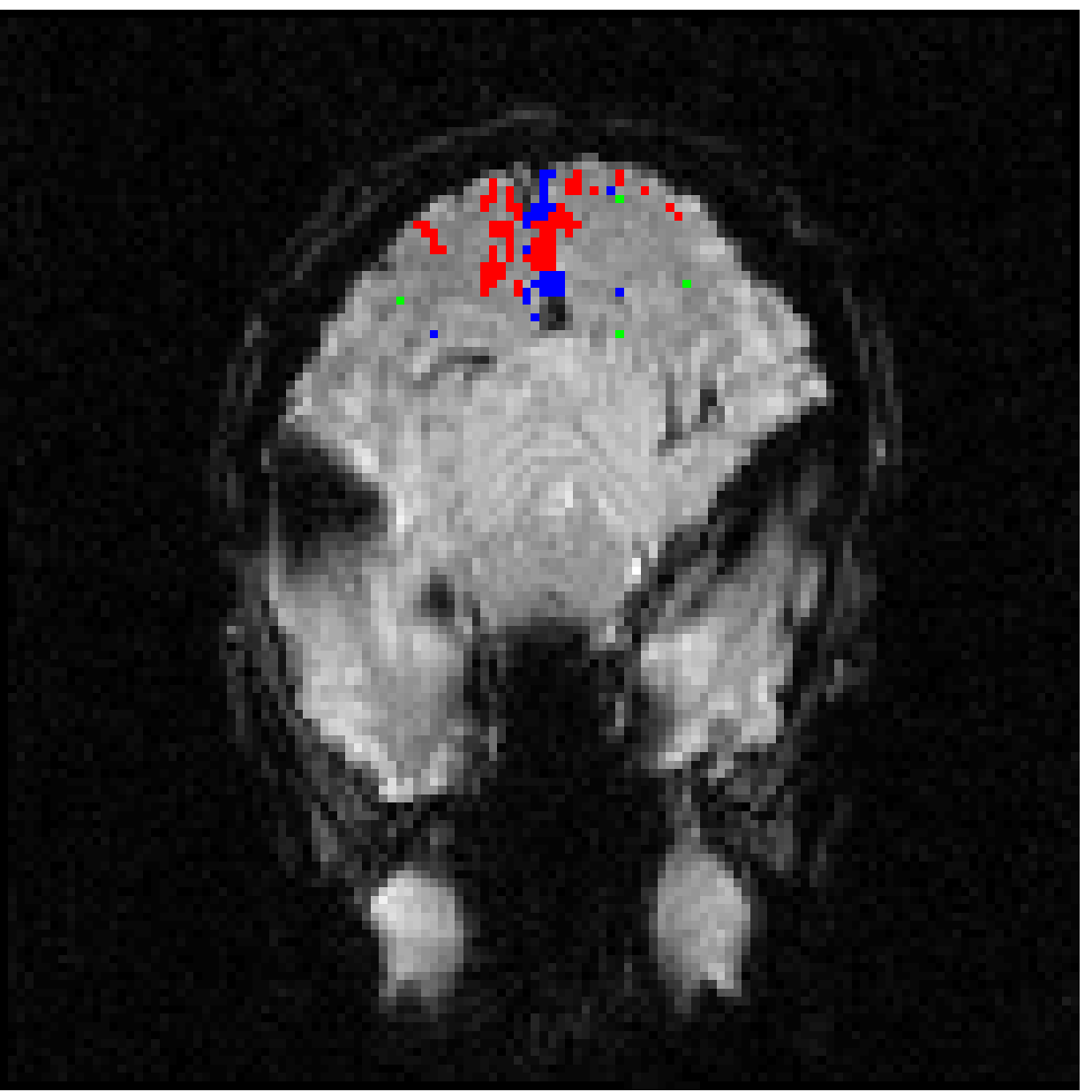}
  \includegraphics[angle=180,width=9pc]{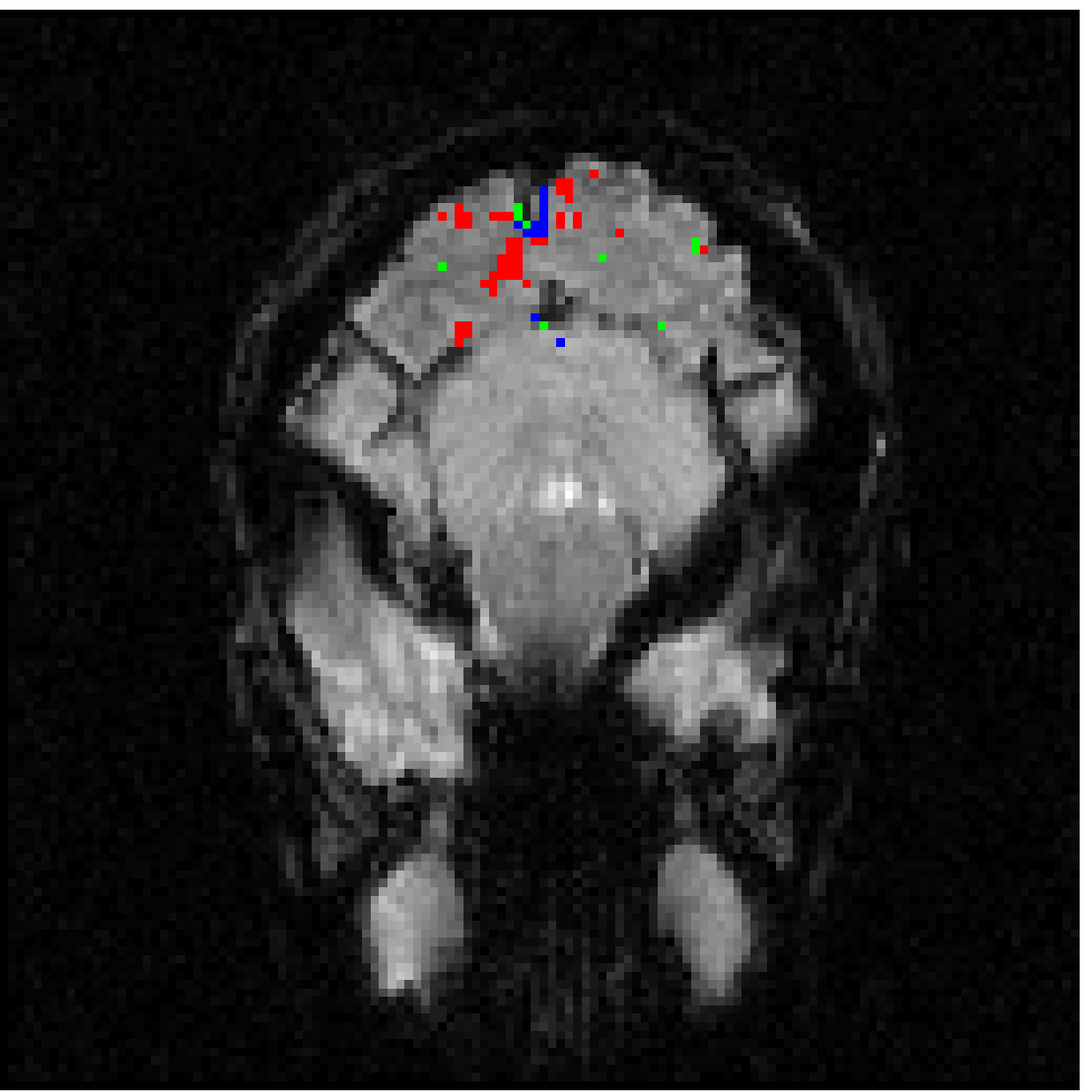}
}
\end{figure}
\noindent  of the branch (close to the background activity).  The time
series from the blue cluster have a high frequency
(Fig. \ref{checker_kmeans_ts}-{\SF B}), and are grouped together
inside the brain (Fig. \ref{checker_map}). These time series could be
related to non-task related physiological responses, such as a
pulsating vein. Finally, the time series from the green cluster are
less structured (Fig. \ref{checker_kmeans_ts}), and are scattered
across the region of analysis (Fig. \ref{checker_map}). We were not
able to interpret the physiological role of these time series.\\

{\noindent \em So what do the eigenvectors look like ?}\\
As explained in section \ref{commute}, we consider the eigenvectors
$\bfi_k$ to be functions of the nodes $i$ of the graph. We can
therefore represent $\bfi_k$ as an image: each voxel $\bx_i$ is
color-coded according to the value of $\bfi_k(i)$. The majority of the
values of $\bfi_2$ are positive (deep red, in
Fig. \ref{checker_eigen}-left). A few voxels take negative values
(yellow and cyan in Fig. \ref{checker_eigen}-left).  The nodal lines
(where $\bfi_2$ changes sign) are localized around the area of
activation in the visual cortex. We can check in
Fig. \ref{checker_embedding} that the activated time series (red
cluster) have a negative $\bfi_2$ coordinate. In fact, $\bfi_2$ is
known as the Fiedler vector \citep{Chung97} and is used to optimally
split a dataset into two parts.  (Fig. \ref{checker_eigen}). We
computed the activation map obtained using a GLM. The regressor is
computed by convolving the haemodynamic response defined by
(\ref{hrf}) with the experimental paradigm.  The activation map
(thresholded at $p=0.001$) is shown in Fig. \ref{checker_map}, and is
consistent with the activation maps obtained by our approach.
\subsection{In vivo data II: event-related dataset}
We apply our method to an event-related dataset.  \cite{Buckner00}
used fMRI to study age-related changes in functional anatomy.  The
subjects were instructed to press a key with their right index finger
upon the visual stimulus onset. The stimulus lasted for
$1.5$s. Functional images were collected using a Siemens $1.5$-T
Vision System with an asymmetric spin-echo sequence sensitive to BOLD
contrast (volume TR = $2.68$s, xy dimension: $64\times64$, voxel size:
$3.75\times 3.75$mm, 16 contiguous axial slices). Each run consists of
$128$ TRs.  For every $8$ images, the subjects were presented with one
of the two conditions: (i) the {\em one-trial} condition where a
single stimulus was presented to the subject, and (ii) the {\em
  two-trial} condition where two consecutive stimuli were
presented. The inter-stimulus interval of $5.36 s.$ was sufficiently
large to guarantee that the overall response would be about twice as
large as the response to the {\em one-trial} condition. We analyzed
one run. After discarding the first and last four scans, the run
included $15$ trials ($8$ {\em one-trial} and $7$ {\em two trial}
conditions) of $8$ temporal samples.  Time series from the {\em
  one-trial} and {\em two-trial} conditions were averaged
separately. Therefore, each voxel gave rise to two average time series
of 8 samples. The linear trend was removed%
\begin{figure}[H]
\centerline{
  \includegraphics[width=13pc]{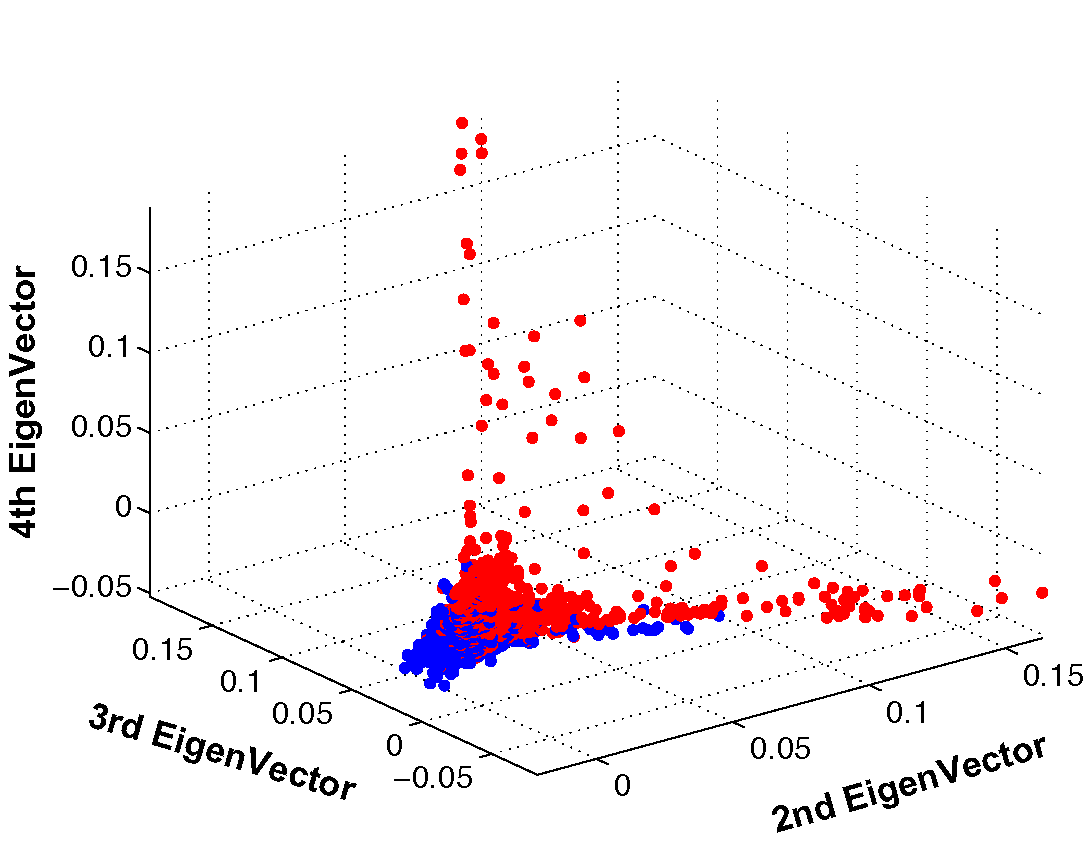}
}
\centerline{
  \includegraphics[width=13pc]{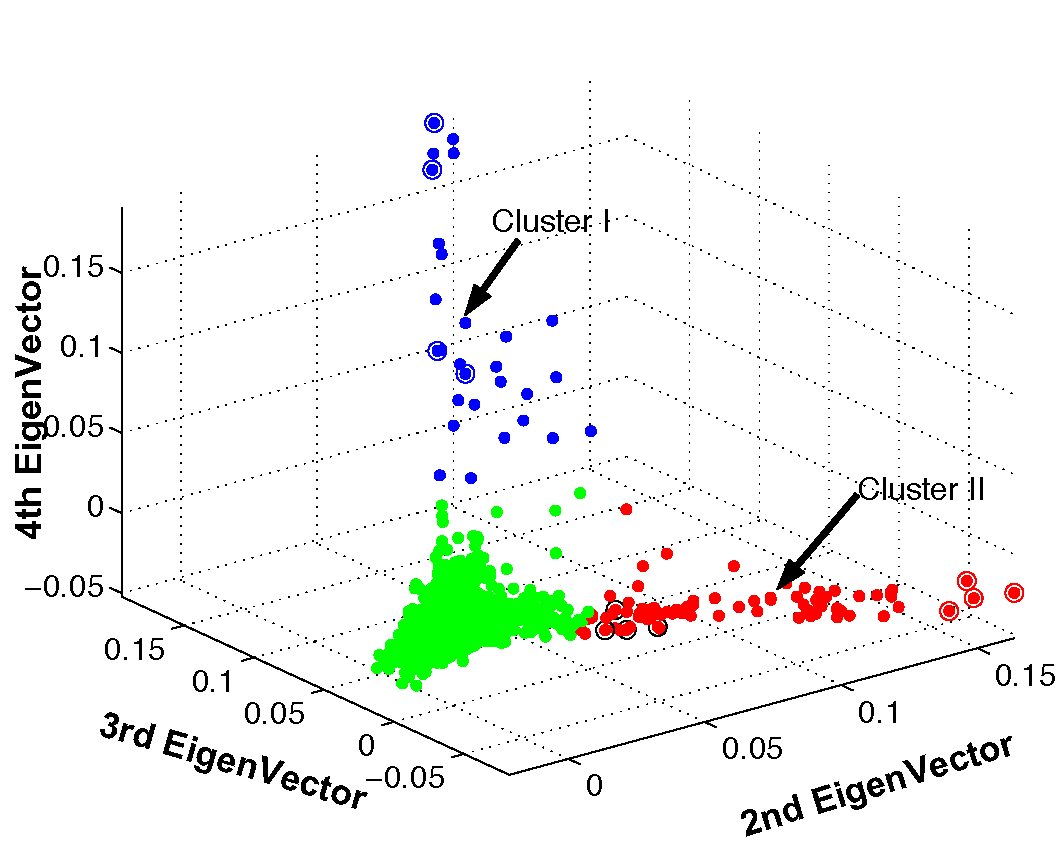}
}
\caption{Three-dimensional embedding. Left: {\em one-trial} (blue) and {\em two-trial} (red) conditions
  are well separated. Right:  cluster~I (blue), cluster~II (red) and background 
  (green). Time series  marked by a circle  are shown in Fig. \ref{dementia_kmean_ts}. 
  \label{dem_embedding}}
\end{figure}
\begin{figure}[H]
\centerline{
  \includegraphics[width=13pc]{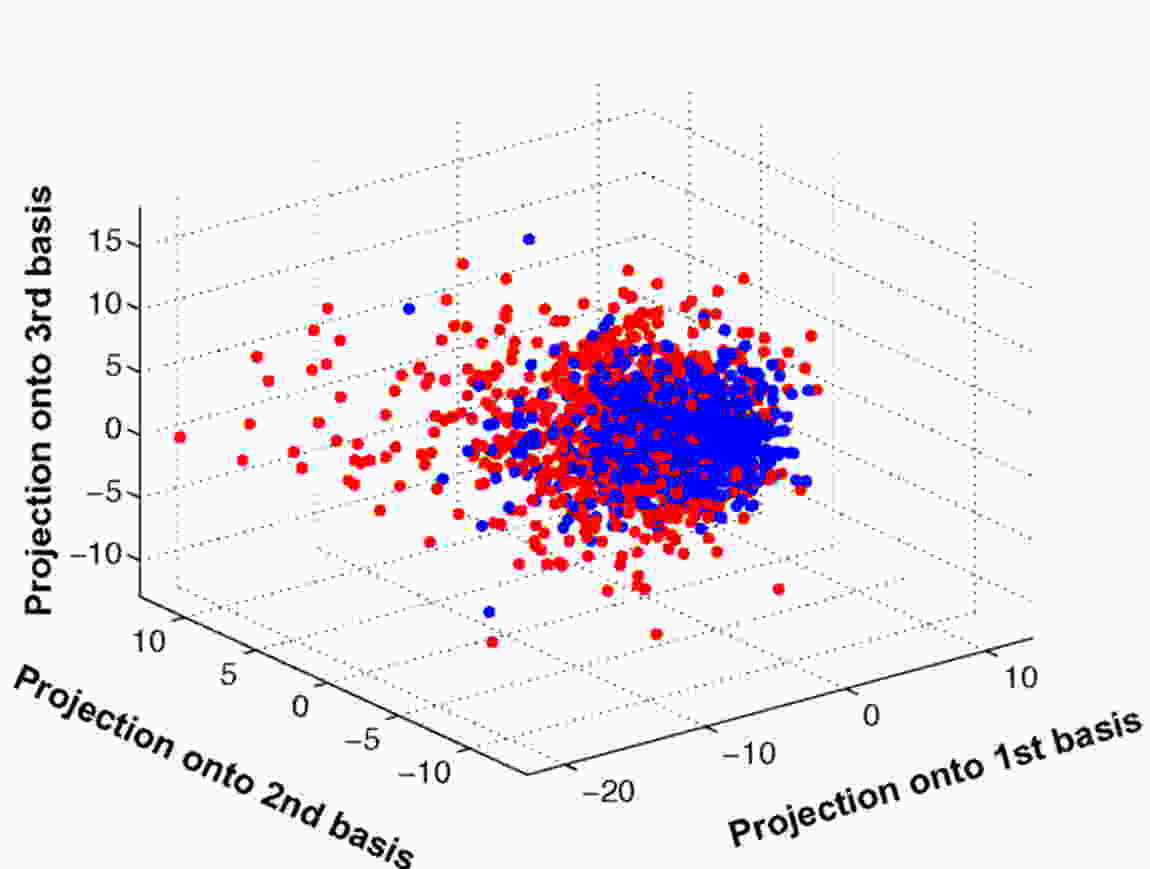}
}
\centerline{
\includegraphics[width=13pc]{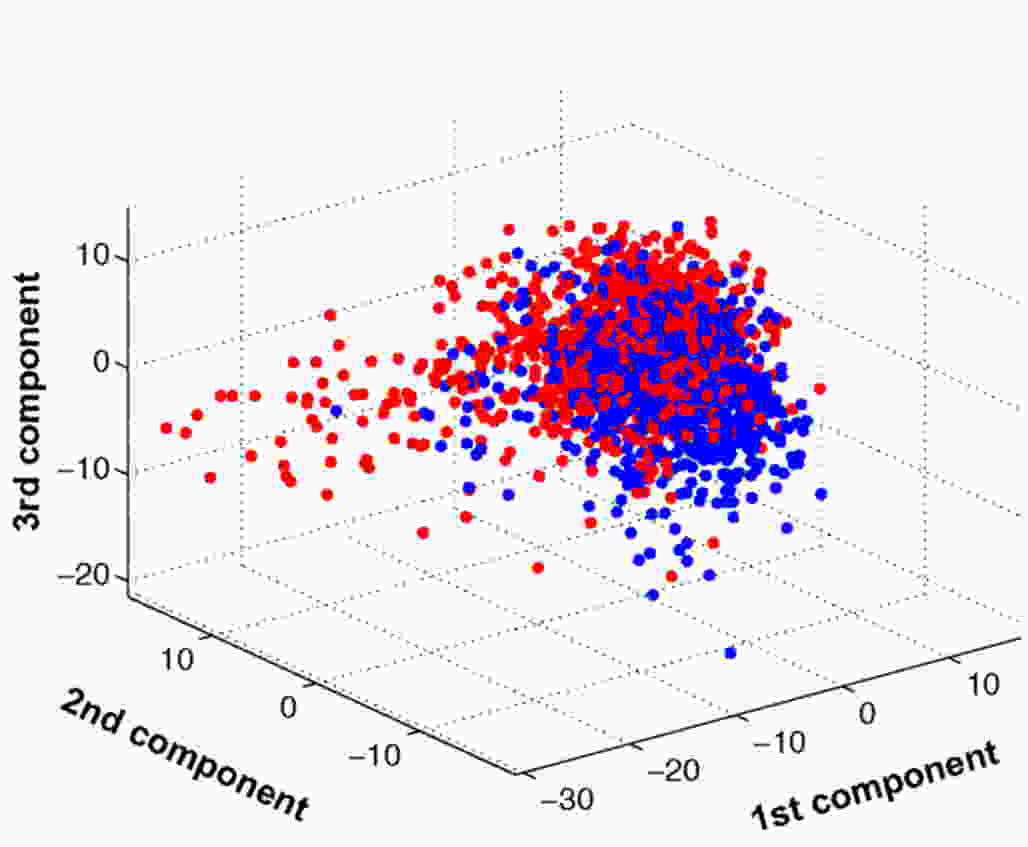}
}
\caption{Low dimensional embedding obtained by PCA (left), and ISOMAP (right).
{\em One-trial} condition (blue) and {\em two-trial} condition (red) are mingled.
\label{dem_pca}}
\end{figure}
\noindent  from all average time
series.  The results published in \citep{Buckner00} show activation in
the visual cortex, motor cortex, and cerebellum. We focus our analysis
on four contiguous axial slices (7, 8, 9 and 10) that extend from the
superior caudate nucleus to the midlevel diencephalon. We extract a
region ($1025$ intracranial voxels) that extends from the occipital
posterior horn of the lateral ventricles until the end of the
occipital lobe.  The total number of time series included in our
analysis is $2050$: two time series ({\em one-trial} and {\em
  two-trial} conditions) of $T=8$ samples for each voxel.

\begin{figure}[H]
\centerline{
\raisebox{6pc}{\hspace*{3pc}\SF A}
\hspace*{-4pc}\includegraphics[width=9pc]{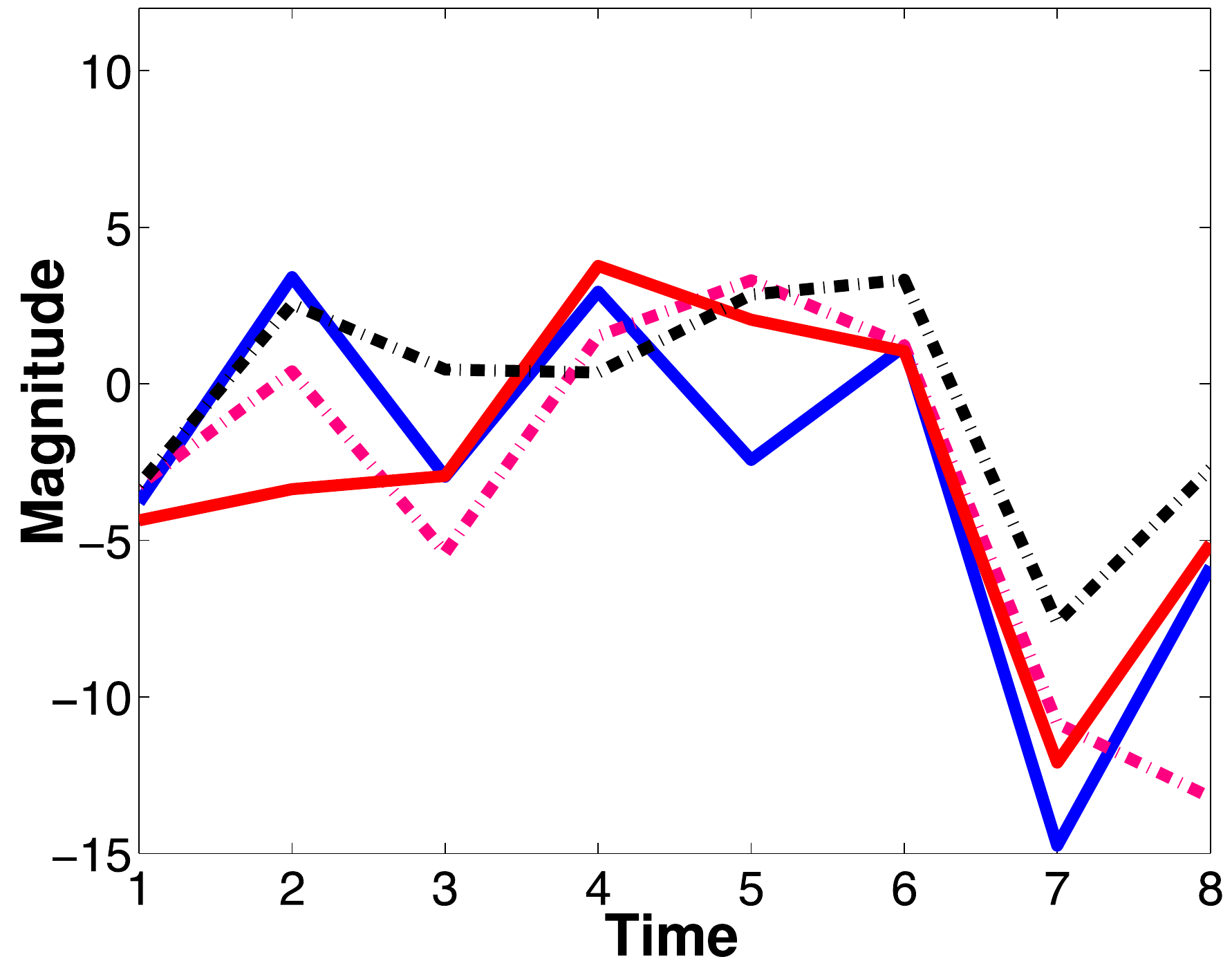}
\raisebox{6pc}{\hspace*{3pc}\SF B}
\hspace*{-4pc}\includegraphics[width=9pc]{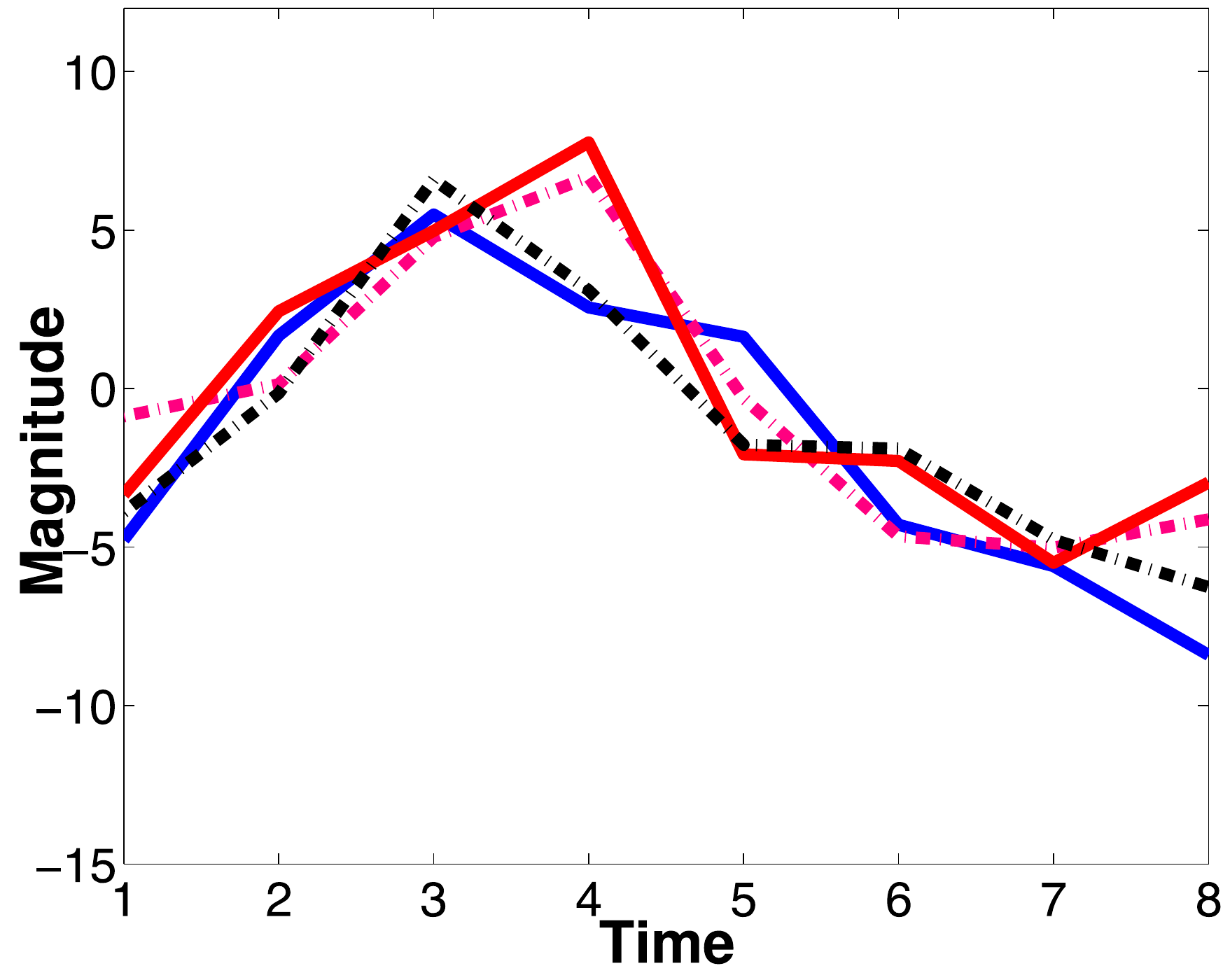}
}
\centerline{
\raisebox{6pc}{\hspace*{1pc}\SF C}
\hspace*{-2pc}\includegraphics[width=9pc]{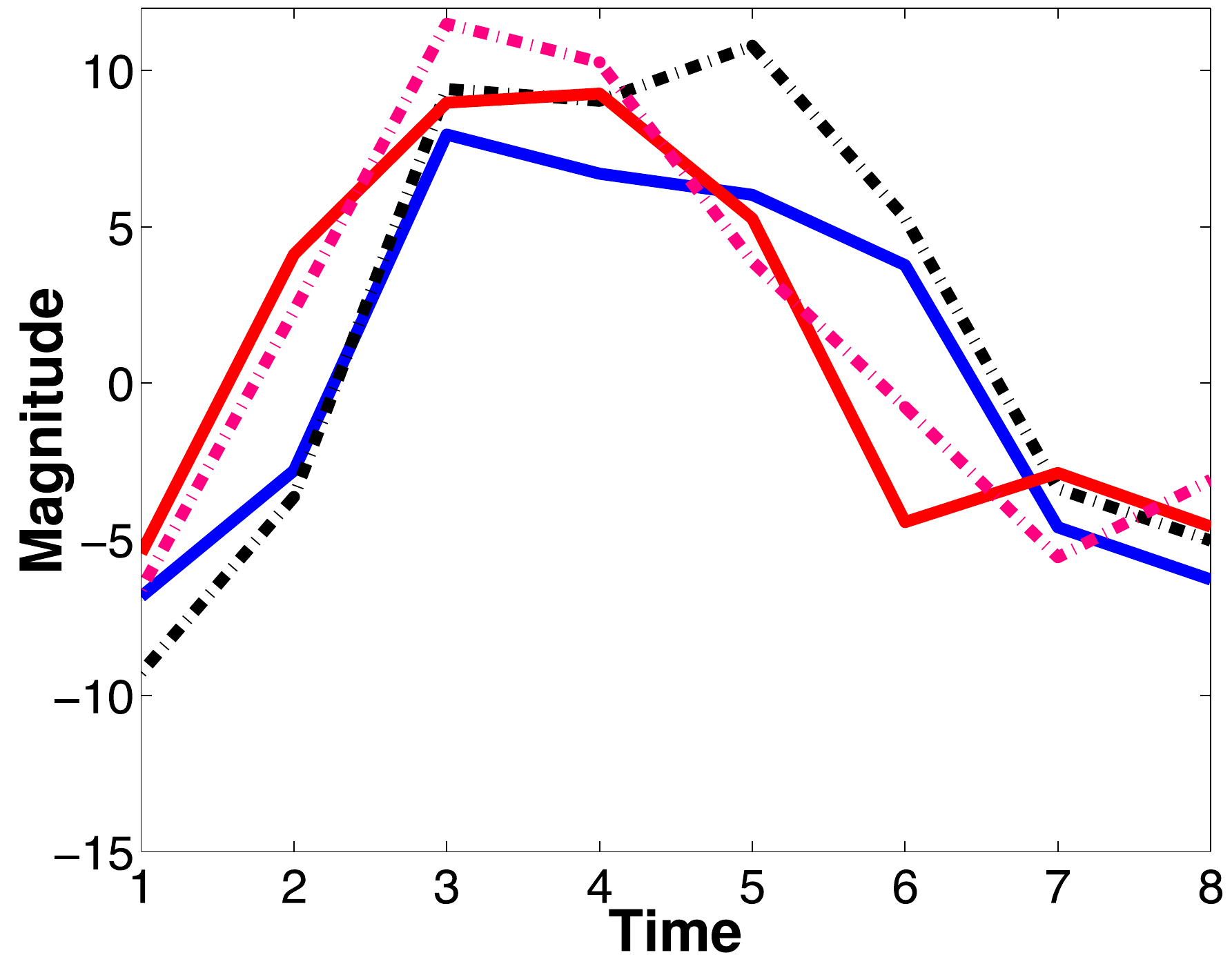}
}
\caption{Four representative time series from cluster~I ({\SF A}) , cluster~II
   with {\em one-trial} condition ({\SF B}), and from cluster~II with {\em two-trial}
   condition ({\SF C)}.
  \label{dementia_kmean_ts} }
\end{figure}

\begin{figure}[H]
\centerline{
  \includegraphics[angle=-90,width=9pc]{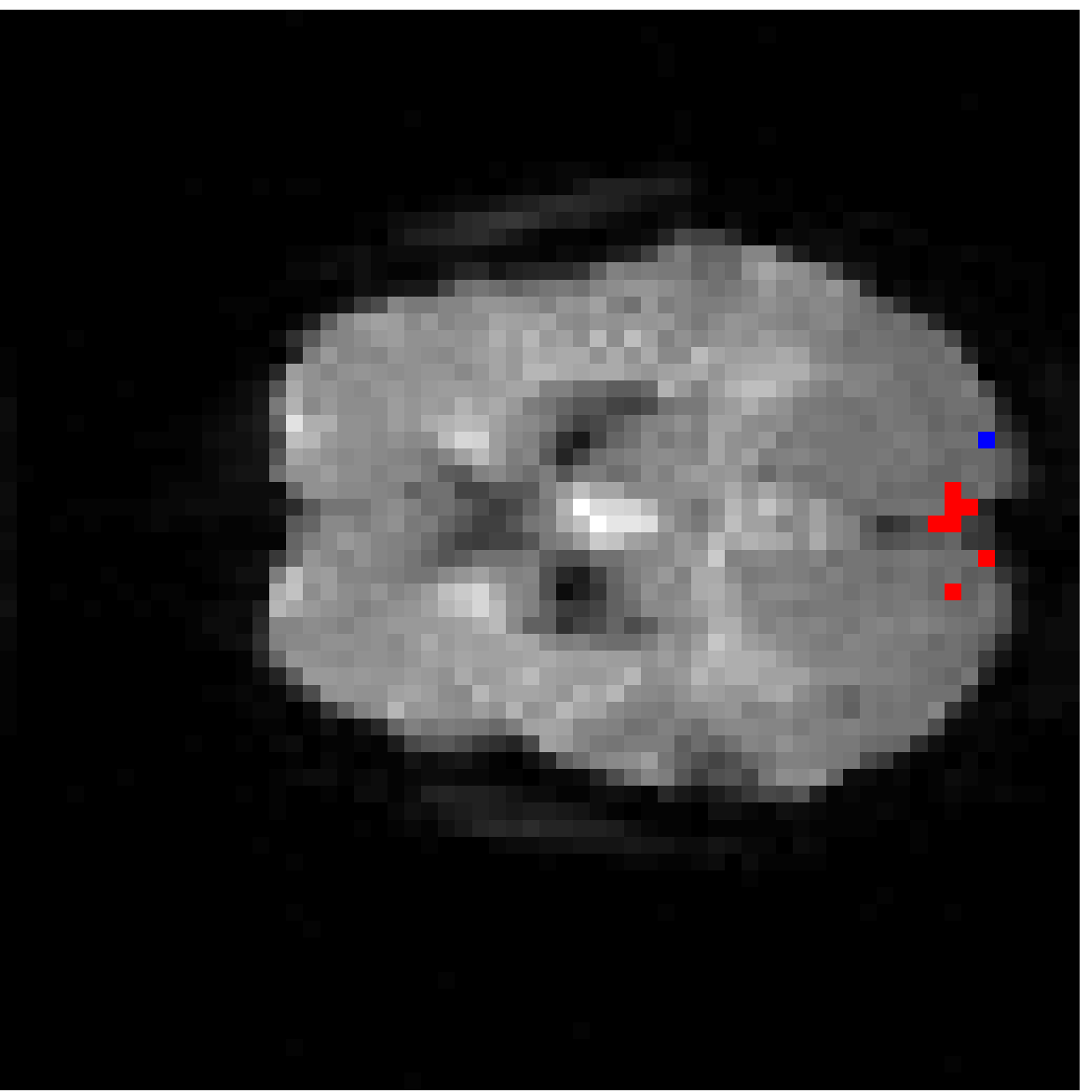}
  \includegraphics[angle=-90,width=9pc]{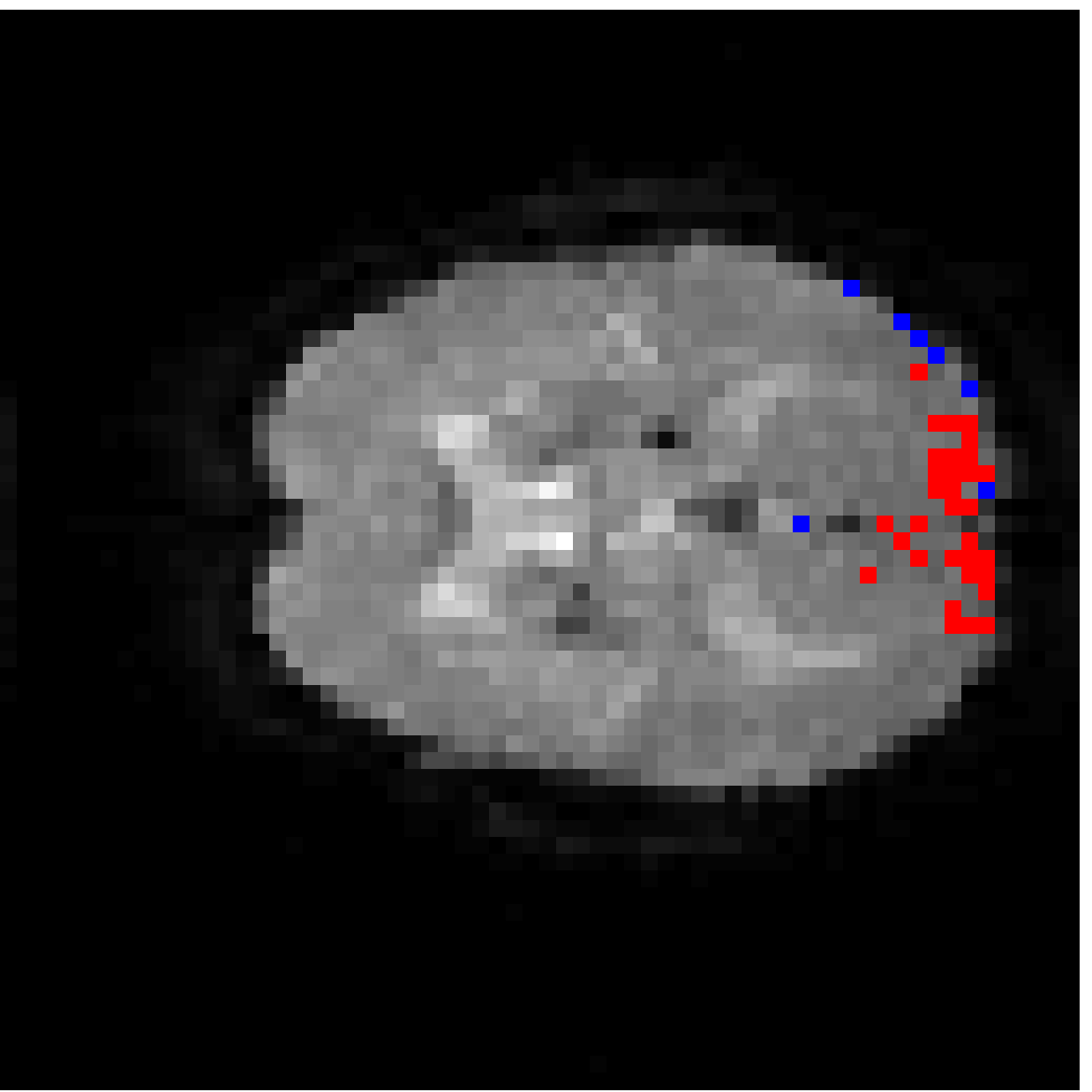}
}
\centerline{
  \includegraphics[angle=-90,width=9pc]{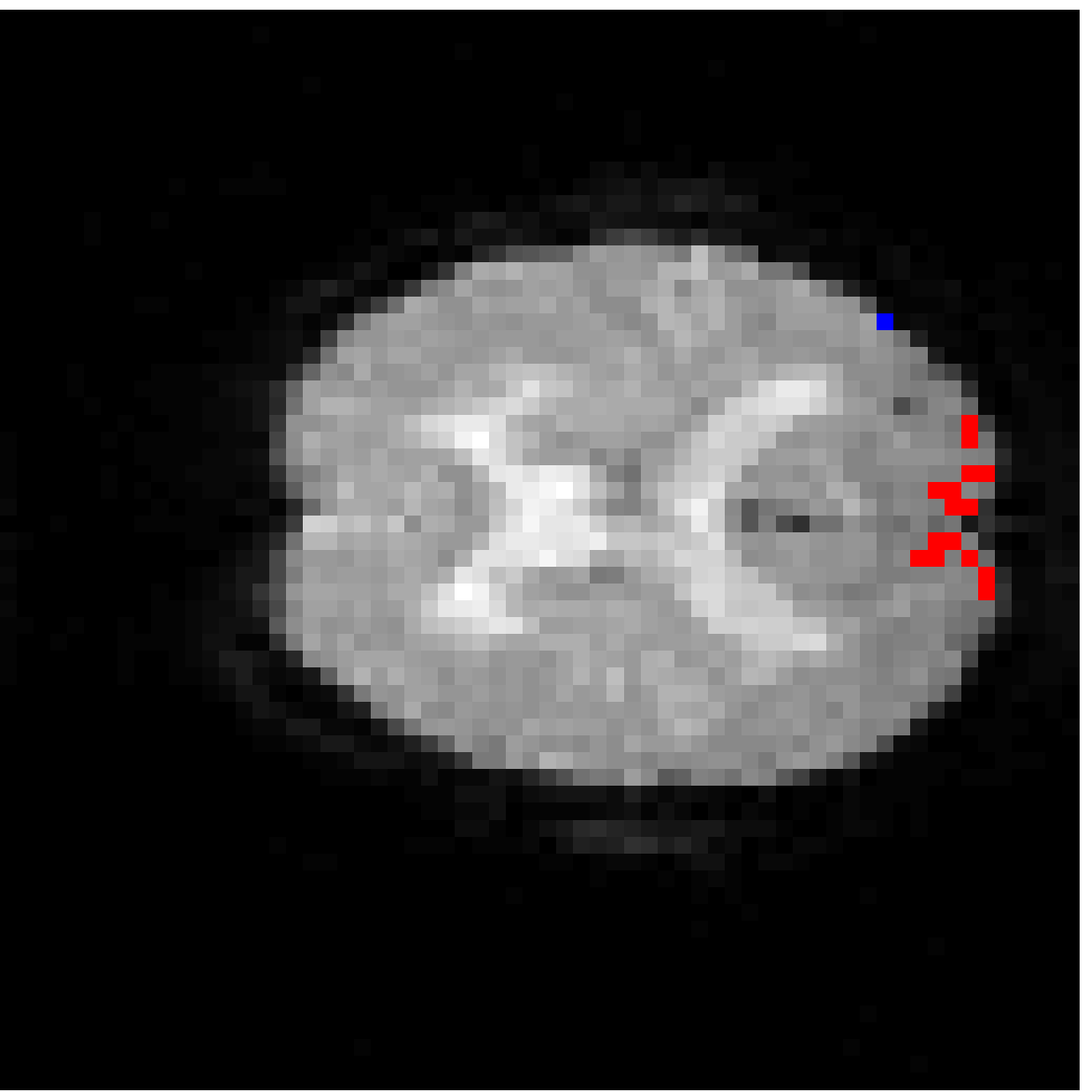}
  \includegraphics[angle=-90,width=9pc]{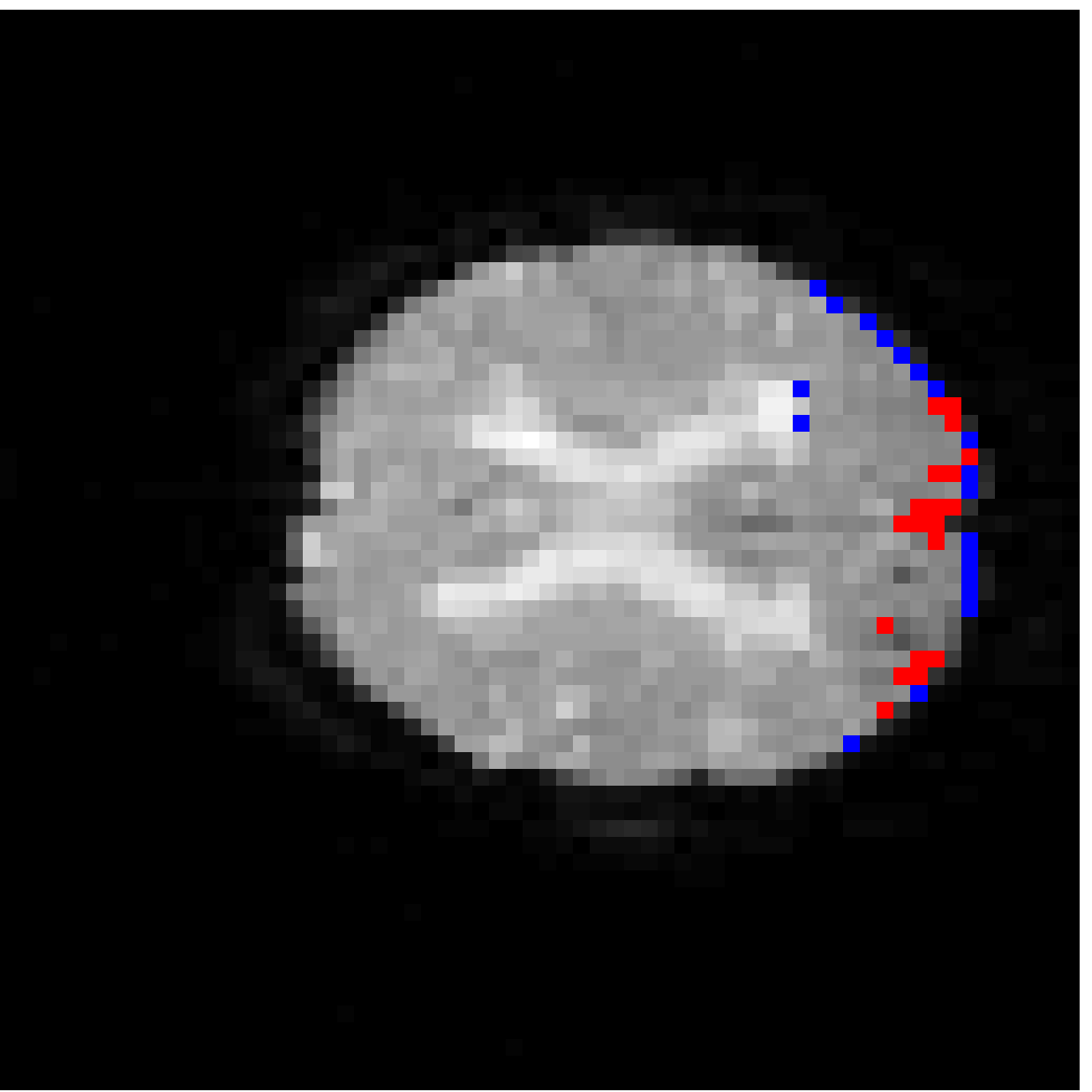}
}
\centerline{
  \includegraphics[angle=-90,width=9pc]{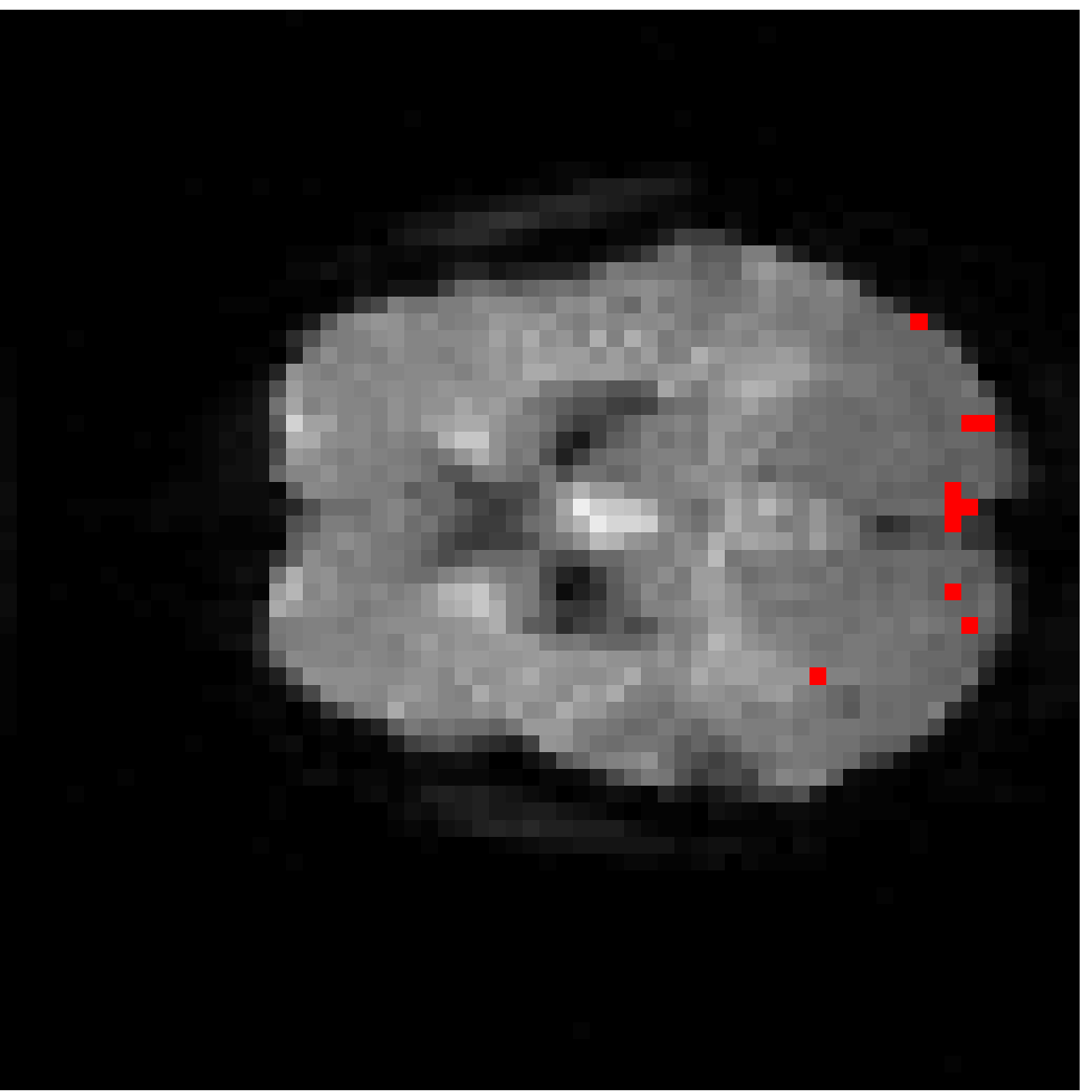}
  \includegraphics[angle=-90,width=9pc]{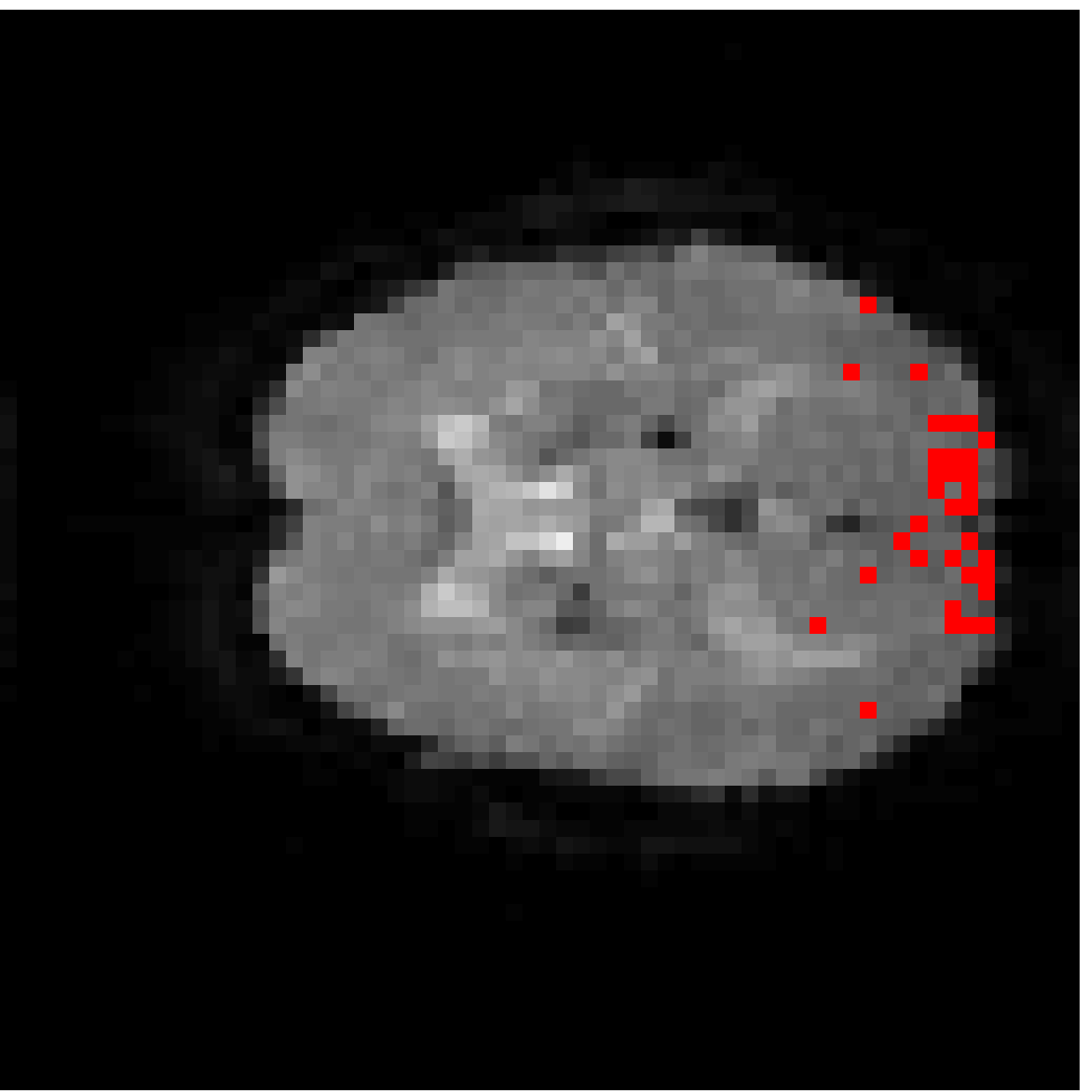}
}
\centerline{
  \includegraphics[angle=-90,width=9pc]{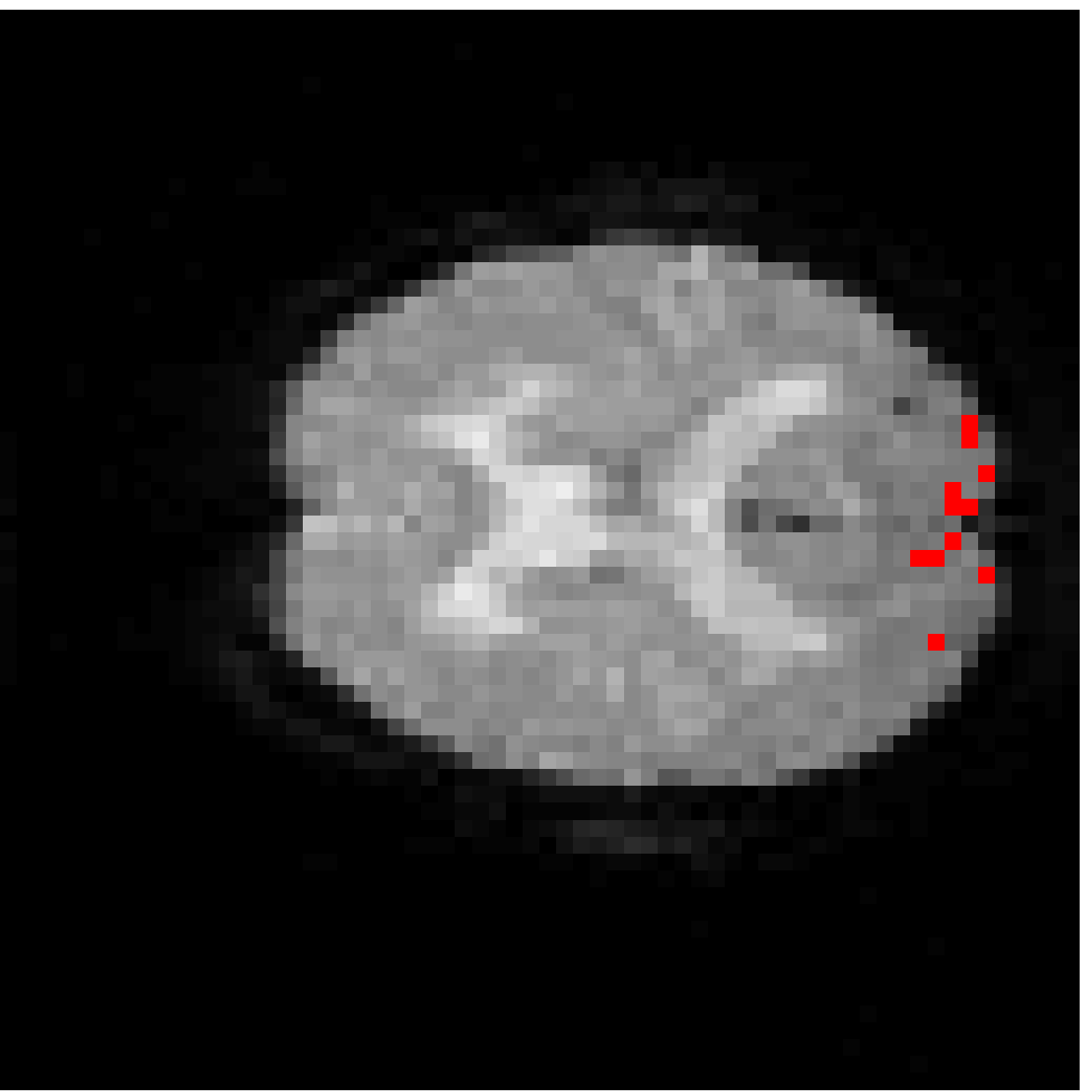}
  \includegraphics[angle=-90,width=9pc]{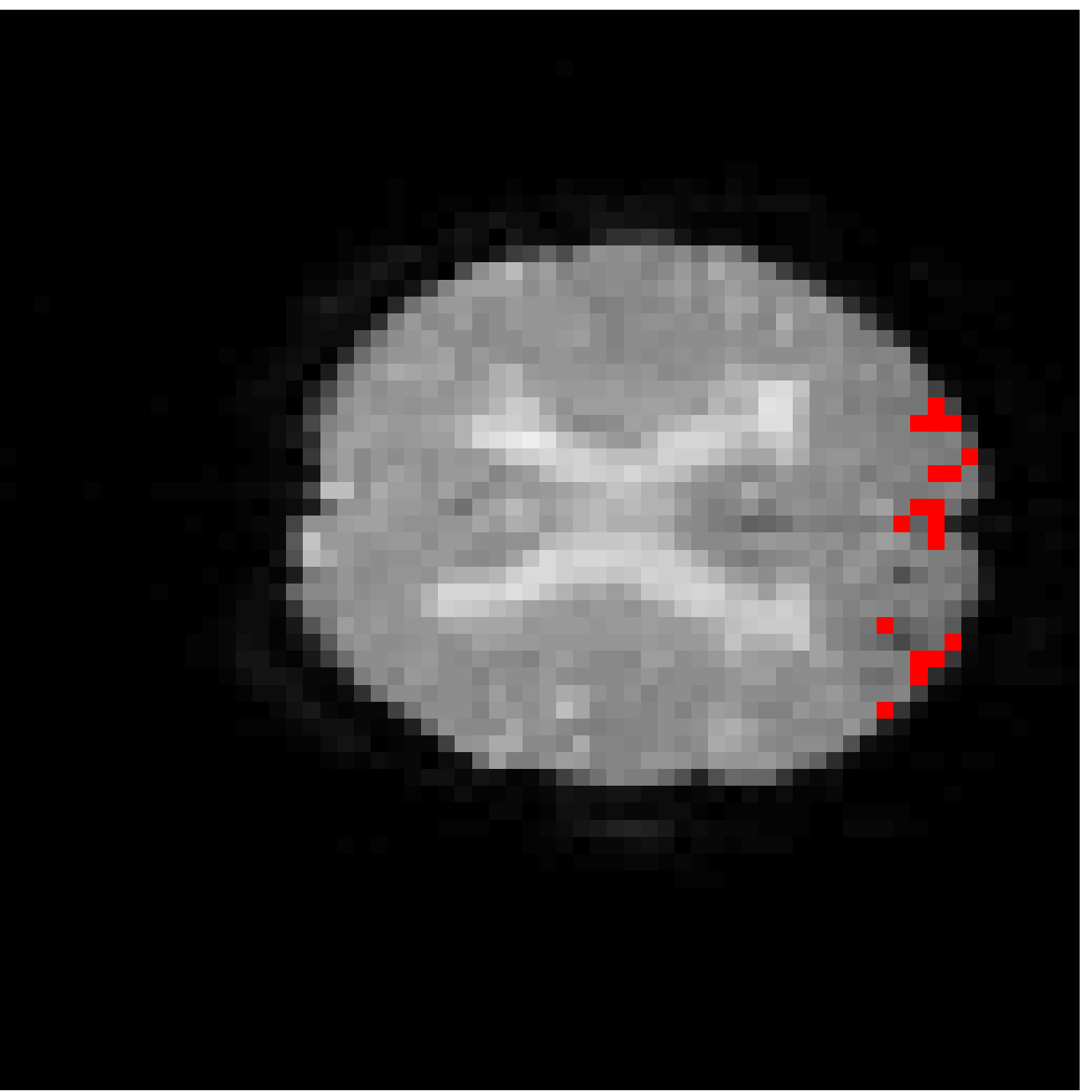}
}
\caption{Top two rows: activation maps: cluster~I (blue) , cluster~II
  (red). Bottom two rows: activation maps obtained using the linear regression model
($p=0.005$).
\label{dementia_map}}
\end{figure}

{\noindent \em Embedding the dataset in 3 dimensions}\\
We display in Fig. \ref{dem_embedding}-left the 2050 time series after
embedding the dataset in three dimension. The {\em one-trial}
condition time series (blue) form a blob at the center. The {\em
  two-trial} (red) conditions time series form a ``V''
(Fig. \ref{dem_embedding}-right).  Both branches of the ``V'' are
nearly one dimensional and are aligned with $\bfi_2$ and $\bfi_4$. The
branch aligned with $\bfi_2$ also includes {\em one-trial} condition
time series at the base of the branch. The branch aligned with
$\bfi_4$ contains only {\em two-trial} condition time series.  The
dataset was partitioned into three clusters
(Fig. \ref{dem_embedding}-right).  We compare our embedding to the
embedding generated by ISOMAP (Fig. \ref{dem_pca}-left) and PCA
(Fig. \ref{dem_pca}-right).  No low-dimensional structures emerge from
the representations given by PCA or ISOMAP, and {\em two-trial} and
{\em one-trial} time series are mixed together. The eigenvectors
$\bfi_2$ and $\bfi_4$ create the largest drop in the residual energy,
and are the most useful coordinates.  We use $K=3$ coordinates:
$\bfi_2, \bfi_3$ and $\bfi_4$ to embed the dataset. To determine the
role of the clusters, we select three groups of four time series
(identified by circles in Fig. \ref{dem_embedding}-right) and we plot
them in Fig.  \ref{dementia_kmean_ts}.  Time series from cluster~I all
have an abrupt dip at $t=7$. The corresponding voxels are located
along the border of the brain (Fig. \ref{dementia_map}).  The original
time series (before averaging) suffer from a sudden drop at time 95,
which could be caused by a motion artifact, that affects the average
time series.  There are two groups of time series selected from
cluster~II: two {\em two-trial} condition time series located at the
tip of the branch, and two {\em one-trial} condition time series
located at the border with the background cluster.  Time series from
cluster~II have a shape similar to an hemodynamic response
(Fig. \ref{dementia_kmean_ts}-B and C), and the corresponding voxels
are located in the visual cortex (Fig. \ref{dementia_map}). Therefore,
we hypothesize that cluster~II contains times series recruited by the
stimulus. Interestingly, the embedding has organized the time series
along the branch%
\noindent of cluster~II according to the strength of the activation:
from {\em two-trial} condition (strong response) at the tip, to {\em
  one-trial} condition (weak response) at the stem (close to the
background time series).  This is a remarkable result since no
information about the stimulus, or the type of trial was provided to
the algorithm. Fig. \ref{dementia_map} shows the location of the
voxels corresponding to the time series of cluster~I (blue) and II
(red). For comparison purposes we computed the activation map obtained
using the GLM. The averaged time series from the {\em two-trial}
condition are used for the regression analysis. We use the hemodynamic
response function defined by \citep{Dale97}, $h(t) =
((t-\delta)/\tau)^2 e^{(t-\delta)/\tau}$, where $\delta=2.5$,
$\tau=1.5$. The regressor was given by $g(t) = h(t) \ast s(t)$, where
$s(t)$ is the stimulus time series. We thresholded the $p$-value at
$p=0.005$, and the activation maps are shown in
Fig.\ref{dementia_map}. The activation maps constructed by our
approach are consistent with the maps obtained with a GLM.
\subsection{Complex natural stimuli}
We demonstrate here that our method can be used to analyze fMRI
datasets collected in natural environments where the subjects are
bombarded with a multitude of uncontrolled stimuli that cannot be
quantified \citep{Golland07,Hason04,Haynes06,Malinen07,Meyer08}.
During such experiments, the subjects are submitted to an abundance of
concurrent sensory stimuli, which makes the analysis with inferential
methods impossible.%
\noindent  Exploratory techniques can help unravel the
different neural processes involved during the experiments
\citep{Malinen07}.  Unlike ICA, our approach does not posit the
existence of a mixture model. Our approach merely explores the
connectivity in the dataset, and proposes a new parameterization that
preserves connectivity.

\begin{figure}[H]
\centerline{
  \includegraphics[width=12pc]{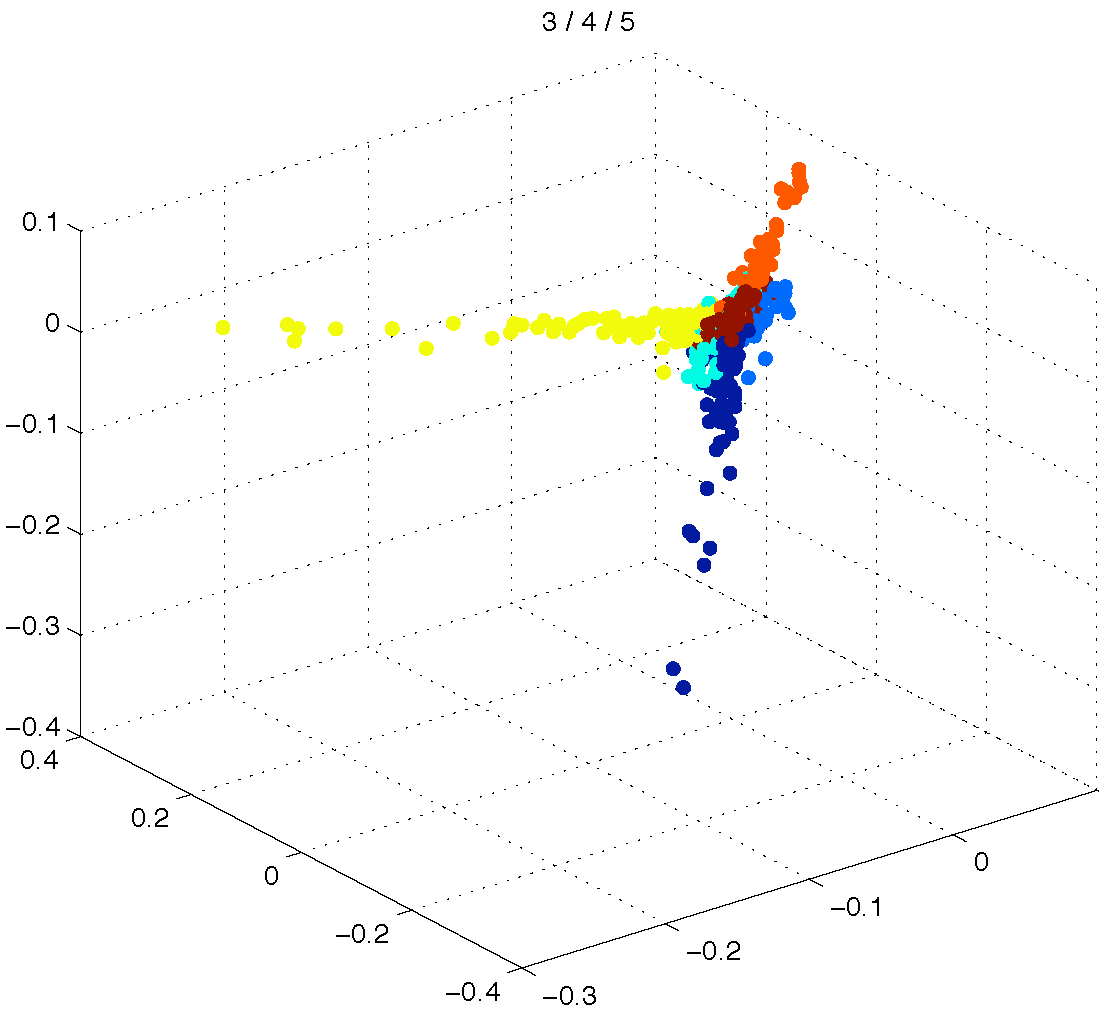}
}
\centerline{
  \includegraphics[width=12pc]{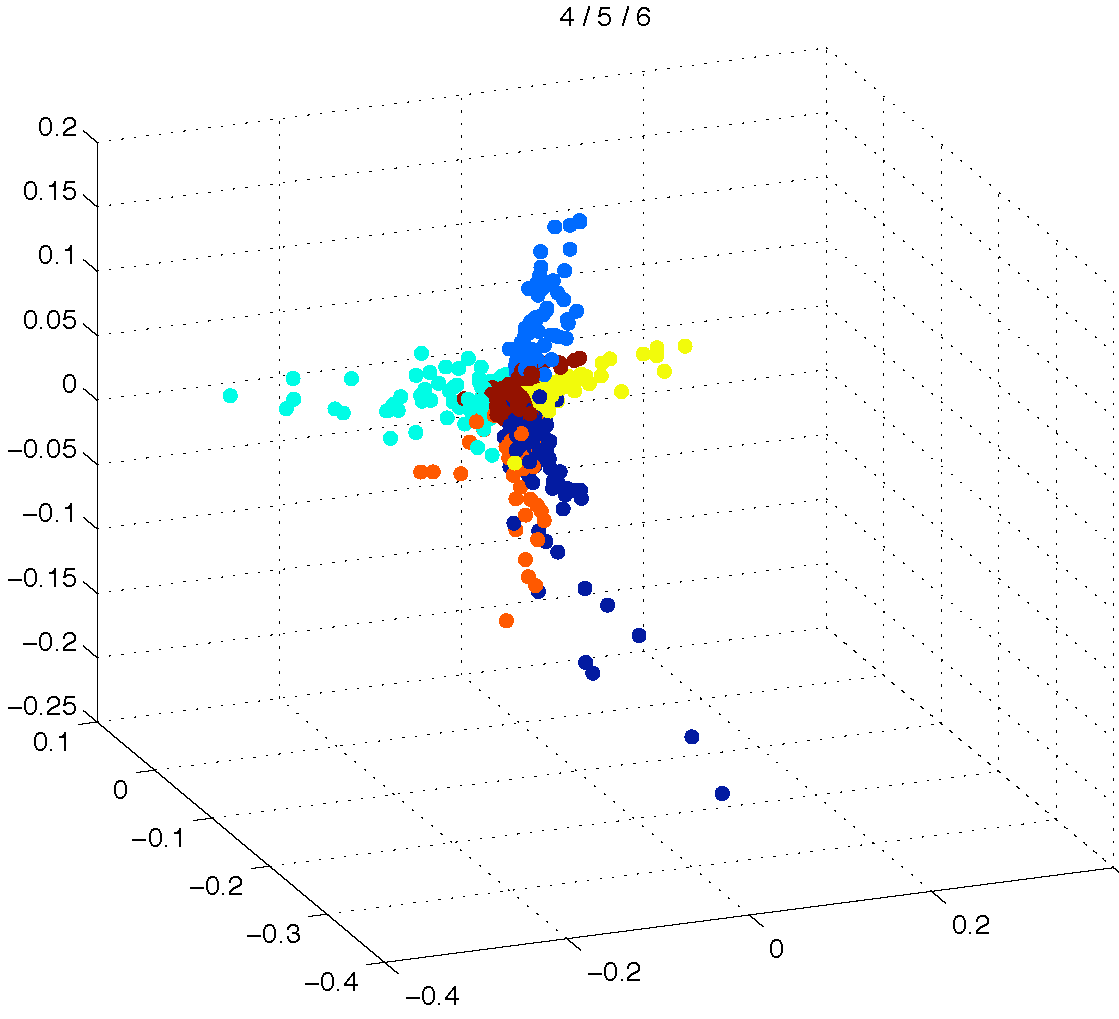}
}
\centerline{
  \includegraphics[width=12pc]{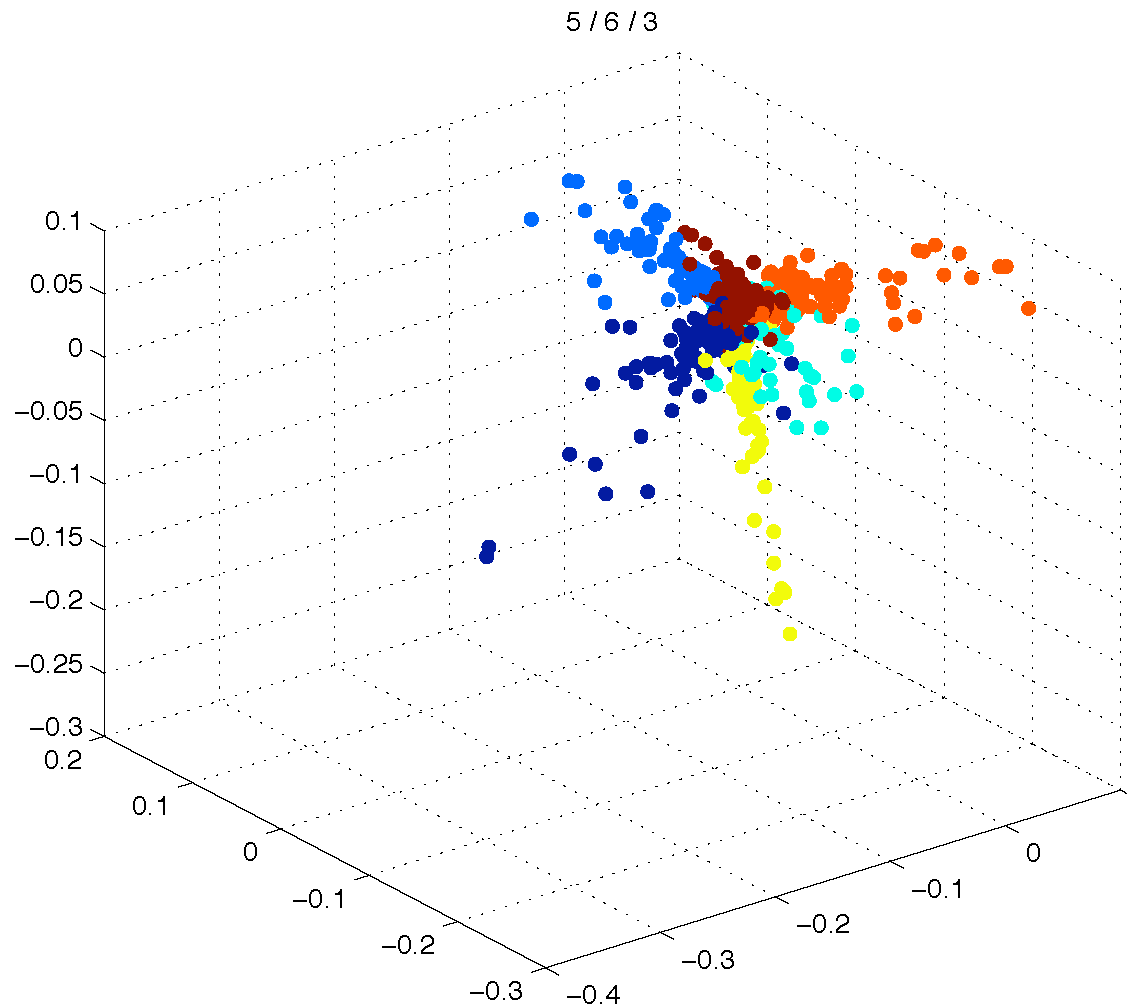}
}
\caption{Subject 14: views of the  embedded dataset using 
$\bfi_3,\bfi_4,\bfi_5$ (left),
$\bfi_4,\bfi_5,\bfi_6$ (center), and
$\bfi_5,\bfi_6,\bfi_3$ (right). The labels are obtained after
clustering the datasets into six clusters. The background activity
(maroon cluster) is centered around 0.
\label{sub14_proj}}
\end{figure}

\begin{figure}[H]
\centerline{
  \includegraphics[width=18pc]{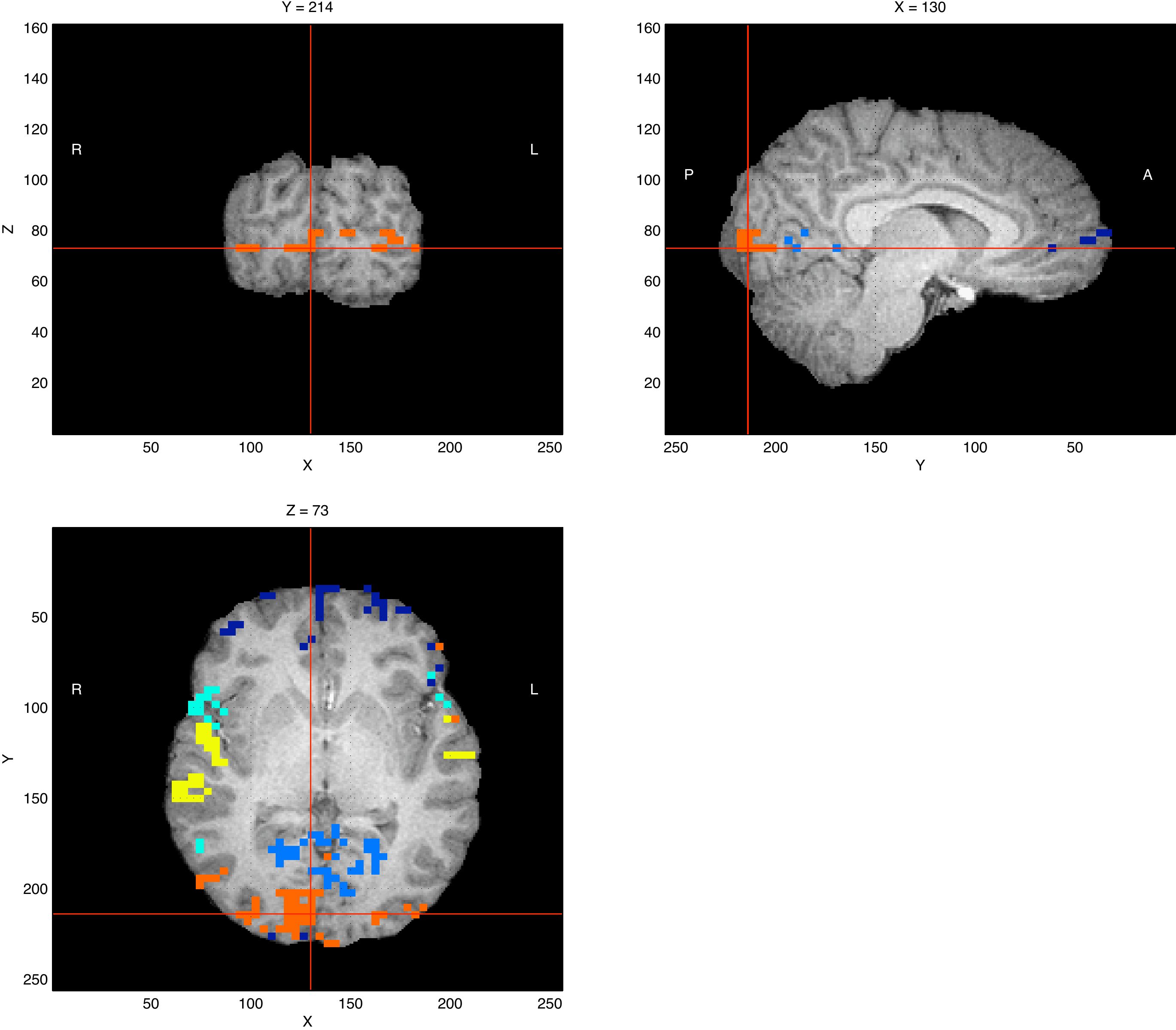}
}
\centerline{
  \includegraphics[width=18pc]{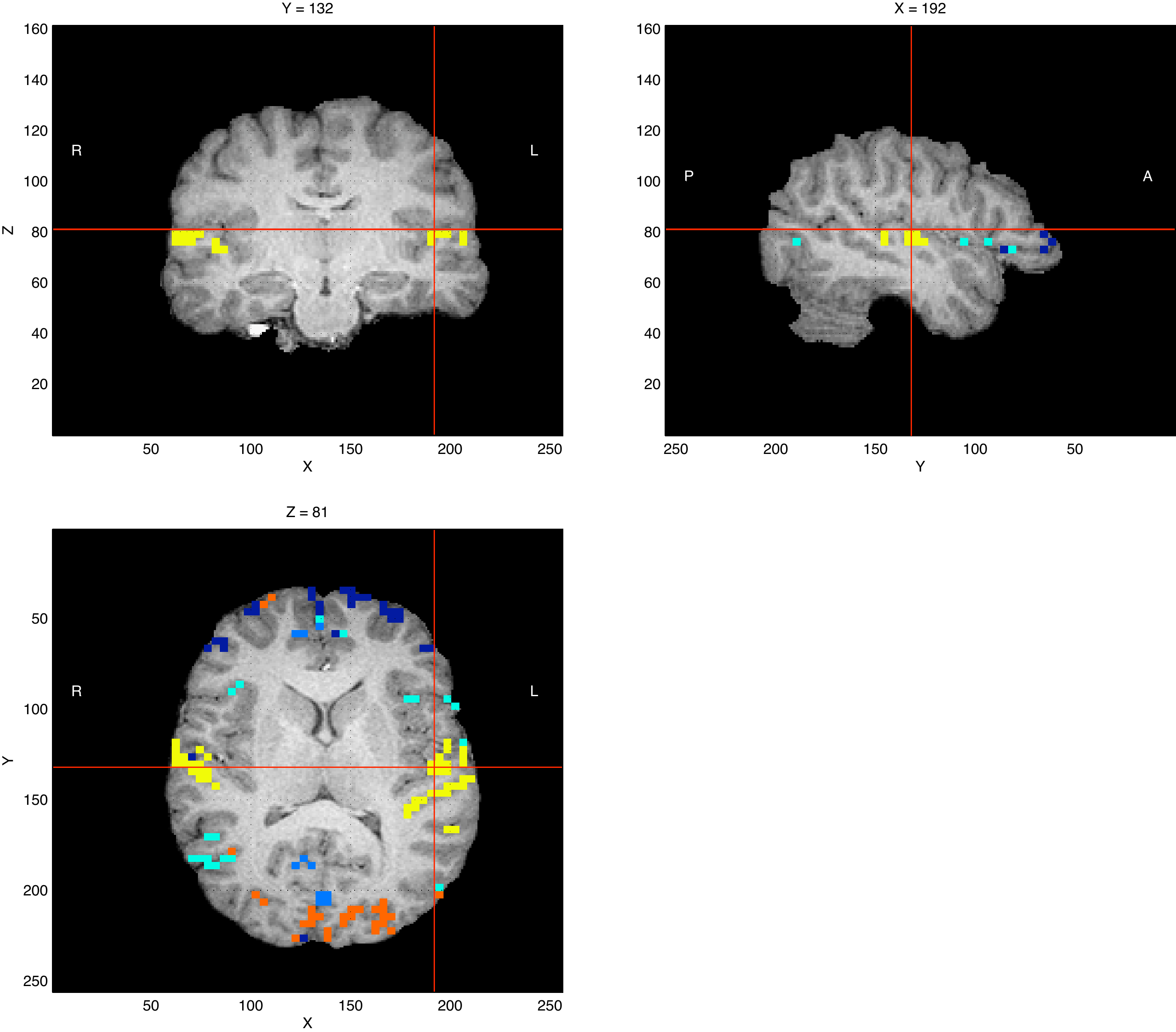}
}
\caption{Subject 14.  Top: V1/V2 (orange);
   representations of the upper visual fields (light blue); Broca area
   (cyan). Bottom:  Wernicke area (yellow); prefrontal cortex
  (dark blue).
\label{sub14_map}}
\end{figure}

\begin{figure}[H]
\centerline{
  \includegraphics[width=12pc]{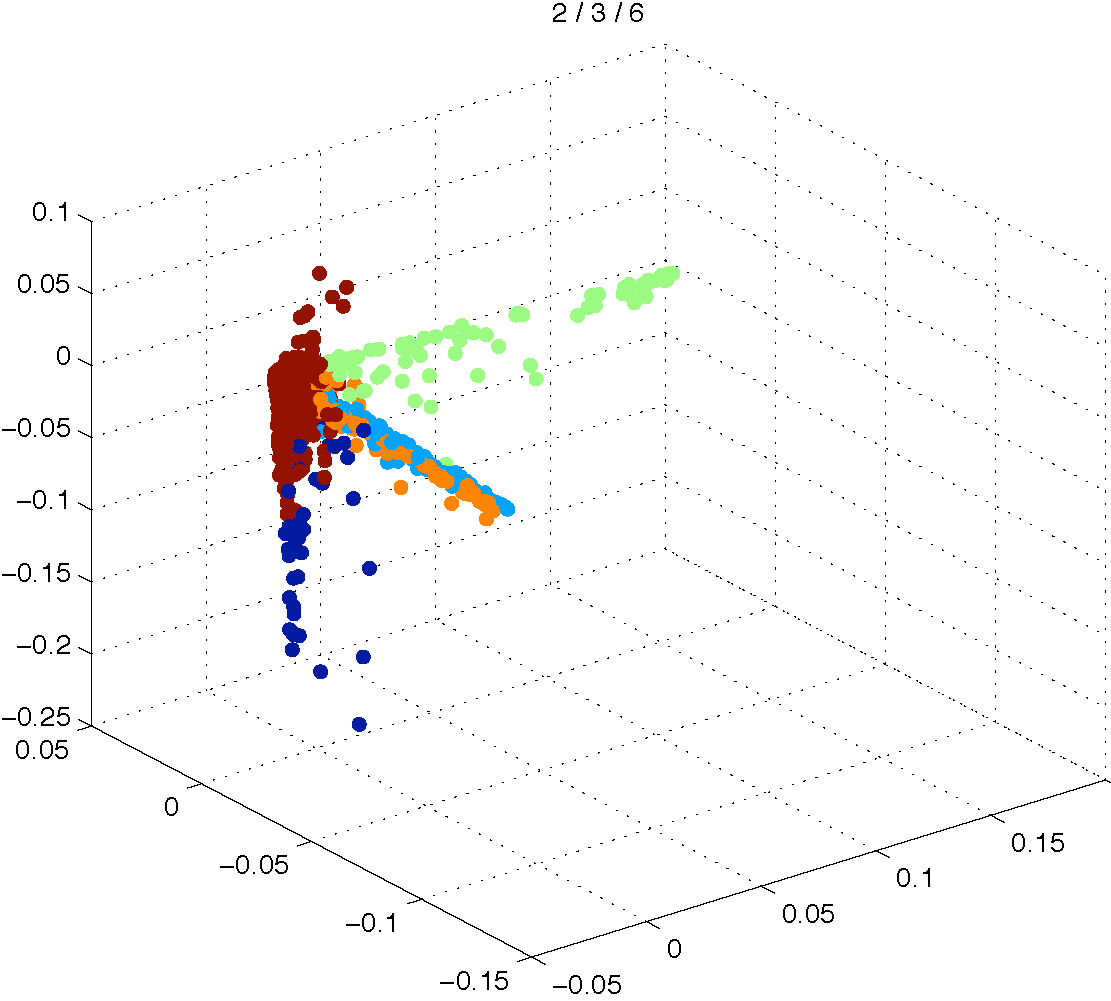}
}
\centerline{
  \includegraphics[width=12pc]{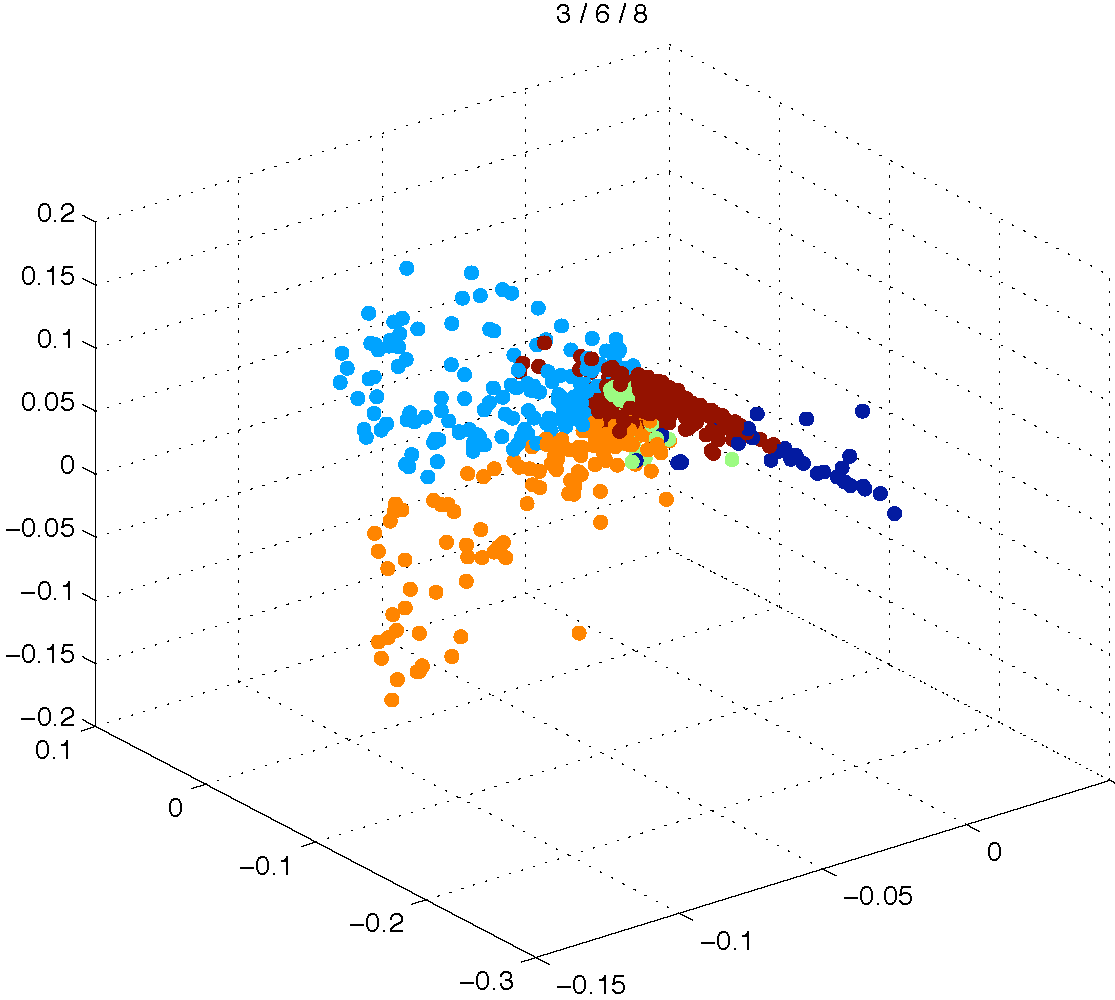}
}
\centerline{
  \includegraphics[width=12pc]{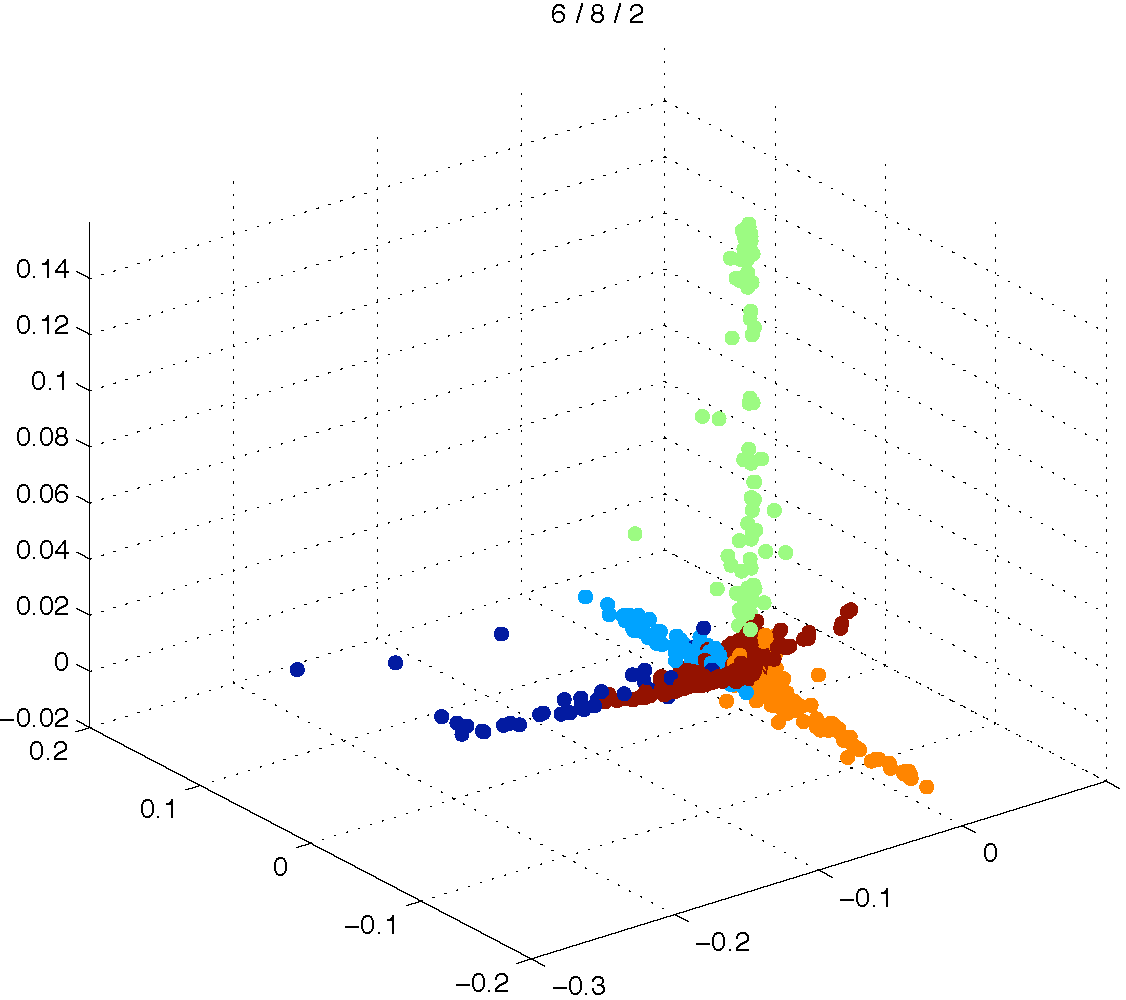}
}
\caption{Subject 13: views of the  embedded dataset using 
$\bfi_2,\bfi_3,\bfi_6$ (left),
$\bfi_3,\bfi_6,\bfi_8$ (center), and
$\bfi_6,\bfi_8,\bfi_2$ (right). The labels are obtained after
clustering the datasets into five clusters. The background activity
(maroon cluster) is centered around 0.
\label{sub13_proj}}
\end{figure}

\begin{figure}[H]
\centerline{
  \includegraphics[width=18pc]{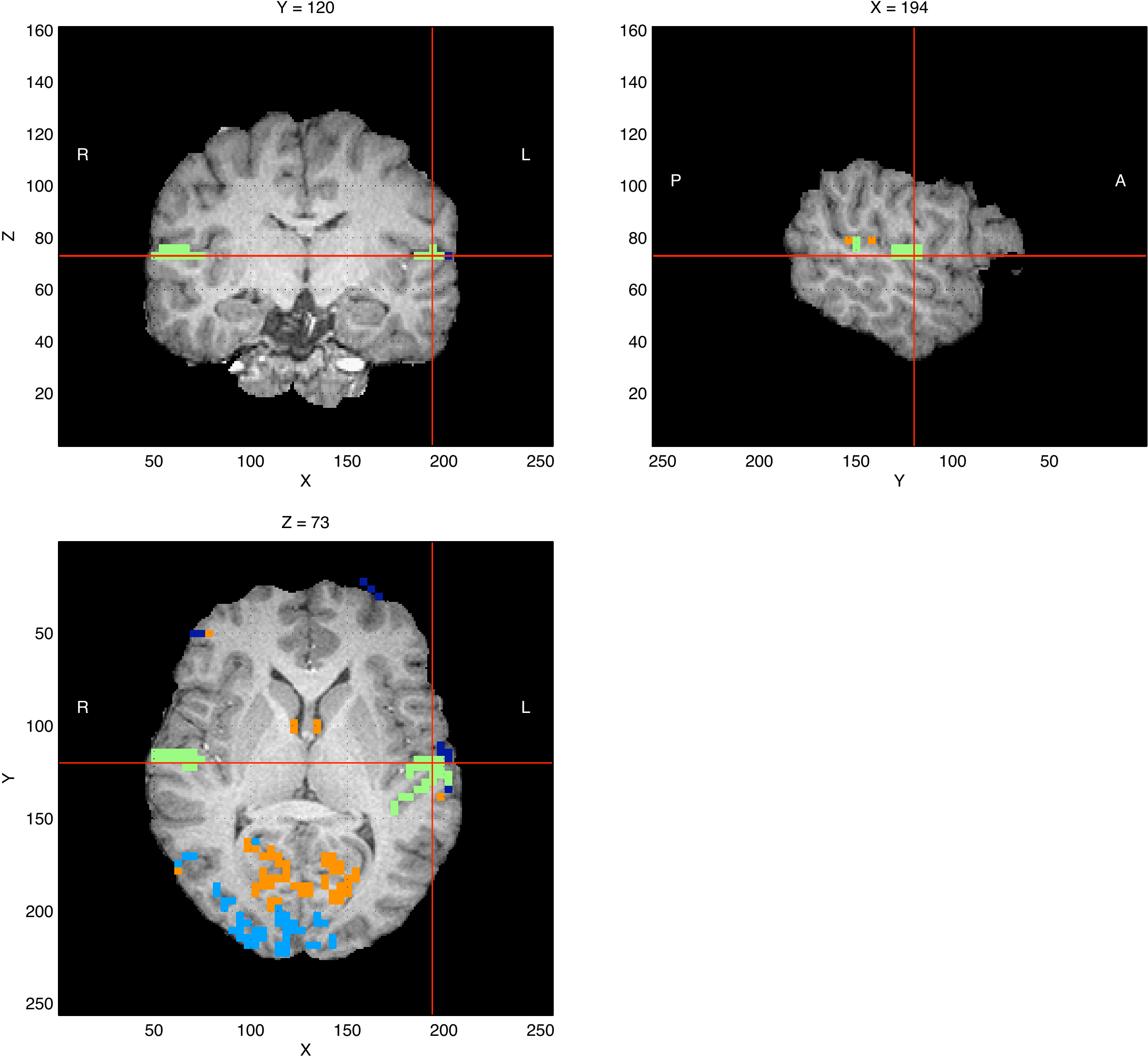}
}
\centerline{
  \includegraphics[width=18pc]{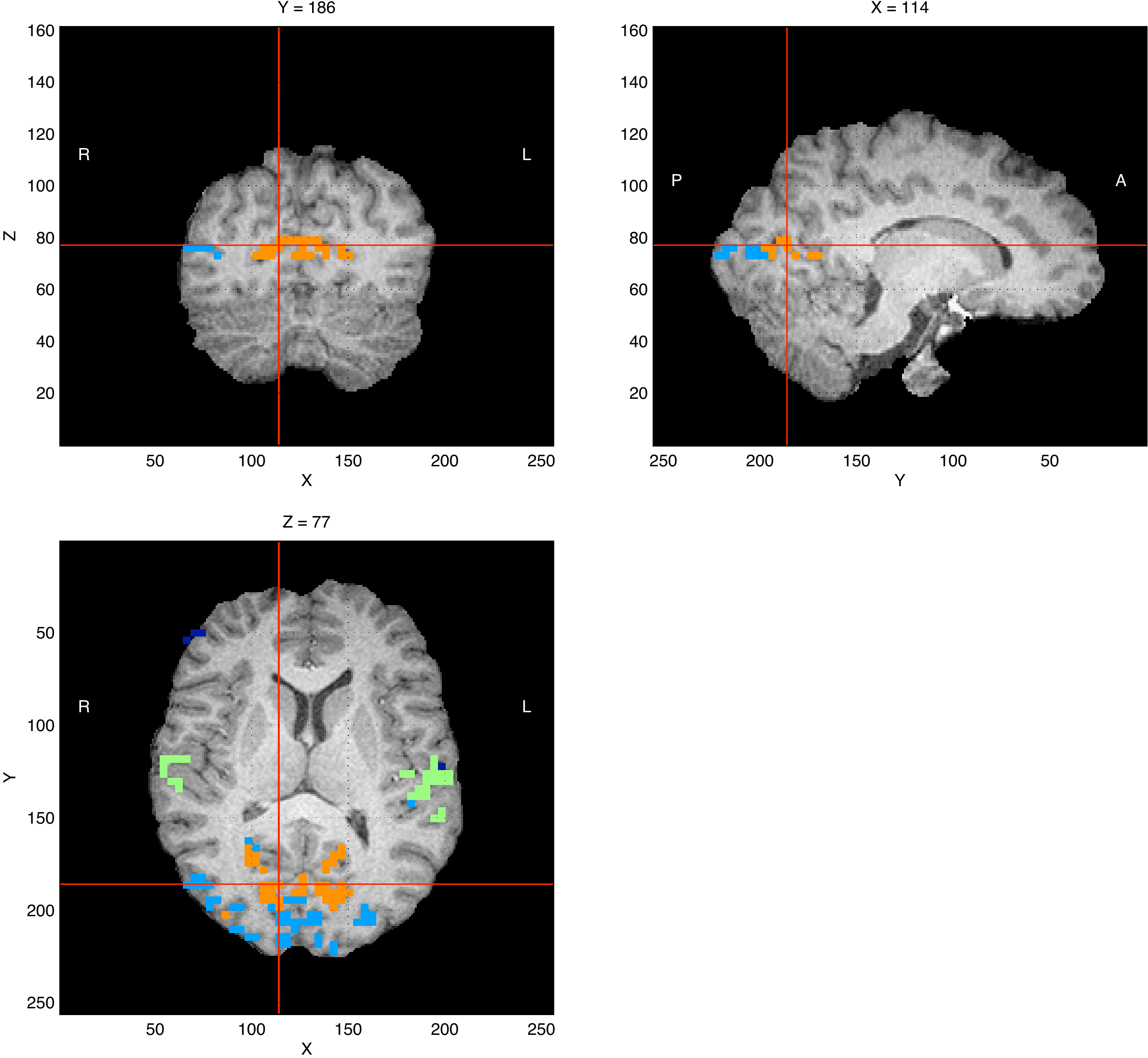}
}
\caption{Subject 13. Top: Wernicke area (green). Bottom: V1/V2 (orange);
   representations of the upper visual fields (light blue); lateral
   areas (light blue): visual motion areas (MT/V5).
   \label{sub13_map}}
\end{figure}

The Experience Based Cognition competition (EBC) \citep{EBC} offers an
opportunity to study complex responses to natural environments.  The
EBC datasets comprise three 20-minute runs (704 TRs in each run) of
subjects interacting in an urban virtual reality environment. Subjects
were audibly instructed to complete three search tasks in the
environment: looking for weapons (but not tools) taking pictures of
people with piercing (but not others), or picking up fruits (but not
vegetables).  The data was collected with a 3T EPI scanner (TR =
$1.75$s, xy dimension: $64\times 64$, voxel size = $3.28 \times
3.28$mm, 34 slices with a thickness of $3.5$mm).  We analyze the
second runs of subjects 14 and 13. For each subject, the matrix $\bX$
is composed of $N=4843$ intra-cranial voxels at $T=704$ TRs. We first
remove the non regionally specific variance captured by the first
eigenmodes of a singular value decomposition of the dataset. We then
compute $\bfi_k,k=2,\cdots 10$ using $n_n=100$ and $\sigma = 2 \times \min_{i<j}
\|\bx_i-\bx_j\|$. After embedding the dataset into four dimensions, we
cluster the voxels. Figs. \ref{sub14_proj} and \ref{sub13_proj}
display the datasets after embedding. Because we cannot display four
dimensions, we show the projections of the dataset on three
consecutive coordinates. All the coordinates contribute to the spread
the dataset along elongated arms, which facilitates the
clustering. Voxels that do not correspond to the background activity
(the maroon cluster in Figs. \ref{sub14_proj} and \ref{sub13_proj})
are superimposed on anatomically registered structural images and
colored according to their cluster label (see Figs. \ref{sub14_map}
and \ref{sub13_map}). For both subjects, the clusters are connected
regions (see Figs. \ref{sub14_map} and \ref{sub13_map}), compactly
organized around functional areas related to the processing of visual,
and auditory stimuli (music, cellphone ringing, dog roaring) in the
virtual environment. It is important to emphasize that our method
never enforces any form of spatial proximity, and is purely based on
functional connectivity.

For subject 14 (Fig. \ref{sub14_map}), the orange cluster corresponds
to activation in the calcarine cortex associated with V1/V2
representations of the lower visual fields, while the light blue
cluster corresponds to representations of the upper
 visual fields. Activation in lateral areas (visual motion
areas, MT/V5) is also present, as well as activity in the posterior
convexial cortex (area VP). The activation is predominantly in the
right hemisphere. Interestingly, the two clusters located in the
visual cortex (light blue and orange) have very similar $\bfi_3$ and
$\bfi_4$ coordinates (see Fig. \ref{sub14_proj}-left). The cyan
cluster corresponds to activation in the right frontal gyrus (Broca
area) associated with language comprehension. The yellow clusters are
located in the right and left superior temporal gyri and medial
temporal gyri (Wernicke area). These regions correlate with activation%
%
\noindent in the auditory cortex and language areas. Finally, the dark blue
cluster corresponds to activation in the prefrontal cortex. A very
similar pattern of activity (Fig.  \ref{sub13_map}) was obtained for
subject 13. The blue and orange clusters, located in the calcarine
cortex, correspond to V1 and V2 areas. Again, these two clusters, both
located in the visual cortex, have similar $\bfi_2$ and $\bfi_6$
coordinates. The green cluster is located in the medial temporal gyrus
(Wernicke area) and is associated with language processing.

We replaced the embedding constructed by our method with the
parametrization produced by PCA, using the same pre-processing
steps. PCA was%
\noindent  unable to produce any meaningful activation maps (results not
shown). In fact, the clustering would usually not converge.
Interestingly, the activation maps obtained with these natural stimuli
are very similar to the extrinsic network, found by \cite{Golland07},
that is composed of areas dedicated to the processing of
sensory information: auditory, visual, somatosensory and language.
\section{Discussion}
We proposed a new method to compute a low dimensional embedding of an
fMRI dataset. The embedding preserves the local functional
connectivity between voxels, and can be used to cluster the time
series into coherent groups.  Our approach, based on%
\noindent a spectral
decomposition of commute time, appears to be more robust than a method
based on the computation of the geodesic distance between time series
\citep{Tenenbaum00}. Our approach is able to detect independently
visual areas (V1/V2, V5/MT), auditory and language areas that are
recruited when subjects interacted in an urban virtual reality
environment.  We believe that this approach offers a new approach for
the analysis of natural stimuli. The method still requires the visual
inspection of the spatial patterns of activations. This is a standard
limitation of exploratory methods. We are currently exploring methods to
combine our approach with standard functional atlases.%
{\noindent \em Choice of the parameters}\\
There are only two parameters that determine the embedding: $n_n$ the
number of nearest neighbors and the scaling factor $\sigma$.  We
described in section \ref{connectgraph} a universal choice for
$\sigma$.  The number of nearest neighbors can also be chosen
according to a universal strategy, $n_n$ can be assigned to the
largest power of ten smaller than the number of time samples (TRs) in
the time series.  For instance, for the datasets for the event-related
and block design paradigm respectively we used $n_n=7$ and $n_n=9$
. The analysis of the EBC datasets that contain $704$ TRs each
required a much larger number of nearest neighbors: $n_n=100$.

{\noindent \em Computational complexity}\\
The complexity of our method is determined by the combined complexity
of the nearest neighbor search and the eigenvalue problem.  We use a
brute force approach to search for nearest neighbors. We use the
restarted Arnoldi method for sparse matrices implemented by the Matlab
function {\tt eigs} to solve the eigenvalue problem.  The algorithm
requires $N{\cal O}(K) + {\cal O}(K^2)$ storage, and ${\cal O}(NK^2) +
{\cal O}(K^3)$ computations.  Our MATLAB code will
   be made available shortly from our webpage.
\ack
The authors are very grateful to the reviewers for their comments and
suggestions. This work was initiated at the Institute for Pure and
Applied Mathematics, UCLA which provided partial support for the
authors. The authors are very grateful to Raphy Coifman, St\'{e}phane
Lafon, and Mauro Maggioni for introducing them to diffusion maps, and
to Ivo Dinov for making the event-related fMRI dataset available for
this study.  FGM was on sabbatical at the Center for the Study of
Brain, Mind and Behavior, Princeton University, when he performed the
analysis of the EBC datasets, and he is very grateful to all the
members of the center for their support and insightful discussions.
\section{Appendix}
The hitting time satisfies the following one-step equation,
\begin{equation}
E_i[T_j] = 1 + \sum_{k;k\neq j} P_{i,k}E_i[T_k].
\label{onestep}
\end{equation}
Iterating this equation yields an expression of $E_i[T_j]$ in terms of
  powers of $\mathbf P$. Let us define the fundamental matrix
  \citep{Bremaud99},  
\begin{equation}
{\mathbf Z}=  I + \sum_{k\ge 1} {\mathbf P^k - \mathbf \Pi}.
\label{fundamental}
\end{equation}
then we have,
\begin{equation}
E_i[T_j] = (Z_{j,j} - Z_{i,j})/\pi_j.
\label{green}
\end{equation}
The proof is a straightforward consequence of (\ref{onestep}), and can be
found in \citep{Bremaud99}. Note that $\mathbf Z= (\mathbf I - ({\mathbf P
  - \mathbf \Pi}))^{-1}$  is also the Green function of the Laplacian,
which explains the connection with the graph Laplacian.
We can consider the  eigenvectors  $\bfi_1, \cdots, \bfi_T$ of the
symmetric matrix
\begin{equation}
{\mathbf D}^{\frac{1}{2}}{\mathbf P}{\mathbf D}^{-\frac{1}{2}},  
\label{N}
\end{equation}
and write $\mathbf P$ in  (\ref{green}) in terms of the eigenvectors. We obtain
\begin{equation}
\kappa(i,j)=\sum_{k=2}^T \frac{1}{1 -\lambda_k} 
\left (\frac{\bfi_k(i)}{\sqrt{\pi_i}} -\frac{\bfi_k(j)}{\sqrt{\pi_j}} \right)^2.
\label{isometry0}
\end{equation}
Finally, we note that $\mathbf D^{1/2} \mathbf P \mathbf D^{-1/2} = \mathbf
D^{-1/2} \mathbf W \mathbf D^{-1/2}$. 
\bibliographystyle{elsart-harv}

\begin{thebibliography}{30}
\expandafter\ifx\csname natexlab\endcsname\relax\def\natexlab#1{#1}\fi
\expandafter\ifx\csname url\endcsname\relax
  \def\url#1{\texttt{#1}}\fi
\expandafter\ifx\csname urlprefix\endcsname\relax\def\urlprefix{URL }\fi

\bibitem[{Achard et~al.(2006)Achard, Salvador, Whitcher, Suckling, and
  Bullmore}]{Achard06}
Achard, S., Salvador, R., Whitcher, B., Suckling, J., Bullmore, E., 2006. A
  resilient, low-frequency, small-world human brain functional network with
  highly connected association cortical hubs. The Journal of Neuroscience
  26(1), 63--72.

\bibitem[{Albert and Barab\'asi(2002)}]{Albert02}
Albert, R., Barab\'asi, A., 2002. Statistical mechanics of complex networks.
  Reviews of modern physics 74, 47--97.

\bibitem[{B.B.Biswal and Ulmer(1999)}]{Biswal99}
B.B.Biswal, Ulmer, J., 1999. Blind source separation of multiple signal sources
  of \mbox{fMRI} data sets using independent component analysis. Journal of
  Computer Assisted Tomography 23(3), 265--271.

\bibitem[{Belkin and Niyogi(2003)}]{Belkin03}
Belkin, M., Niyogi, P., 2003. Laplacian eigenmaps for dimensionality reduction
  and data representation. Neural Computations 15, 1373--1396.

\bibitem[{B\'erard et~al.(1994)B\'erard, Besson, and Gallot}]{Berard94}
B\'erard, P., Besson, G., Gallot, S., 1994. Embeddings \mbox{Riemannian}
  manifolds by their heat kernel. Geometric and Functional Analysis 4(4),
  373--398.

\bibitem[{Bremaud(1999)}]{Bremaud99}
Bremaud, P., 1999. Markov Chains. Springer Verlag.

\bibitem[{Buckner et~al.(2000)Buckner, Snyder, Sanders, Raichle, and
  Morris}]{Buckner00}
Buckner, R., Snyder, A., Sanders, A., Raichle, M., Morris, J., 2000. Functional
  brain imaging of young, nondemented, and demented older adults. Journal of
  Cognitive Neuroscience 12, 24--34.

\bibitem[{Bullmore et~al.(2001)Bullmore, Long, Suckling, Fadili, Calvert, and
  {F. Zelaya {\em et al.}}}]{Bullmore01}
Bullmore, E., Long, C., Suckling, J., Fadili, J., Calvert, G., {F. Zelaya {\em
  et al.}}, 2001. Colored noise and computational inference in
  neurophysiological (\mbox{fMRI}) time series analysis~: resampling methods in
  time and wavelet domain. Human Brain Mapping 78, 61--78.

\bibitem[{Caclin and Fonlupt(2006)}]{Caclin06}
Caclin, A., Fonlupt, P., 2006. Effect of initial \mbox{fMRI} data modeling on
  the connectivity reported between brain areas. NeuroImage 33, 515--521.

\bibitem[{Chung(1997)}]{Chung97}
Chung, F., 1997. Spectral Graph Theory. CBNS-AMS.

\bibitem[{Coifman and Lafon(2006)}]{Coifman06a}
Coifman, R., Lafon, S., 2006. Diffusion maps. Applied and Computational
  Harmonic Analysis 21, 5--30.

\bibitem[{Dale and Buckner(1997)}]{Dale97}
Dale, A.~M., Buckner, R.~L., 1997. Selective averaging of rapidly presented
  individual trials using \mbox{fMRI}. Human Brain Mapping 5, 329--340.

\bibitem[{Egu\mbox{\'i}luz et~al.(2005)Egu\mbox{\'i}luz, Chialvo, Cecchi,
  Baliki, and Apkarian}]{Eguiluz05}
Egu\mbox{\'i}luz, V., Chialvo, D., Cecchi, G., Baliki, M., Apkarian, A., 2005.
  Scale-free brain functional networks. Physical Review Letters 94(018102).

\bibitem[{Fox et~al.(2005)Fox, Snyder, Vincent, Corbetta, \mbox{Van Essen}, and
  Raichle}]{Fox05}
Fox, M., Snyder, A., Vincent, J., Corbetta, M., \mbox{Van Essen}, D., Raichle,
  M., 2005. The human brain is intrinsically organized into dynamic,
  anticorrelated functional networks. Proc. of the National Academy of Sciences
  102(27), 9673--78.

\bibitem[{Friston(2005)}]{Friston05}
Friston, K., 2005. Models of brain function in neuroimaging. Annual Reviews of
  Psychology 56, 57--87.

\bibitem[{Friston(1998)}]{Friston98b}
Friston, K.~J., 1998. Modes or models: a critique on independent component
  analysis for \mbox{fMRI}. Trends in Cognitive Sciences 2~(10), 373--375.

\bibitem[{Golland et~al.(2007)Golland, Bentin, Gelbard, Benjamini, Heller, Nir,
  Hasson, and Malach}]{Golland07}
Golland, Y., Bentin, S., Gelbard, H., Benjamini, Y., Heller, R., Nir, Y.,
  Hasson, U., Malach, R., 2007. Extrinsic and intrinsic systems in the
  posterior cortex of the human brain revealed during natural sensory
  stimulation. Cerebral Cortex 17, 766--777.

\bibitem[{Haynes and Rees(2006)}]{Haynes06}
Haynes, J.-D., Rees, G., Apr. 2006. Decoding mental states from brain activity
  in humans. Nature Neuroscience 7.

\bibitem[{Malinen et~al.(2007)Malinen, Hlushchuk, and Hari}]{Malinen07}
Malinen, S., Hlushchuk, Y., Hari, R., 2007. Towards natural stimulation in
  \mbox{fMRI}--issues of data analysis. NeuroImage 35, 131--139.

\bibitem[{McKeown et~al.(2003)McKeown, Hansen, and Sejnowski}]{McKeown03}
McKeown, M., Hansen, L., Sejnowski, T., 2003. Independent component analysis of
  functional \mbox{MRI}: what is signal and what is noise? Current Opinion in
  Neurobiology 13, 620--629.

\bibitem[{Meyer and Stephens(2008)}]{Meyer08}
Meyer, F., Stephens, G., 2008. Locality and low-dimensions in the prediction of
  natural experience from \mbox{fMRI}. In: Advances in Neural Information
  Processing Systems. MIT Press.

\bibitem[{Petersson et~al.(1999)Petersson, Nichols, Poline, and
  Holmes}]{Petersson99b}
Petersson, K.~M., Nichols, T., Poline, J.-B., Holmes, A., 1999. Statistical
  limitations in functional neuroimaging \mbox{II. Signal} detection and
  statistical inference. Phil. Trans. R. Soc. Lond. B~(354), 1261--81.

\bibitem[{Raichle and Mintun(2006)}]{Raichle06a}
Raichle, M., Mintun, M., 2006. Brain work and brain imaging. Annual review of
  neuroscience 29, 449--76.

\bibitem[{Sporns et~al.(2000)Sporns, Toroni, and Edelman}]{Sporns00}
Sporns, O., Toroni, G., Edelman, G., 2000. Theoretical neuroanatomy: relating
  anatomical and functional connectivity in graphs and cortical connection
  matrices. Cerebral Cortex 10, 127--141.

\bibitem[{Tanabe et~al.(2002)Tanabe, Miller, Tregellas, Freedman, and
  Meyer}]{Tanabe01a}
Tanabe, J., Miller, D., Tregellas, J., Freedman, R., Meyer, F., 2002.
  Comparison of detrending methods for optimal \mbox{fMRI} pre-processing.
  NeuroImage 15, 902--907.

\bibitem[{Tenenbaum et~al.(2000)Tenenbaum, de~Silva, and
  Langford}]{Tenenbaum00}
Tenenbaum, J., de~Silva, V., Langford, J., 2000. A global geometric framework
  for nonlinear dimensionality reduction. Science 290, 2319--2322.

\bibitem[{Thirion et~al.(2006)Thirion, Dodel, and Poline}]{Thirion06}
Thirion, B., Dodel, S., Poline, J.-B., 2006. Detection of signal
  synchronizations in resting-state \mbox{fMRI} datasets. Neuroimage 2,
  321--27.

\bibitem[{Thirion and Faugeras(2003)}]{Thirion03c}
Thirion, B., Faugeras, O., 2003. Dynamical components analysis of \mbox{fMRI}
  data through kernel \mbox{PCA}. Neuroimage 20, 34--49.

\bibitem[{U.Hason et~al.(2004)U.Hason, Nir, Levy, Fuhrmann, and
  Malach}]{Hason04}
U.Hason, Nir, Y., Levy, I., Fuhrmann, G., Malach, R., 2004. Intersubject
  synchronization of cortical activity during natural vision. Science 303,
  1634--1640.

\bibitem[{U\mbox{niversity of Pittsburgh}(2007)}]{EBC}
U\mbox{niversity of Pittsburgh}, 2007. The experience based cognition project,
  http://www.ebc.pitt.edu.

\end{thebibliography}


\end{document}